\documentclass[12pt]{article}
\usepackage{graphicx,psfrag,epsf,color}
\usepackage{amssymb}
\usepackage{epsfig, graphicx, amsmath}
\usepackage{natbib}
\usepackage{multirow}
\usepackage{bm, color}
\usepackage{comment}
\usepackage{algorithmic}
\usepackage[linesnumbered, ruled, vlined]{algorithm2e}
\usepackage{hyperref}
\usepackage{caption}
\usepackage{subcaption}

\hypersetup{
	colorlinks=true,
	linkcolor=red,
	urlcolor=blue,
	citecolor=blue
}
\usepackage{enumerate}
\usepackage{url} 
\usepackage{array}
\usepackage{booktabs}
\usepackage{float}
\usepackage[font=small,labelfont=bf]{caption}
\setcitestyle{citesep={;}}
\usepackage[total={6.4in,8.6in}]{geometry}
\newcommand{\wh}[1]{\widehat{#1}}

\newtheorem{lemma}{Lemma}
\newtheorem{theorem}{Theorem}
\newtheorem{assumption}{Assumption}
\newtheorem{remark}{Remark}

\newenvironment{proof}[1][Proof]{\noindent \textbf{#1.} }{\  \rule{0.5em}{0.5em}}

\newtheorem{definition}{Definition}[section]

\def\EE{\mathbb{E}}

\def\given{\, | \,}

\def\begar{$$\begin{array}{lll}}
	\def\endar{\end{array}$$}
\def\begarlab{\begin{equation} \begin{array}{lll} \label}
		\def\endarlab{\end{array} \end{equation}}
\def\argmax{\arg\max}
\def\argmin{\arg\min}

\def\ds1{{\mathrm{1 \hspace{-2.6pt} I}}}

\def\calA{{\cal A}}
\def\calB{{\cal B}}

\def\calD{{\cal D}}

\def\calF{{\cal F}}
\def\calG{{\cal G}}
\def\calH{{\cal H}}
\def\calI{{\cal I}}

\def\calL{{\cal L}}

\def\calN{{\cal N}}
\def\calP{{\cal P}}

\def\calR{{\cal R}}

\def\calS{{\cal S}}
\def\calT{{\cal T}}

\def\calV{{\cal V}}
\def\calW{{\cal W}}

\def\calZ{{\cal Z}}

\def\floor#1{\lfloor #1 \rfloor}

\usepackage{xr}

\newcommand{\abs}[1]{|#1|}

\newcommand{\norm}[1]{\|#1\|}

\newcommand{\revise}[1]{{\color{blue}{#1}}}

\newcommand{\blind}{0}
\begin{document}

\def\spacingset#1{\renewcommand{\baselinestretch}%
	{#1}\small\normalsize} \spacingset{1}
%
%
\if0\blind
{
	\title{\bf STEEL: Singularity-aware \\ Reinforcement Learning\footnotetext{Author
	names are sorted alphabetically.\\
	This work does not relate to the position at Amazon.\\
 We thank Elynn Chen, Xi Chen, Max Cytrynbaum, Lars Peter Hansen,
Clifford M. Hurvich, Chengchun Shi, Jin Zhou and other seminar participants at LSE, UCL, Cambridge, NYU Stern, University of Chicago for useful comments.}}
		\date{First version: January 2023; Revised: \today }
\author{Xiaohong Chen$^{1}$, 
	Zhengling Qi$^{2}$ and Runzhe Wan$^{3}$\\ 
	\smallskip\\
	$^1$Cowles Foundation for Research in Economics, Yale University\\ $^2$Department of Decision Sciences, George Washington University \\ $^3$ Amazon}
	
	\maketitle
} \fi
\if1\blind
{
	\bigskip
	\bigskip
	\bigskip
	\begin{center}
		{\Large\bf STEEL: Singularity-aware \\ Reinforcement Learning}
	\end{center}
	\medskip
} \fi

\begin{abstract}
	Batch reinforcement learning (RL) aims at leveraging pre-collected data to find an optimal policy that maximizes the expected total rewards in a dynamic environment. The existing methods require absolutely continuous assumption (e.g., there do not exist non-overlapping regions) on the distribution induced by target policies with respect to the data distribution over either the state or action or both.
 We propose a new batch RL algorithm that allows for singularity for both state and action spaces (e.g., existence of non-overlapping regions between offline data distribution and the distribution induced by the target policies)
 in the setting of an infinite-horizon Markov decision process with continuous states and actions. 
 We call our algorithm STEEL: SingulariTy-awarE rEinforcement Learning.
 Our algorithm is motivated by a new error analysis on off-policy evaluation, where we use maximum mean discrepancy, together with distributionally robust optimization, to characterize the error of off-policy evaluation caused by the possible singularity and to enable model extrapolation.
 By leveraging the idea of pessimism and under some technical conditions, we derive a first finite-sample regret guarantee for our proposed algorithm under singularity. 
 Compared with existing algorithms, 
    by requiring only minimal data-coverage assumption, STEEL improves the applicability and robustness of batch RL. In addition, a two-step adaptive STEEL, which is nearly tuning-free, is proposed.
 Extensive simulation studies and one (semi)-real experiment on personalized pricing  demonstrate the superior performance of our methods in dealing with possible singularity in batch RL.
\end{abstract}

\baselineskip=21pt
\section{Introduction}\label{sec: introduction}

We study batch reinforcement learning (RL), which aims to learn an optimal policy using the pre-collected data without further interacting with the environment. Arguably, the biggest challenge for solving such a problem is the distributional mismatch between the batch data distribution and those induced by some candidate policies \citep{levine2020offline}. 

To avoid the issue of distributional mismatch, prior works require \textit{absolute continuity} assumption, e.g., there does not exist a non-overlapping region between the batch data distribution and the distribution induced by some policies, for finding an optimal policy. For example, classical methods such as \citep{antos_learning_2008} require the batch data to contain every possible state-action pair. Recent literature \citep[e.g.,][]{kumar2019stabilizing,jin2021pessimism} has found that, by incorporating the idea of pessimism, the requirement on the batch data is relaxed to contain state-action pairs induced by an optimal policy. However, the absolute continuity is not known a priori and could be easily violated. In this case, \textit{singularity} arises, e.g.,, there exists a non-overlapping region between the batch data distribution and the target distribution induced by some policy such as an optimal one, and all aforementioned methods fail. See formal definitions of absolute continuity and singularity in Section \ref{sec: prelim}.


To address the issue of singularity, in this paper, we propose an efficient and singularity-aware policy-iteration-type algorithm for finding an optimal policy under the framework of an infinite-horizon Markov decision process (MDP) with continuous states and actions. By saying an efficient algorithm, we refer to that the sample complexity requirement in finding an optimal policy is polynomial in terms of key parameters \citep{zhan2022offline}. We call our algorithm STEEL: SingulariTy-awarE rEinforcement Learning.



Our algorithm is motivated by a new error analysis on off-policy evaluation (OPE), which has been extensively studied in the recent literature \citep{shi2020statistical}. 
The goal of OPE is to use the batch data for evaluating the performance of other policies. 
By leveraging Lebesgue's Decomposition Theorem, we decompose the error of OPE into two parts: the absolutely continuous part and the singular one with respect to the data distribution.  For the absolutely continuous part, which can be calibrated by the behavior policy using the change of measure, standard OPE methods can be applied for controlling the error. For the singular part, we use the maximum mean discrepancy and leverage distributionally robust optimization to characterize its worst-case error measured by the behavior policy. 
Once we understand these two sources of the OPE error induced by any given target policy, a new estimating method for OPE without requiring absolute continuity can be formulated, based on which a policy iteration algorithm with pessimism is proposed. 

From theoretical perspective, we show that under some technical conditions, the regret of our estimated policy converges to zero at a satisfactory rate in terms of total decision points in the batch data even when singularity arises. 
This novel result demonstrates that our method can be more applicable in solving batch RL problems, in comparison to existing solutions.
More specifically, when the distribution induced by the optimal policy is covered by our batch data generating process in the usual manner, we recover the existing theoretical results given by those pessimistic RL algorithms \cite[e.g.,][]{xie2021bellman,fu2022offline}. In contrast, when absolute continuity fails, our algorithm is provable to find an optimal policy with a finite-sample regret warranty. To the best of our knowledge, this is the first finite-sample regret guarantee in  batch RL without assuming any form of absolute continuity. In contrast, existing methods such as \citep{antos_learning_2008,chen2016personalized,kallus2020doubly} can only allow singularity for certain state-action space and still require absolute continuity to some extent. For example, \cite{chen2016personalized,kallus2020doubly} leverage kernel smoothing techniques to address the issue of singularity on the action space, which is known to suffer from the curse of dimensionality when the action space is large. Moreover, our work can provide theoretical guidance on the existing algorithms aiming to find a deterministic optimal policy in a continuous action space.  For example, our STEEL method and the related theoretical results can be helpful for improving the understanding of the celebrated deterministic policy gradient method \citep{silver2014deterministic}.

To further demonstrate the effectiveness of our algorithm, we consider a contextual bandit problem and conduct extensive simulation studies. 
We show that compared with two existing baseline methods, under a possible singularity, our algorithm has a significantly better finite-sample performance.  In particular, we observe that our algorithm converges faster and is more robust when the singularity arises. We also apply our method to a personalized pricing application using the data from a US auto loan company, and find that our method outperforms the existing baseline methods.

Lastly, we propose a two-step adaptive STEEL that relieves the selection of several tuning parameters in the original STEEL. We conduct a simulation study to demonstrate its superior performance.

The rest of the paper is organized as follows. 
In Section \ref{sec: prelim}, we introduce our setup, the discrete-time homogeneous MDP with continuous states and actions. 
We introduce related notations and also the problem formulation. 
Then, an illustrative example on the contextual bandit problem is presented in Section \ref{sec: contextual bandit prem} for describing the challenge of policy learning under singularity. 
We also briefly introduce our solution in this section. 
In Section \ref{sec: method}, we formally introduce STEEL for finding an optimal policy and its adaptive version. 
A comprehensive theoretical study of our algorithm is given in Section \ref{sec: Theory}. 
Section \ref{sec: numerical} demonstrates the empirical performance of our methods using both simulation studies and a (semi-)real data application. 
Section \ref{sec: conclusion} concludes this paper. In the appendix, we discuss our work related to the existing literature, present some additional discussions on the theoretical results and assumptions, show all technical proofs and provide more numerical results.

%
%
%

\section{Preliminaries and Notations}\label{sec: prelim}
In this section, we briefly introduce a framework of the discrete-time  homogeneous MDP, which is the foundation of many RL algorithms, and the related notations \citep{sutton1998introduction}. 
Consider a trajectory $\{S_t, A_t, R_t   \}_{t \geq 0}$, where $S_t$ denotes the state of the environment at the decision point $t$, $A_t$ is the action and $R_t$ is the immediate reward received from the environment. We use $\calS$ and $\calA$ to denote the state and action spaces, respectively. Throughout this paper, we assume both $\calS \subseteq \mathbb R^{d_S}$ and $\calA\subseteq \mathbb R^{d_A}$ are continuous and multi-dimensional. Specifically, let $\calA = \bigotimes_{i = 1}^{d_A} \calA_i$ with $d_A \geq 1$, where $\calA_i$ is the space for $i$-th coordinate of the action.  
Under the setup of MDP, the following standard assumptions are imposed on the trajectory $\{S_t, A_t, R_t   \}_{t \geq 0}$.
\begin{assumption}\label{ass: Markovian}
	There exists a time-invariant transition kernel $P$ such that for every $t\geq 0$, $s \in \calS$, $a \in \calA$ and any set $F \in \calB(\calS)$,
	$$
	\Pr(S_{t+1}  \in F \given S_t = s, A_t = a, \left\{S_j, A_j, R_j\right\}_{0 \leq j < t}) = P(S_{t+1}  \in F \given S_t = s, A_t = a),
	$$
	where $\calB(\calS)$ is the family of Borel subsets of $\calS$ and $\left\{S_j, A_j, R_j\right\}_{0 \leq j < t} = \emptyset$ if $t = 0$. In addition, assume that for each $(s, a) \in \calS \times \calA$, $P(\bullet \given s, a)$ is absolutely continuous with respect to the Lebeque measure and thus there exists a probability density function $q(\bullet \given s, a)$ associated with $P(\bullet \given s, a)$.
\end{assumption}
\begin{assumption}\label{ass: reward}
	The immediate reward $R_t$ is a measurable function of $(S_t, A_t, S_{t+1})$, i.e., $R_t = \widetilde{R}(S_t, A_t, S_{t+1})$ for any $t \geq 0$, where $\widetilde{R}: \mathbb{R}^{2d_S+d_A} \rightarrow \mathbb{R}$. In addition, we assume $R_t$ is uniformly bounded, i.e., there exists a constant $R_{\max}$ such that $\abs{R_t} \leq R_{\max}$ for every $t \geq 0$.
\end{assumption}
By Assumption \ref{ass: reward}, we define a reward function as $r(s, a) = \EE[R_t \given S_t = s, A_t = a]$ for $t \geq 0$. The uniformly bounded assumption on the immediate reward $R_t$ is used to simplify the technical proofs and can be relaxed. We remark that one can implement hypothesis testings to justify the use of a time-homogeneous MDP to model the batch data \citep{shi2020does,li2022testing}.

An essential goal of RL is to find an optimal policy that maximizes some utility. In this paper, we consider expected discounted sum of rewards to measure the performance of each policy. A policy is defined as a decision rule that an agent chooses her action $A_t$ based on the environment state $S_t$ at each decision point $t$. 
In this paper, we focus on the stationary policy $\pi$, which is a vector-valued function mapping from the state space $\calS$ into a probability distribution over the action space $\calA$. Specifically, $\pi = \bigotimes_{i = 1}^{d_A} \pi_i$, where $\pi_i$ is a function mapping from $\calS$ into a distribution over $\calA_i$.
For each policy $\pi$, one can define a $Q$-function to measure its performance starting from any  state-action pair $(s, a) \in \calS \times \calA$, denoted by
\begin{align}\label{def: Q-function}
	Q^\pi(s, a) = \EE^\pi\left[\sum_{t=0}^{\infty}\gamma^t R_t \given S_0 = s, A_0 = a\right],
\end{align}
where $\EE^\pi$ refers to the expectation that all actions along the trajectory follow the stationary policy $\pi$. 
We evaluate the overall performance of a policy by the so-called \textit{policy value} defined as
\begin{align}\label{def: policy value}
	\calV(\pi) \triangleq (1-\gamma) \EE^\pi_{S_0 \sim \nu}\left[\sum_{t=0}^{\infty}\gamma^t R_t\right] = (1-\gamma) \EE_{S_0 \sim \nu}[Q^\pi(S_0, \pi(S_0))],
\end{align}
where $\nu$ is some \textit{known} reference distribution. 
Here, for any function $Q$ defined over $\calS \times \calA$, we define $Q(s, \pi(s)) = \int_{a \in \calA}Q(s, a)\pi(a \given s)da$. Our goal in this paper is to search for an \textit{in-class} optimal  policy $\pi^\ast$ such that
\begin{align}\label{def: optimal policy}
	\pi^\ast \in \underset{\pi \in \Pi}{\argmax} \, \calV(\pi),
\end{align}
where $\Pi = \bigotimes_{i = 1}^{d_A} \Pi_i$ and $\Pi_i$ is some pre-specified class of stationary policies for $i$-th coordinate of $\calA$. 
Some commonly used policy classes include linear decision functions, neural networks and decision trees. 
The sufficiency of focusing on the stationary policies is guaranteed by Assumptions \ref{ass: Markovian} and \ref{ass: reward}. 
See Section 6.2 of \cite{puterman1994markov} for the justification. 
We also remark that stationary policy does not rule out the deterministic policy, which is a function mapping from the state space $\calS$ into the action space $\calA$. 

In this work, we focus on the batch setting, where the observed data consist of $N$ independent and identically distributed copies of $\{S_t, A_t, R_t\}_{t \geq 0}$ up to $T$ decision points. 
For the $i$-th trajectory, where $1 \leq i \leq N$, the data can be represented by $\{S_{i, t}, A_{i, t}, R_{i, t}, S_{i, t+1}\}_{0 \leq t < T}$. 
We aim to leverage the batch data to find an in-class optimal policy $\pi^\ast$ defined in \eqref{def: optimal policy}. 
The following standard assumption is imposed on our batch data generating process. 

\begin{assumption}\label{ass: DGP}
	The batch dataset $\calD_N = \{(S_{i,t},A_{i,t},R_{i,t},S_{i,t+1})\}_{0\le t< T, 1 \le i \le N}$ is generated by a stationary policy $\pi^b$. 
\end{assumption}
Next, we introduce the average visitation probability measure. 
Let $q^{\pi^b}_t$ be the marginal probability measure over $\calS \times \calA$ at the decision point $t$ induced by the behavior policy $\pi^b$. Then the average visitation probability measure across $T$ decision points is defined as
$$
\bar d^{\pi^b}_T = \frac{1}{T}\sum_{t = 0}^{T-1}q^{\pi^b}_t. 
$$
The corresponding expectation with respect to $\bar d^{\pi^b}_T$ is denoted by $\overline \EE$. 
Similarly, we can define the discounted visitation probability measure over $\calS \times \calA$ induced by a policy $\pi$ as
\begin{align}\label{def: discounted visitation probability}
	d_\gamma^\pi = (1-\gamma) \sum_{t=0}^{\infty} \gamma^t q_t^\pi,
\end{align}
where $q_t^\pi$ is the marginal probability measure of $(S_t, A_t)$ induced by the policy $\pi$ with the initial state distribution $\nu$.  For notation simplicity, we also use $\bar d^{\pi^b}_T$ and $d_\gamma^\pi$ to denote their corresponding probability density function when there is no confusion.  Lastly, we use $\mathbb{Q} \ll \mathbb{P}$ when $\mathbb{Q}$ is absolutely continuous with respect to $\mathbb{P}$, and $\mathbb{Q} \perp \mathbb{P}$ when $\mathbb{Q}$ is singular to $\mathbb{P}$. More specifically, $\mathbb{Q} \ll \mathbb{P}$ means that if for any measurable set $\calF$ such that $\mathbb{P}(\calF) = 0$, then $\mathbb{Q}(\calF) = 0$ as well. Lastly $\mathbb{Q} \perp \mathbb{P}$ means that there exists a measurable set $\calF$ such that $\mathbb{Q}(\calF) = \mathbb{P}(\calF^c) = 0$.

\textit{Additional notations}: For generic sequences $\{\varpi(N)\}$ and $\{\theta(N)\}$, the notation $\varpi(N) \gtrsim  \theta(N)$ (resp. $\varpi(N) \lesssim \theta(N)$) means that there exists a sufficiently large constant (resp. small) constant $c_1>0$ (resp. $c_2>0$) such that $\varpi(N) \geq c_1 \theta(N)$ (resp. $\varpi(N) \leq c_2 \theta(N)$). We use $\varpi(N) \asymp \theta(N)$ when $\varpi(N) \gtrsim \theta(N)$ and $\varpi(N) \lesssim \theta(N)$. For matrix and vector norms, we use $\norm{\bullet}_{\ell_q}$ to denote  either  the vector $\ell_q$-norm or operator norm induced by the vector $\ell_q$-norm, for $1 \leq q < \infty$, when there is no confusion. For any random variable $X$, we use $L^q_{\mathbb{P}}(X)$ to denote the class of all measurable functions with finite $q$-th moments for $1 \leq q \leq \infty$, where the underlying probability measure is $\mathbb{P}$. We may also write it as $L^q_{\mathbb{P}}$ when there is no confusion about the underlying random vectors. Then the $L^q$-norm is denoted by $\norm{\bullet}_{L^q_{ \mathbb{P}}(X)}$. 
We use $\norm{\bullet}_\infty$ to denote the point-wise sup-norm for a vector-valued function. For example, when $\pi$ is a deterministic policy, $\norm{\pi}_{\infty}$ returns a vector with sup-norm on each coordinate for $\pi \in \Pi$. In addition, we often use $(S, A, R, S')$ or $(S, A, S')$ to represent a generic transition tuple.  
Lastly, we define the maximum mean discrepancy (MMD) between two probability distributions $\mathbb{P}$ and $\mathbb{Q}$ as
\begin{align}
	\text{MMD}_k(\mathbb{P}, \mathbb{Q}) = \sup_{f \in \calH_k, \norm{f}_{\calH_k} \leq 1} \left\abs{\EE_{X \sim \mathbb{P}}[f(X)] - \EE_{X \sim \mathbb{Q}}[f(X)]\right}, 
\end{align}
where $\calH_k$ is a reproducing kernel Hilbert space (RKHS) with the kernel $k$ and the corresponding norm is denoted by $\norm{\bullet}_{\calH_k}$.

{\section{An Illustrative Example: A Contextual Bandit under Singularity}\label{sec: contextual bandit prem}

In this section, we use the offline contextual bandit problem \citep{manski2004statistical} as an example to illustrate the challenge of policy learning under singularity and  our main idea to address it. 
Note that the contextual bandit problem is a special case of batch RL (i.e., $T = 0$). 

Suppose we have a batch dataset $\calD^0_N = \{S_{i, 0}, A_{i, 0}, R_{i, 0}\}_{1 \leq i \leq N}$ which contains i.i.d. copies from $(S_0, A_0, R_0)$, where $S_0 \sim \nu_0$ and $\nu_0$ is some unknown distribution. 
The target is to find an  in-class optimal policy $\pi^\ast$ such that
\begin{align}\label{def: optimal in contextual bandit}
	\pi^\ast \in \argmax_{\pi \in \Pi} \left\{\calV_0(\pi) \triangleq \EE_{S_0 \sim \nu}\left[r(S_0, \pi(S_0))\right]\right\},
\end{align}
Recall that $r(s, a) = \EE\left[R_0 \given S_0 = s, A_0 =a\right]$. 

Most existing methods for solving \eqref{def: optimal in contextual bandit} rely on a key assumption that $\nu \times \pi \ll \nu_0 \times \pi^b$ for all $\pi \in \Pi$, under which one is able to use the batch data $\calD^0_N$ to calibrate $\pi$ and evaluate its performance via estimating $\calV_0(\pi)$ for all $\pi \in \Pi$.  For instance, the regression-based approaches   \citep{qian2011performance,bhattacharya2012inferring} consider  a discrete action space, and require (i) $\pi^b(a \given s)$ is uniformly bounded away from 0 for all $s$ and $a$, and (ii) $\nu = \nu_0$. 
In this case, the absolute continuity holds. Meanwhile, the popular classification-based approaches \citep[e.g.,][]{zhao2012estimating}, which crucially rely on the inverse propensity weighting formulation, also have the same requirement. However, due to the distributional shift induced by $\pi$, such absolutely continuous assumption may not hold in general. There is a recent streamline of research studying contextual bandits under the framework of distributionally robust optimization such as \cite{mo2021learning,qi2022robustness,adjaho2022externally,si2023distributionally} and the reference therein. These works investigate policy learning in the presence of distributional shifts in the covariates or rewards when deploying the policy in the future. In particular, \cite{adjaho2022externally} considered to use Wasserstein distance for quantifying the distributional shift and can allow the singularity for such a shift. However, none of them study the policy learning when there is a singularity issue in terms of the policy during the training procedure. 
For example, when $\pi^b$ refers to some stochastic policy used to collect the data and $\Pi$ is a class of deterministic policies, $\pi$ becomes singular to $\pi^b$, i.e., $\pi \perp \pi^b$ for all $\pi \in \Pi$. In this case, most existing methods may fail to find a desirable policy. 

To address such singularity issue induced by the deterministic policy with respect to the stochastic one, existing solutions either adopt the kernel smoothing techniques on $\pi$ in order to approximately estimate $\calV_0(\pi)$ \citep[e.g.,][]{chen2016personalized} or impose some (parametric) structure assumption on the reward function $r$ \citep[e.g.,][]{chernozhukov2019semi}. The first type of approaches requires the selection of kernel and the tuning on the bandwidth for approximating all deterministic policies, while the latter one could suffer from the model mis-specification due to the strong (parametric) model assumption. It is also well-known that the performance of the kernel smoothing deteriorates when the dimension of the action space increases. In addition, these methods cannot handle general settings such as that $\nu$ is also singular to $\nu_0$, i.e., there exists a \textit{covariate shift} problem. 

In the following, we briefly introduce our method for policy learning without requiring absolute continuity of the target distribution (i.e., $\nu \times \pi$) with respect to the data distribution (i.e., $\nu_0 \times \pi^b$) in this contextual bandit problem. 
The idea relies on the following observation. Consider a fix policy $\pi \in \Pi$. Let $\widetilde{r}$ be any estimator for the reward function. It can be easily seen that
\begin{align}\label{eqn: reward functiond decomposition}
	\calV_0(\pi) - \EE_{S_0 \sim \nu}\left[\widetilde r(S_0, \pi(S_0))\right] = \EE_{(S_0, A_0) \sim \nu\times\pi}\left[(r - \widetilde{r})(S_0, A_0) \right].
\end{align}
By leveraging Lebesgue's Decomposition Theorem, we can always represent $\nu\times\pi$ as
$$
\nu\times\pi = \widetilde \lambda_1^\pi + \widetilde \lambda_2^\pi,
$$
where $\widetilde \lambda_1^\pi \ll (\nu_0\times\pi^b)$ and $\widetilde \lambda_2^\pi \perp (\nu_0\times\pi^b)$. Since $\widetilde \lambda^\pi_1$ and $\widetilde \lambda^\pi_2$ are finite measures, we normalize them and rewrite the above decomposition as
$$
\nu\times\pi = \widetilde \lambda_1^\pi(\calS \times \calA) \lambda_1^\pi + \widetilde \lambda_2^\pi(\calS \times \calA)\lambda_2^\pi,
$$
where $\lambda^\pi_1$ and $\lambda^\pi_2$  are two probability measures. In particular, if $ \widetilde  \lambda_i^\pi(\calS \times \calA) = 0$, the corresponding $\lambda_i^\pi$ can be chosen as an arbitrary probability measure. See Theorem 5.6.2 of \cite{gray2009probability} for more details. Given this decomposition, one can further show that 
\begin{align*}
	&\calV_0(\pi) - \EE_{S_0 \sim \nu}\left[\widetilde r(S_0, \pi(S_0))\right] \\
	=& \widetilde \lambda_1^\pi(\calS \times \calA) \times \underbrace{\EE_{(S_0, A_0) \sim \nu_0\times\pi^b}\left[\frac{\lambda^\pi_1(S_0, A_0)}{(\nu_0\times\pi^b)(S_0, A_0)}(r - \widetilde{r})(S_0, A_0) \right]}_{\text{absolutely continuous part}} \\
	+ & \widetilde \lambda_2^\pi(\calS \times \calA) \times  \underbrace{\EE_{(S_0, A_0) \sim\lambda_2^\pi}\left[(r - \widetilde{r})(S_0, A_0) \right]}_{\text{singular part}}.
\end{align*}
In order to have an accurate estimation of $\calV_0(\pi)$, one needs to find an $\widetilde{r}$ such that the absolutely continuous and singular parts in the above equality are minimized.
For the absolutely continuous part, since $\lambda_1^\pi$ is unknown, define $\calW$ as some class of symmetric functions and assume $\lambda^\pi_1/(\nu_0\times\pi^b) \in \calW$, we can show that
\begin{align*}
	& \left\abs{\EE_{(S_0, A_0) \sim \nu_0\times\pi^b}\left[\frac{\lambda^\pi_1(S_0, A_0)}{(\nu_0\times\pi^b)(S_0, A_0)}(r - \widetilde{r})(S_0, A_0) \right]\right} \\
	\leq &\sup_{w \in \calW} \EE_{(S_0, A_0) \sim \nu_0\times\pi^b}\left[w(S_0, A_0)(R_0 - \widetilde{r}(S_0, A_0)) \right].
\end{align*}
Then the right-hand side of the above inequality can be properly controlled using the batch data because $(S_0, A_0)$ now follows $\nu_0 \times \pi^b$.
Essentially, one can find a good $\widetilde{r}$ that minimizes the right hand side of the above inequality so as to control the error induced by the absolutely continuous part.
To handle the singular part, we adopt the idea of distributionally robust optimization \cite[e.g., a review paper by][]{rahimian2019distributionally} with MMD. We can show that
\begin{align*}
	\left\abs{\EE_{(S_0, A_0) \sim\lambda_2^\pi}\left[(r - \widetilde{r})(S_0, A_0) \right] \right} \leq \sup_{\mathbb{P}: \text{MMD}_k(\mathbb{P}, \nu_0 \times \pi^b) \leq \text{MMD}_k(\lambda_2^\pi, \nu_0 \times \pi^b) } \left\abs{ \EE_{(S_0, A_0) \sim \mathbb{P}}\left[r(S_0, A_0) - \widetilde{r}(S_0, A_0)  \right]\right}.
\end{align*}
Summarizing the above derivations together, we can quantify the error of using $\widetilde r$ for estimating $\calV_0(\pi)$ as
\begin{align*}
	&\left\abs{\calV_0(\pi) - \EE_{S_0 \sim \nu}\left[\widetilde r(S_0, \pi(S_0))\right]\right} \\
	\leq & \widetilde \lambda_1^\pi(\calS \times \calA) \sup_{w \in \calW} \EE_{(S_0, A_0) \sim \nu_0\times\pi^b}\left[w(S_0, A_0)(R_0 - \widetilde{r}(S_0, A_0)) \right] \\
	+ & \widetilde \lambda_2^\pi(\calS \times \calA) \sup_{\mathbb{P}: \text{MMD}_k(\mathbb{P}, \nu_0 \times \pi^b) \leq \text{MMD}_k(\lambda_2^\pi, \nu_0 \times \pi^b) } \left\abs{ \EE_{(S_0, A_0) \sim \mathbb{P} }\left[r(S_0, A_0) - \widetilde{r}(S_0, A_0)  \right]\right}\\
	\leq &  \sup_{w \in \calW} \EE_{(S_0, A_0) \sim \nu_0\times\pi^b}\left[w(S_0, A_0)(R_0 - \widetilde{r}(S_0, A_0)) \right] \\
	+ & \widetilde \lambda_2^\pi(\calS \times \calA) \times \text{MMD}_k(\lambda_2^\pi, \nu_0 \times \pi^b)  \times \norm{\EE\left[R_0 - \widetilde{r}(S_0, A_0) \given S_0 = \bullet, A_0 = \bullet \right]}_{\calH_k},
\end{align*}
where the last inequality is given by Lemma \ref{lm: OPE decomposition with mean embedding} in the later section under some mild conditions.
Moreover, it can be checked that if $\widetilde{r} = r$, the right-hand side of the above inequality vanishes. 
These facts motivate us to estimate the reward function $r$ and later $\calV_0(\pi)$ via
\begin{align}\label{eqn: value estimation in contextual bandit}
	\min_{\widetilde{r}} & \left\{ \sup_{w \in \calW} \EE_{(S_0, A_0) \sim \nu_0\times\pi^b}\left[w(S_0, A_0)(R_0 - \widetilde{r}(S_0, A_0)) \right] \right.\\
	&	\left. +\widetilde \lambda_2^\pi(\calS \times \calA) \times \text{MMD}_k(\lambda_2^\pi, \nu_0 \times \pi^b)  \times \norm{\EE\left[R_0 - \widetilde{r}(S_0, A_0) \given S_0 = \bullet, A_0 = \bullet \right]}_{\calH_k}\right\}\nonumber.
\end{align}
It is remarked that the goal of analyzing or upper bounding the estimation error is to construct a valid loss function when there is a singularity issue.
Moreover, $\widetilde \lambda_2^\pi(\calS \times \calA)$ and $\text{MMD}_k(\lambda_2^\pi, \nu_0 \times \pi^b)$ are irrelevant to $\tilde r$ and could be estimated when $\pi$ is known. 
Once we are able to provide a valid estimation for $\calV_0(\pi)$, an optimization rountine can be implemented to estimate $\pi^\ast$, where we incorporate the idea of pessimism. Specifically, for each given $\widetilde{r}$, we first implement the kernel ridge regression using $\calD_N^0$ for estimating $\EE\left[R_0 - \widetilde{r} \given S_0 = \bullet, A_0 = \bullet\right]$. Denote the resulting estimator by $\widehat \EE\left[R_0 - \widetilde{r} \given S_0 = \bullet, A_0 = \bullet\right]$. Note that $\widetilde{r}$ is some candidate reward function. Then we propose to compute the optimal policy via solving the maximin problem 
\begin{align}\label{eqn: pessimistic policy learning bandit}
	\max_{\pi \in \Pi}  \min_{\widetilde{r} \in \calR}& \quad \EE_{S_0 \sim \nu}\left[\widetilde r(S_0, \pi(S_0))\right],\\
	\text{subject to} & \quad \sup_{w \in \calW} \EE_N\left[w(S_0, A_0)(R_0 - \widetilde{r}(S_0, A_0)) \right]  \leq \varepsilon_1 \nonumber \\
	& \quad \norm{\widehat \EE\left[R_0 - \widetilde{r} \given S_0 = \bullet, A_0 = \bullet\right]}_{\calH_k} \leq \varepsilon_2 \nonumber,
\end{align}
where $\calR$ is some pre-specified class of functions for modeling the true reward function $r$ and $\EE_{N}$ refers to the empirical average over the batch data $\calD^0_N$. Two constants $\varepsilon_1 >0$ and $\varepsilon_2 > 0$ are used to quantify the uncertainty for estimating $r$ and also control the degree of pessimism. We remark that the RKHS norm in \eqref{eqn: pessimistic policy learning bandit} can be computed explicitly when kernel ridge regression is used in estimating $\widehat \EE\left[R_0 - \widetilde{r} \given S_0 = \bullet, A_0 = \bullet\right]$.  Under some technical conditions, with properly chosen $\varepsilon_1$ and $\varepsilon_2$, we can show that the true reward function $r$ always belongs to the feasible set of \eqref{eqn: pessimistic policy learning bandit}. 
Therefore, by implementing the above algorithm, we search a policy that maximizes the most pessimistic estimation of its value $\calV_0(\pi)$ within the corresponding uncertainty set. 
The proposed algorithm will produce a valid policy with regret guaranteed and without assuming absolute continuity.}

\section{Policy Learning in Batch RL}\label{sec: method}

In this section, we consider policy learning in  batch RL under the possible singularity, which generalizes the previous contextual bandit example. Our proposed algorithm for obtaining $\pi^\ast$ is motivated
by the following analysis of OPE. 

\subsection{Off-policy Evaluation with Potential Singularity}
For any given policy $\pi$, the target of OPE is to use the batch data to estimate the policy value  $\calV(\pi)$ defined in \eqref{def: policy value}. Note that by the definition of the discounted visitation probability measure, we can show that
$$
\calV(\pi)  = \int_{\calS \times \calA}  r(s, a) d_\gamma^\pi(\text{d}s, \text{d}a).
$$
A direct approach to perform OPE is via estimating $Q^\pi$ defined in \eqref{def: Q-function}.
Let $\widetilde{Q}$ be any estimator for $Q^\pi$ and define the corresponding estimator for the policy value as $\widetilde{\calV}(\pi) = (1-\gamma) \EE_{S_0 \sim \nu}[\widetilde Q(S_0, \pi(S_0))]$. Then we have the following lemma to characterize the estimation error of $\widetilde{\calV}(\pi)$ to $\calV(\pi)$.
\begin{lemma}\label{lm: OPE decomposition}
	Under Assumptions \ref{ass: Markovian} and \ref{ass: reward}, we have
	\begin{align}\label{eqn: deomposition}
		\calV(\pi) - \widetilde{\calV}(\pi) = \EE_{(S, A) \sim d_\gamma^\pi} \left[ R + \gamma \widetilde{Q}(S', \pi(S')) - \widetilde{Q}(S, A) \right].
	\end{align}
\end{lemma}
Based on Lemma \ref{lm: OPE decomposition}, if $d_\gamma^\pi \ll \bar d^{\pi^b}_T$, then by change of measure and assuming that $\frac{d_\gamma^\pi}{\bar d^{\pi^b}_T} \in \calW$, which is a symmetric class of functions, we can obtain that
\begin{align*}
	\abs{\calV(\pi) - \widetilde{\calV}(\pi)} &= \left\abs{\overline \EE\left[\frac{d_\gamma^\pi(S, A)}{\bar d^{\pi^b}_T(S, A)}\left(R + \gamma \widetilde{Q}(S', \pi(S')) - \widetilde{Q}(S, A) \right)\right]\right} \\
	& \leq \sup_{w \in \calW} \overline \EE\left[w(S, A)\left(R + \gamma \widetilde{Q}(S', \pi(S')) - \widetilde{Q}(S, A) \right)\right].
\end{align*}
This naturally motivates a minimax estimating approach to learn $Q^\pi$ and hence $\calV(\pi)$, i.e.,
$$
\min_{\widetilde Q}\sup_{w \in \calW} \overline \EE\left[w(S, A)\left(R + \gamma \widetilde{Q}(S', \pi(S')) - \widetilde{Q}(S, A) \right)\right],
$$
which has been used in the literature of OPE \citep[e.g.,][]{jiang2020minimax}. 

However, as discussed before, due to the potentially large state-action space, it is quite usual that some state-action pairs induced by the target policy $\pi$ are not covered by the batch data generating process, causing the issue of non-overlapping region. In addition, there are many applications where $\pi \in \Pi$ is a deterministic policy but the behavior one is stochastic \citep[e.g.,][]{silver2014deterministic,lillicrap2015continuous}. In either of the two cases,  $d_\gamma^\pi \ll \bar d^{\pi^b}_T$ fails to hold and singularity arises. Then the above min-max estimating approach may no longer be valid for OPE.  To address this issue, with some abuse of notations, we first decompose
$$
d_\gamma^\pi = \widetilde \lambda_1^\pi(\calS \times \calA) \lambda_1^\pi + \widetilde \lambda_2^\pi(\calS \times \calA)\lambda_2^\pi,
$$
where $\lambda^\pi_1$ and $\lambda^\pi_2$  are two probability measures, and $ \widetilde \lambda_i^\pi(\calS \times \calA) >0$ with $\widetilde \lambda_1^\pi(\calS \times \calA) + \widetilde \lambda_2^\pi(\calS \times \calA) = 1$.  Then we can decompose the OPE error as
\begin{align*}
	\calV(\pi) - \widetilde{\calV}(\pi) &=\widetilde \lambda_1^\pi(\calS \times \calA)  \underbrace{\EE_{(S, A) \sim \lambda_1^\pi} \left[ R + \gamma \widetilde{Q}(S', \pi(S')) - \widetilde{Q}(S, A) \right]}_{\text{absolute continuous part}}\\
	& + \widetilde \lambda_2^\pi(\calS \times \calA) \underbrace{\EE_{(S, A) \sim \lambda_2^\pi} \left[ R + \gamma \widetilde{Q}(S', \pi(S')) - \widetilde{Q}(S, A) \right]}_{\text{singular part}}.
\end{align*}
The absolutely continuous part can be properly controlled by using the above min-max formulation. However, the singularity of $\lambda_2^\pi$ with respect to $\bar d^{\pi^b}_T$ is the major obstacle that makes most existing OPE approaches not applicable.  To address this issue, one must rely on the extrapolation ability of $Q^\pi$. As the difficulty of OPE largely comes from the distributional mismatch between $d_\gamma^\pi$ and $\bar d^{\pi^b}_T$, we leverage the kernel mean embedding approach to quantifying the difference between $\lambda_2^\pi$ and $\bar d^{\pi^b}_T$ in order to control the singular part. See Lemma \ref{lm: existence of embeddings} in the appendix for the existence of kernel mean embeddings, which follows from Lemma 3 of \cite{gretton2012kernel}.

Next, we present our formal result in bounding the OPE error of $\widetilde \calV(\pi)$.  Define the Bellman residual operator as
$$
\calT^\pi\widetilde{Q}= \calB^\pi{\widetilde Q}- \widetilde Q,
$$
where $\calB^\pi$ is the Bellman operator, i.e., 
$$
\calB^\pi\widetilde Q(\bullet, \bullet) = \EE[R + \gamma \widetilde{Q}(S', \pi(S'))  \given S = \bullet, A = \bullet]
$$
for any transition tuple $(S, A, R, S')$. Since $\lambda_1^\pi \ll \bar{d}^{\pi^b}_T$, 
by the Radon-Nikodym Theorem, 
we can define the Radon-Nikodym derivative as
$$
\omega^\pi(s, a) = \frac{\lambda_1^\pi(s, a)}{\bar d^{\pi^b}_T(s, a)}, 
$$
for every $(s, a) \in \calS \times \calA$. 
We have the following key lemma for our proposal. For notation simplicity, let $Z = (S, A)$ and denote $\calZ = \calS \times \calA$.
\begin{lemma}\label{lm: OPE decomposition with mean embedding}
	For any policy $\pi$, assume that the kernel $k(\bullet, \bullet): \calZ \times \calZ \rightarrow \mathbb{R}$ is measurable with respect to  both $\lambda_2^\pi$ and $\bar d^{\pi^b}_T$, and $\max\{\overline{\EE}[\sqrt{k(Z, Z)}], \EE_{Z \sim \lambda_2^\pi}[\sqrt{k(Z, Z)}] \} < +\infty$. In addition, suppose $\{\pm 1\} \in \calW$ and $\calT^\pi \widetilde{Q} \in \calH_k$. Then the following holds.
	\begin{align}\label{eqn: decomposition inequality}
		\abs{\calV(\pi) - \widetilde{\calV}(\pi)} \leq & 
 \sup_{w \in \calW} \overline \EE\left[w(S, A)\left(R + \gamma \widetilde{Q}(S', \pi(S')) - \widetilde{Q}(S, A) \right)\right] \nonumber\\
		+ &  \widetilde \lambda_2^\pi(\calS \times \calA)  \times \text{MMD}_k(\bar d^{\pi^b}_T, \lambda_2^\pi) \times  \norm{\calT^\pi \widetilde{Q}}_{\calH_k}.
	\end{align}
\end{lemma}
Lemma \ref{lm: OPE decomposition with mean embedding} provides an upper bound for the estimation error of OPE using any estimator $\widetilde{Q}$ for $Q^\pi$. 
It can be further checked that if $\widetilde{Q} = Q^\pi$, by Bellman equation, the RHS of \eqref{eqn: decomposition inequality} becomes $0$, which demonstrates the validity of using it as an objective function to find $Q^\pi$. Since the first term on the right-hand-side of Equation \eqref{eqn: decomposition inequality} can be calibrated by the data distribution due to absolute continuity, under some mild conditions, one can expect that it can be estimated efficiently. In contrast, the second term becomes hard to estimate, whose convergence rate is typically slower than the first one if consistent. Nevertheless, based on the above analysis, we propose to estimating $Q^\pi$ by solving the following minimax problem. Without loss of generality, assume that $\pm 1 \in \calW$. 
\begin{align}\label{eqn: Q-estimation old}
	\min_{\widetilde{Q}} \sup_{w \in \calW} &\quad 
	\overline\EE\left[w(S, A)\left(R + \gamma \widetilde{Q}(S', \pi(S')) - \widetilde{Q}(S, A) \right)\right] \\
	& \quad + \widetilde \lambda_2^\pi(\calS \times \calA)  \times \text{MMD}_k(\bar d^{\pi^b}_T, \lambda_2^\pi) \times  \norm{\calT^\pi \widetilde{Q}}_{\calH_k}.\nonumber
\end{align}
To use the above procedure for OPE, for each policy $\pi$, one needs to estimate $\widetilde \lambda_1^\pi(\calS \times \calA), \widetilde \lambda_2^\pi(\calS \times \calA)$ and $\text{MMD}_k(\bar d^{\pi^b}_T, \lambda_2^\pi)$ first or find some constants that serve as upper bounds. This new method will serve as the foundation of our proposed algorithm developed below.

\subsection{Policy Optimization under Potential Singularity}
Recall that our goal is to leverage the batch data $\calD_N$ to estimate the in-class optimal  policy $\pi^\ast$. For any policy $\pi$, we can implement the empirical version of \eqref{eqn: Q-estimation old} for obtaining the estimation of $Q^\pi$ and hence perform the policy evaluation. 
This motivates us to develop a pessimistic policy iteration algorithm for finding $\pi^\ast$. 

Pessimism in  batch RL serves as the main tool for efficient policy optimization by quantifying the uncertainty of the estimation and discouraging the exploration of the learned policy from visiting the less explored state-action pair in the batch data. The success of the pessimsitic-typed algorithms has been demonstrated in many applications \citep[e.g.,][]{kumar2019stabilizing,bai2022pessimistic}. In terms of the data coverage, instead of the full-coverage assumption (i.e., $\bar{d}^{\pi^b}_T$ is uniformly bounded away from $0$) required by many classic RL algorithms, algorithms with a proper degree of pessimism only require that the  in-class optimal policy $\pi^\ast$ is covered by the behavior one, which is thus more desirable. Adapting the pessimism idea to our setting without assuming absolute continuity, we expect to develop an algorithm that finds the optimal policy by at most requiring the data coverage assumption such that $\omega^{\pi^\ast} \in \calW$. This requirement is much weaker than those needed by all existing RL algorithms.

Let $\calF$ be some pre-specified class of functions over $\calS \times \calA$, which is used to model $Q^\pi$ with $\pi \in \Pi$. 
A key building block of our proposed algorithm is to construct the following two uncertainty sets for $Q^\pi$ 
\begin{align}\label{def: feasible set of Q}
	\Omega_1(\pi, \calF, \calW, \varepsilon^{(1)}_{NT}) &= \left\{Q \in \calF \, \mid \,   \sup_{w \in \calW}  \overline\EE_{NT}\left[w(S, A)\left(R + \gamma {Q}(S', \pi(S')) - {Q}(S, A) \right)\right]  \leq \varepsilon^{(1)}_{NT}    \right\}, \\
	\Omega_2(\pi, \calF, \varepsilon^{(2)}_{NT}) &= \left\{Q \in \calF \, \mid \,  \norm{\widehat \calT^\pi {Q}}_{\calH_k} \leq \varepsilon^{(2)}_{NT}    \right\}, \nonumber
\end{align}
where  $\varepsilon^{(1)}_{NT} > 0$ and $\varepsilon^{(2)}_{NT} > 0$ are some tuning parameters possibly depending on $N$ and $T$, which will be specified later. Here
\begin{align*}
	\overline \EE_{NT}\left[f(S, A, R, S')\right]=  \frac{1}{NT}\sum_{i=1}^{N}\sum_{t = 0}^T f(S_{i, t}, A_{i, t}, R_{i, t}, S_{i, t+1}) 
\end{align*}
for any function $f$, and $\widehat{\calT}^\pi$ is an estimator of $\calT^\pi$ to be specified later.

Based on the defined uncertainty  sets $\Omega_j, j = 1, 2$, we propose to estimate the in-class optimal policy $\pi^\ast$ via
\begin{align}\label{eqn: policy optimization algorithm}
	\max_{\pi \in \Pi} \min_{Q \in \Omega_1(\pi, \calF, \calW, \varepsilon^{(1)}_{NT}) \cap \Omega_2(\pi, \calF,  \varepsilon^{(2)}_{NT}) } (1-\gamma)\EE_{S_0 \sim \nu}\left[Q(S_0, \pi(S_0))\right],
\end{align}
i.e., maximizing the worst-case policy value within the uncertainty sets. Note that our optimization problem \eqref{eqn: policy optimization algorithm} is \textit{free} of $\widetilde \lambda_1^\pi(\calS \times \calA)$, $\widetilde \lambda_2^\pi(\calS \times \calA)$ and $ \text{MMD}_k(\bar d^{\pi^b}_T, \lambda_2^\pi)$, compared with OPE using \eqref{eqn: Q-estimation old}.  Denote the resulting estimated policy as $\widehat{\pi}$. In Section \ref{sec: Theory}, we provide a regret guarantee for $\widehat{\pi}$ in finding $\pi^\ast$ under minimal data coverage assumption. The policy optimization problem \eqref{eqn: policy optimization algorithm}, in its original form, may be difficult to solve because of the constraint set for $Q$, especially when $\calF$ and $\calW$ are highly complex such as neural networks. In the following, we propose to solve it via the dual formulation of the inner minimization problem in \eqref{eqn: policy optimization algorithm}.

\subsection{Dual Problem and Algorithm}


Let $\rho = (\rho_1, \rho_2)$ be dual variables (i.e., Lagrange multipliers). Define a Lagrangian function as
\begin{align}\label{def: Lagrangian function}
	L_{NT}(Q, \rho, \pi) &= (1-\gamma)\EE_{S_0 \sim \nu}\left[Q(S_0, \pi(S_0))\right]\\
	& +\rho_1 \times \left\{\sup_{w \in \calW}  \overline\EE_{NT}\left[w(S, A)\left(R + \gamma {Q}(S', \pi(S')) - {Q}(S, A) \right)\right]  - \varepsilon^{(1)}_{NT}\right\}\nonumber \\ 
	& + \rho_2 \times \left\{\norm{\widehat \calT^\pi {Q}}_{\calH_k} - \varepsilon^{(2)}_{NT} \right\}\nonumber.
\end{align}
Based on the formulation of $L_{NT}(Q, \rho, \pi) $, 
we consider an alternative way to estimating the optimal policy by solving the equivalent optimization problem 
\begin{align}\label{eqn: dual policy optimization}
	\max_{\pi \in \Pi, \rho \succeq 0}\min_{Q \in \calF} L_{NT}(Q, \rho, \pi),
\end{align}
where $\succeq$ refers to the component-wise comparison. 
Note that compared with \eqref{eqn: policy optimization algorithm}, Problem \eqref{eqn: dual policy optimization} can be solved in a more efficient manner as the optimization with respect to the two dual variables can be easily performed. 

Denote the resulting estimated policy given by \eqref{eqn: dual policy optimization}  as $\widehat{\pi}_{\text{dual}}$. By the weak duality, it can always
be shown that
\begin{align*}
	\max_{\rho \succeq 0}\min_{Q \in \calF} L_{NT}(Q, \rho, \pi) \leq \min_{Q \in \Omega_1(\pi, \calF, \calW, \varepsilon^{(1)}_{NT}) \cap \Omega_2(\pi, \calF,  \varepsilon^{(2)}_{NT}) } (1-\gamma)\EE_{S_0 \sim \nu}\left[Q(S_0, \pi(S_0))\right].
\end{align*}
Therefore, solving Problem \eqref{eqn: dual policy optimization} can also be viewed as a policy optimization algorithm with pessimism. However, the strong duality for the primal and dual problems may not hold as the primal problem is not convex with respect to $Q$, which therefore leads to that $\widehat{\pi} \neq \widehat{\pi}_{\text{dual}}$ in general. More seriously, the existence of the duality gap will result in a non-negligible regret of $\widehat{\pi}_{\text{dual}}$ in finding $\pi^\ast$. Nevertheless, in Section \ref{sec: Theory} below, we show that under one additional mild assumption, $\widehat{\pi}_{\text{dual}}$ indeed can achieve the same regret guarantee as that of $\widehat{\pi}$.

Now, to solve  \eqref{eqn: dual policy optimization}, we first estimate $\calT$ via kernel ridge regression. Specifically, 
\begin{align}\label{eqn: krr}
	\widehat{\calT}^\pi Q \in \argmin_{f \in \calH_k} \, \, \overline\EE_{NT}\left[\left(R + \gamma Q(S', \pi(S')) - Q(S, A) - f(S, A) \right)^2\right] + \zeta_{NT} \norm{f}^2_{\calH_k},
\end{align}
with the regularization parameter $\zeta_{NT} > 0$. Denote
\begin{align*}
	Y(\pi, Q) & = \left(Y_{1, 0}(\pi, Q), \cdots, Y_{1, T-1}(\pi, Q), Y_{2, 0}, \cdots, Y_{N, T-1} (\pi, Q)\right) \in \mathbb{R}^{NT} \quad \text{with}\\
	Y_{i, t}(\pi, Q)  & = R_{i, t} + \gamma {Q}(S_{i, t+1}, \pi(S_{i, t+1})) - {Q}(S_{i, t}, A_{i, t}), \quad \text{and let}\\
	K & \in \mathbb{R}^{NT \times NT} \quad \text{with} \quad K_{i, j} = k(Z_{\floor{i/T}+1, i \, \text{mod} \, T - 1}, Z_{\floor{j/T}+1, j \, \text{mod} \, T - 1}).
\end{align*}
Thanks to the representer theorem, we can show that
\begin{align}\label{eqn: RKHS solution}
	\norm{\widehat{\calT}^\pi Q}^2_{\calH_k} = Y(\pi, Q)^\top (K + \zeta_{NT}I_{NT})^{-1}K(K + \zeta_{NT}I_{NT})^{-1}Y(\pi, Q),
\end{align}
where $I_{NT}$ is an identity matrix with the dimension $NT$.

Similarly, we use a RKHS to model $\calW$. 
We set $\calW = \{w \, \mid \, \norm{w}_{\calH_k} \leq C \}$ for some constant $C>0$. Based on the discussion after Lemma \ref{lm: OPE decomposition with mean embedding}, ideally $C =\text{MMD}_k(\bar d^{\pi^b}_T,  d^{\pi^\ast}_\gamma)$, which is unknown in practice. But one can first enforce $\rho_1 = 0$ to compute a good estimated optimal policy and then approximate $\text{MMD}_k(\bar d^{\pi^b}_T,  d^{\pi^\ast}_\gamma)$ by it.
Once a suitable $C$ is determined, by the representer property, we can obtain a closed-form optimal value for 
$$
\max_{w \in \calW}\overline\EE_{NT}\left[w(S, A)\left(R + \gamma {Q}(S', \pi(S')) - {Q}(S, A) \right)\right],  
$$
which is given as
$$
\frac{C}{NT} \sqrt{Y(\pi, Q)^\top K Y(\pi, Q)}.
$$

With these two choices, \eqref{eqn: dual policy optimization} is equivalent to 
\begin{align}\label{eqn: primal-dual}
	\max_{\pi \in \Pi, \rho \succeq 0} \min_{Q\in \calF} &  \left\{ \left(1-\gamma\right) \EE_{S_0 \sim \nu}\left[Q(S_0, \pi(S_0))\right]\right. \\
	&\left. +\rho_1 \times \left[\frac{1}{(NT)^2} Y(\pi, Q)^\top K Y(\pi, Q) - (\varepsilon^{(1)}_{NT})^2\right]\nonumber \right. \\ 
	&\left. + \rho_2 \times \left[Y(\pi, Q)^\top (K + \zeta_{NT}I_{NT})^{-1}K(K + \zeta_{NT}I_{NT})^{-1}Y(\pi, Q)- (\varepsilon^{(2)}_{NT})^2 \right]
	\right\}\nonumber.
\end{align}
Lastly, if letting $\calF$ and $\Pi$ be any pre-specified functional classes such as neural network architectures and RKHS, we can apply stochastic gradient decent to solve the primal-dual problem \eqref{eqn: primal-dual} and obtain $\widehat \pi_{\text{dual}}$. 
See a pseudo algorithm of STEEL in Algorithm \ref{alg:MBTS-general}. 


\SetKwInput{KwData}{Input}
\SetKwInput{KwData}{Input}

\newcommand\mycommfont[1]{\footnotesize\ttfamily\textcolor{blue}{#1}}
\SetCommentSty{mycommfont}

\begin{algorithm}[!ht]
\setcounter{AlgoLine}{1}
\KwData{Data $\calD_N = \{(S_{i,t},A_{i,t},R_{i,t},S_{i,t+1})\}_{0\le t< T, 1 \le i \le N}$, kernel function $k(\cdot, \cdot)$,  discount factor $\gamma$, initial state distribution $\nu$, function class $\calF$ and $\Pi$, tuning parameters $\zeta_{NT}$, $\varepsilon^{(1)}_{NT}$, and $\varepsilon^{(2)}_{NT}$, and learning rates $(lr_{\rho, 1}, lr_{\rho, 2}, lr_{\pi}, lr_{Q})$. 
}



Initialize the values of $\rho = (\rho_1, \rho_2)$, $Q$, and $\pi$

Compute the pairwise kernel matrix 
\begin{align*}
	K \in \mathbb{R}^{NT \times NT} \quad \text{with} \quad K_{i, j} = k(Z_{\floor{i/T}+1, i \, \text{mod} \, T - 1}, Z_{\floor{j/T}+1, j \, \text{mod} \, T - 1}).
\end{align*}

\Repeat{convergence}{

Compute the objective function \eqref{eqn: primal-dual} and take a few 
gradient descent steps with respect to $Q$ with learning rate $lr_{Q}$ 

Take a 
gradient ascent step  with respect to $\rho_1$
\begin{align*}
    \rho_1 = \rho_1 + lr_{\rho, 1} \times \left\{\frac{1}{(NT)^2} Y(\pi, 
Q)^\top K Y(\pi, Q) - (\varepsilon^{(1)}_{NT})^2\right\}
\end{align*}

Take a 
gradient ascent step with respect to $\rho_2$
\begin{align*}
\rho_2 = \rho_2 + lr_{\rho, 2} \times \left\{Y(\pi, Q)^\top (K + \zeta_{NT}I_{NT})^{-1}K(K + \zeta_{NT}I_{NT})^{-1}Y(\pi, Q)- (\varepsilon^{(2)}_{NT})^2 \right\}
\end{align*}

Compute the objective function \eqref{eqn: primal-dual} and take a 
gradient ascent step with respect to $\pi$ with learning rate $lr_{\pi}$
}


\KwResult{An estimated Policy $\widehat{\pi}$}
\caption{A pseudo algorithm of STEEL.}
\label{alg:MBTS-general}
\end{algorithm}



\subsection{Adaptive STEEL}
In this subsection, we propose an adaptive version of STEEL to finding the optimal policy under singularity. It consists of three steps. In the first step, we run STEEL by only considering the constraint set $\Omega_2(\pi, \calF, \varepsilon^{(2)}_{NT})$ with a sufficiently large $\varepsilon^{(2)}_{NT}$. Essentially, we implement the algorithm without assuming any overlapping region between the behavior policy and the optimal one. Denote the resulting estimated policy as $\widehat \pi_0$ and the estimated $Q$-function as $\widehat Q$ after solving \eqref{eqn: policy optimization algorithm} or \eqref{eqn: dual policy optimization}. In the second step, based on $\widehat \pi_0$, we aim to construct a single constraint set that takes account for both singularity and absolute continuity. Basically, we leverage the initially estimated policy $\widehat \pi_0$ to calibrate the absolutely continuous and singular parts with respect to the batch data distribution. To achieve this goal, we need to estimate several related quantities including $\widetilde \lambda_1^{\widehat \pi_0}, \widetilde \lambda_2^{\widehat \pi_0}, \lambda_1^{\widehat \pi_0}, \lambda_2^{\widehat \pi_0}, \delta^{\widehat \pi_0} = \text{MMD}(\lambda_2^{\widehat \pi_0}, \bar{d}_T^{\pi^b})$, and  $\omega^{\widehat \pi_0}$. One can first estimate the behavior policy and transition probability using the batch data. Then $\bar{d}_T^{\pi^b}$ and $d_\gamma^{\widehat \pi_0}$ can be estimated correspondingly, where the estimators are denoted by $\widehat{d}_T^{\pi^b}$ and $\widehat {d}_\gamma^{\widehat \pi_0}$ respectively. To find $\omega^{\widehat \pi_0}$, one can solve the following minimization problem:
\begin{equation}\label{eqn: estimate ac part}
 \underset{\omega \in \calW}{\text{minimize}} \int_{\calS \times \calA} \abs{d_\gamma^{\widehat \pi_0}(s, a) - \omega(s, a)\bar{d}_T^{\pi^b}(s, a)}\text{d}s\text{d}a.  
\end{equation}
Denote the optimal solution as $\widehat \omega^{\widehat \pi_0}$. Based on the relationship that $\widetilde \lambda_1^{\widehat \pi_0}\times\lambda_1^{\widehat \pi_0} = \widehat \omega^{\widehat \pi_0} \times \bar{d}_T^{\pi^b}$, we can obtain an estimator of $\widetilde \lambda_1^{\widehat \pi_0}$ denoted by $\widehat{\widetilde \lambda_1^{\widehat \pi_0}} = \int_{\calS \times \calA} \widehat \omega^{\widehat \pi_0}(s,a) \times \widehat{d}_T^{\pi^b}(s, a)$ and an estimator of $\lambda_1^{\widehat \pi_0}(s, a)$ denoted by $\widehat \lambda_1^{\widehat \pi_0}(s, a) = \widehat \omega^{\widehat \pi_0}(s,a) \times \bar{d}_T^{\pi^b}(s, a)/\widehat{\widetilde \lambda_1^{\widehat \pi_0}}$. By the Lebesgue's Decomposition, we can find $\widetilde \lambda_2^{\widehat \pi_0}\times\lambda_2^{\widehat \pi_0} = d_\gamma^{\widehat \pi} - \widehat \omega^{\widehat \pi_0}\times \bar{d}_T^{\pi^b}$. Correspondingly, we can construct their estimators denoted by $\widehat{ \widetilde \lambda_2^{\widehat \pi_0}}$ and $\widehat \lambda_2^{\widehat \pi_0}$ correspondingly. Finally, an estimator of $\delta^{\widehat \pi}$ denoted by $\widehat \delta^{\widehat \pi_0}$ can be computed by standard methods such as \citep{gretton2012kernel}. Based on the aforementioned quantities, we construct the following uncertainty set denoted by $\Omega$.
\begin{align}\label{eqn: adaptive confidence set}
    \Omega = &\{Q \in \calF \, \mid \, \widehat{\widetilde \lambda_1^{\widehat \pi_0}}(\calS, \calA) \times \EE_{(S, A) \sim \widehat \lambda_1^{\widehat \pi_0}}[R + \gamma Q(S',  \pi(S')) - Q (S, A)]\\
   &+ \widehat{\widetilde \lambda_2^{\widehat \pi_0}}(\calS, \calA) \times \EE_{NT}[R + \gamma  Q(S', \pi(S')) -  Q (S, A)] + \widehat{\widetilde \lambda_2^{\widehat \pi_0}}(\calS, \calA)\times \widehat \delta^{\widehat \pi_0}\norm{\widehat \calT^\pi { Q}}_{\calH_k}\leq \varepsilon_0\}.\nonumber,
\end{align}
where the constant $\varepsilon_0$ is computed as follows.
\begin{align}\label{eqn: adaptive varepsilon}
    \varepsilon_0 &= \widehat{\widetilde \lambda_1^{\widehat \pi_0}}(\calS, \calA) \times \EE_{(S, A) \sim \widehat \lambda_1^{\widehat \pi_0}}[R + \gamma \widehat Q(S', \widehat \pi_0(S')) - \widehat Q (S, A)]\nonumber \\
    & + \widehat{\widetilde \lambda_2^{\widehat \pi_0}}(\calS, \calA) \times \EE_{NT}[R + \gamma \widehat Q(S', \widehat \pi_0(S')) - \widehat Q (S, A)] + \widehat{\widetilde \lambda_2^{\widehat \pi_0}}(\calS, \calA)\times \widehat \delta^{\widehat \pi_0}\norm{\widehat \calT^{\widehat \pi_0} {\widehat Q}}_{\calH_k}.
\end{align}
Note that $\Omega$ does not rely on any tuning parameter. The last step is to run our proposed STEEL again with a single contraint set $\Omega$, which gives an adaptive policy $\widehat \pi_1$.  The full algorithm can be found in Table \ref{alg:STEEL-adaptive}.

\begin{algorithm}[!ht]
\setcounter{AlgoLine}{1}
\KwData{Data $\calD_N = \{(S_{i,t},A_{i,t},R_{i,t},S_{i,t+1})\}_{0\le t< T, 1 \le i \le N}$
}



Run STEEL algorithm with the constraint set $\Omega_2$ only and a sufficiently large $\varepsilon^{(2)}_{NT}$. Output $\widehat \pi_0$ and $\widehat Q$.

Estimate $\widetilde \lambda_1^{\widehat \pi_0}, \widetilde \lambda_2^{\widehat \pi_0}, \lambda_1^{\widehat \pi_0}, \lambda_2^{\widehat \pi_0}, \delta^{\widehat \pi_0}, \bar{d}_T^{\pi^b},$ and $\omega^{\widehat \pi_0}$, and compute $\varepsilon_0$ and construct $\Omega$ given in Equations  \eqref{eqn: adaptive varepsilon} and \eqref{eqn: adaptive confidence set} respectively.

Run STEEL with the constraint set $\Omega$.

\KwResult{An estimated policy $\widehat{\pi}_1$}
\caption{A pseudo algorithm of adaptive STEEL.}
\label{alg:STEEL-adaptive}
\end{algorithm}

\section{Theoretical Results}\label{sec: Theory}
The performance of a policy optimization algorithm is often measured by the difference between the value of the (in-class) optimal policy and that of the estimated one, which is referred to as the \textit{regret}. In our case, we evaluate the performance of any estimated policy $\widetilde{\pi} \in \Pi$ via the regret defined as
\begin{align}\label{def: regret}
	\text{Regret}(\widetilde{\pi}) = \calV(\pi^\ast) - \calV(\widetilde{\pi}).
\end{align}
Clearly, $\text{Regret}(\widetilde{\pi}) \geq 0$.
In this section, we aim to derive the finite-sample upper bounds for both $\text{Regret}(\widehat{\pi})$ and $\text{Regret}(\widehat{\pi}_{\text{dual}})$. We leave the theoretical analysis of adaptive STEEL as future work. To begin with, we list several important assumptions and discuss their corresponding implications. Other assumptions can be found in Appendix \ref{app: Theory}.

\subsection{Technical Assumptions}\label{subsec: technical assumptions}

\begin{assumption}\label{ass: stationary}
	The stochastic process $\{S_t, A_t\}_{t\geq 0}$ induced by the behavior policy $\pi^b$ is stationary, exponentially $\boldsymbol{\beta}$-mixing. The $\boldsymbol{\beta}$-mixing coefficient at time lag $j$ satisfies that $\beta(j) \leq \beta_0 \exp(-\beta_1 j)$ for $\beta_0 \geq 0$ and $\beta_1 > 0$. The induced stationary distribution is denoted by $d^{\pi^b}$.
\end{assumption}
Our batch data consist of multiple trajectories, where observations in each trajectory follow a MDP and thus are dependent.  Assumption \ref{ass: stationary}  characterizes the dependency among those observations, instead of assuming transition tuples are all independent as in many previous works. This assumption has been used in recent works \citep[e.g.,][]{shi2020statistical,chen2022well}. The  upper bound on the $\boldsymbol{\beta}$-mixing coefficient at time lag $j$ indicates that the dependency between $\{S_t, A_t\}_{t \leq k }$	and $\{S_t, A_t\}_{t \geq (k+j) }$ decays to 0 at least exponentially fast with respect to $j$. See \cite{bradley2005basic} for the exact definition of the exponentially $\boldsymbol{\beta}$-mixing.  Recall that we have let $Z = (S, A)$ and denoted $\calZ = \calS \times \calA$. Define an operator $T^\pi_{\zeta_{NT}}: \calF \rightarrow \calH_k$ such that
\begin{align}\label{def: bias term oracle RKHS}
	T^\pi_{\zeta_{NT}}Q = \argmin_{f \in \calH_k} \overline{\EE}\left[\left(R + \gamma Q(S', \pi(S')) - Q(S, A) - f(S, A)\right)^2\right] + \zeta_{NT} \norm{f}_{\calH_k}^2,
\end{align}
for every $Q \in \calF$. Consider the following class of vector-valued functions as
\begin{align*}
	\calG & =  \left\{g: \calS \times \calA \times \calS \rightarrow \calH_k \, \mid \,  g(s, a, s') = \left(Y(Q, \pi, s, a, s') -  \calT^\pi_{\zeta_{NT}}Q(s, a)\right)k((s, a), \bullet) \right.\\
	& \left. \, \text{with} \quad Q \in \calF, \pi \in \Pi  \right\},
\end{align*}
where $Y(Q, \pi, s, a, s') = \widetilde{R}(s, a, s') + \gamma Q(s', \pi(s')) - Q(s, a)$. For $g \in \calG$, we use $\norm{g}_{\calH_k, \infty}$ to denote $\sup_{(s, a, s') \in \calS \times \calA \times \calS}\norm{g(s, a, s')}_{\calH_k}$.   In the following, we quantify the complexities of function classes used in our policy optimization algorithm via the $\epsilon$-covering number.  An $\epsilon$-covering number of a set $\Theta$ denoted by $\calN(\epsilon, \Theta, \widetilde d)$ is the infimum of the cardinality of $\epsilon$-cover of $\Theta$ measured by a (semi)-metric $\widetilde d$. See definition of the $\epsilon$-covering number in  \ref{def: covering} of the appendix. 

\begin{assumption}\label{ass: function class}
	The following conditions hold: 
	\begin{enumerate}[(a)]
		\item \label{ass: Q-function class} For any $Q \in \calF$, $\norm{Q}_\infty \leq c_{\calF} < +\infty$. For any $Q \in \calF$, $s \in \calS$, and $a_1, a_2 \in \calA$,
		\begin{align}\label{eqn: lips on Q}
			\abs{Q(s, a_1) - Q(s, a_2)} \lesssim \norm{a_1 - a_2}_{\ell_1}.
		\end{align}
		In addition, for any $\epsilon > 0$, we have
		\begin{align}\label{eqn: entropy for Q}
			\calN(\epsilon, \calW, \norm{\bullet}_\infty) \lesssim \calN(\epsilon, \calF, \norm{\bullet}_\infty) \lesssim \left(\frac{1}{\epsilon} \right)^{v(\calF)},
		\end{align}
		where $v(\calF) > 0$ is some constant.
  \item \label{ass: policy class} The policy class $\Pi$ is a class of deterministic policies, i.e., $\pi: \calS \rightarrow \calA$. 
  The action space $\calA$ is bounded. For any $\epsilon > 0$ and $1 \leq i \leq d_A$, we have
		\begin{align}\label{eqn: entropy for pi}
			\calN(\epsilon, \Pi_i, \norm{\bullet}_\infty) \lesssim \left(\frac{1}{\epsilon} \right)^{v_i(\Pi)},
		\end{align}
		where $v_i(\Pi) > 0$ is some constant. Let $v(\Pi) = \sum_{i = 1}^{d_A} v_i(\Pi).$
		\item \label{ass: vector-valued function class} For any $g \in \calG$, $\norm{g}_{\calH_k, \infty} \leq c_{\calG}$ for some constant $c_{\calG}$. In addition, for any $\epsilon > 0$, we have
		\begin{align}\label{eqn: entropy for ratio}
			\calN(\epsilon, \calG, \norm{\bullet}_{\calH_k, \infty}) \lesssim \left(\frac{1}{\epsilon} \right)^{v(\calG)},
		\end{align}
		where $v(\calG) > 0$ is some constant.
	\end{enumerate}
\end{assumption}

Assumption \ref{ass: function class} imposes metric entropy conditions on function classes $\calF$, $\calW$, $\Pi$ and $\calG$. For simplicity, we consider uniformly bounded classes for deriving the exponential inequalities for both scalar-value and vector-value function classes \citep{van1996weak,park2022towards}. 
In Assumption \ref{ass: function class}~\eqref{ass: policy class}, we only focus on the policy class which consists of the deterministic policies because there always exists an optimal policy that is deterministic. One can generalize Assumption \ref{ass: function class} to include more complicated function classes such as neural networks. See Lemma 5 of \cite{schmidt2020nonparametric} for an example. We also remark that $v(\calF)$, $v(\Pi)$ and $v(\calG)$ could increase with respect to $N$ and $T$. In practice, one can specify $\calF$, $\Pi$ and $\calW$ and compute its corresponding complexity. In Appendix \ref{app: Theory}, we provide a sufficient condition for computing $v(\calG)$. The Lipschitz-typed condition in \eqref{eqn: lips on Q} is imposed so that the $\epsilon$-covering number of the following class of functions:
\begin{align*}
	\widetilde{\calF} = \left\{(S, A, S') \rightarrow R + \gamma Q(S', \pi(S'))-Q(S, A) \, \mid \, Q \in \calF, \pi \in \Pi    \right\}
\end{align*}
could be properly controlled by the $\epsilon$-covering numbers of $\Pi_i$ for $1 \leq i \leq d_A$ and $\calF$ \citep{van1996weak}.

Next, let $\{e_j\}_{j \geq 1}$ be a non-increasing sequence of eigenvalues of the convolution operator defined in Appendix \ref{app: Theory} towards $0$ and $\{\phi_j\}_{j \geq 1}$ be orthonormal bases of $L^2_{d^{\pi^b}_T}(\calZ)$. Define a space $\calH_k^{2c}$ as
\begin{align*}
	\calH_k^{2c} = \left\{f = \sum_{i = 1}^{\infty}\bar{e}_i \phi_i \,  \mid \, \forall i \geq 1, \,  \bar{e}_i \in \mathbb R \, \text{such that} \,  \sum_{i = 1}^{\infty} \frac{\bar{e}^2_i }{e_i^{2c}} < +\infty \right\}.
\end{align*}
We impose the following assumption.
\begin{assumption}\label{ass: regularity}
	For any $\pi \in \Pi$, any $Q \in \calF$, and some constant $c \in (1/2, 3/2]$, $\calB^\pi Q \in \calF \subseteq \calH_k^{2c}$. In addition, $\calF$ is symmetric and closed under the limit operation and addition.
\end{assumption}
The first statement of Assumption \ref{ass: regularity} is closely related to Bellman completeness, which has been widely imposed in the literature of batch RL \citep[e.g.,][]{antos_value-iteration_2007}. This assumption will typically hold when the transition and the reward functions are smooth enough. Since $\calF$ is closed under the limit operator, Assumption \ref{ass: regularity}  implies that $Q^\pi \in \calF$ for $\pi \in \Pi$ by the contraction property of Bellman operator. In other words,  there is no model misspecification error for estimating $Q^\pi$ for every $\pi \in \Pi$. Moreover, together with symmetry and closed under addition, Assumption \ref{ass: regularity} also implies that for any $\pi \in \Pi$, any $Q \in \calF$, $\calT^\pi Q \in \calF \subseteq \calH_k^{2c}$.  This further states that for any $\pi \in \Pi$, $Q \in \calF$, there exists $g_{\pi, Q} \in L^2_{d^{\pi_b}}(\calZ)$ such that $\calL_k^cg_{\pi, Q} = \calT^\pi Q$, where $L_k$ is an operator associated with the kernel $k$ and its definition can be found in Appendix \ref{app: Theory}. In the standard literature of the kernel ridge regression \citep[e.g.,][]{smale2007learning,caponnetto2007optimal}, for fixed $Q$ and $\pi$, the same type of the smoothness assumption is also imposed. $c \in (1/2, 3/2]$ is the range where the kernel ridge regression can adapt to the smoothness of the true function. 
Here we strengthen this smoothness assumption by considering for all $ \pi \in \Pi$ and $Q \in \calF$, which is used for controlling the bias uniformly over $\calF$ and $\Pi$ 
induced by the regularization in the kernel ridge regression \eqref{eqn: krr}.  See more discussion related to this in Remark 2 of Theorem \ref{thm: regret bound for primal}. It is worth mentioning that one can choose $\calF = \calH_k^{2c}$ and modify Assumption \ref{ass: function class}~\eqref{ass: Q-function class} accordingly.

\subsection{Regret Bounds}\label{subsec: regret bound}
To derive the regret bounds for $\widehat{\pi}$ and $\widehat{\pi}_{\text{dual}}$, we first show that for every $\pi \in \Pi$, $Q^\pi$ is a feasible solution to the inner minimization problem of \eqref{eqn: policy optimization algorithm} with a high probability. Let
\begin{align*}
	\Omega(\pi, \calF,  \calW, \varepsilon^{(1)}_{NT}, \varepsilon^{(2)}_{NT}) \triangleq \Omega_1(\pi, \calF, \calW, \varepsilon^{(1)}_{NT}) \cap \Omega_2(\pi, \calF,  \varepsilon^{(2)}_{NT}).
\end{align*}
\vspace{-1cm}
\begin{theorem}\label{thm: feasible set}
	Suppose Assumptions \ref{ass: Markovian}-\ref{ass: regularity}, and Assumption \ref{ass: kernel} in Appendix are satisfied, by letting
	\begin{align*}
		\varepsilon_{NT}^{(1)} &\asymp \log(NT)\sqrt{\left(v(\Pi) +  v(\calF)   \right)/NT} \quad \text{and}\\
		\varepsilon_{NT}^{(2)} &\asymp 
		\left(\log(NT)\sqrt{\left(v(\Pi) +  v(\calF) + v(\calG)  \right)/NT}\right)^{\frac{2c-1}{2c+1}}, 
	\end{align*}
	where $c \in (1/2, 3/2]$ is defined in Assumption \ref{ass: regularity},
	we have with probability at least $1- 1/NT$, for every $\pi \in \Pi$,
	\begin{align}\label{eqn: Q-function feasible}
		Q^\pi \in  \Omega(\pi, \calF,  \calW, \varepsilon^{(1)}_{NT}, \varepsilon^{(2)}_{NT}).
	\end{align}
\end{theorem}
Theorem \ref{thm: feasible set} is the key for establishing the regret bounds of our proposed algorithms. By showing \eqref{eqn: Q-function feasible}, we can guarantee that for every $\pi \in \Pi$, $Q^\pi$ is bounded below by the optimal value of the inner minimization problem in \eqref{eqn: policy optimization algorithm} with a high probability. This verifies the use of pessimism in our algorithms, i.e., evaluating the value of each $\pi \in \Pi$ by a lower bound during the policy optimization procedure. Since we require $c \in (1/2, 3/2]$, the 
smallest value that $\varepsilon^{(2)}_{NT}$ can be set is of the order $\sqrt{\log(NT)}(NT)^{-1/4}$. Based on Theorem \ref{thm: feasible set}, we have the following regret warranty for $\widehat{\pi}$.
\begin{theorem}[Regret warranty for $\widehat{\pi}$]\label{thm: regret bound for primal}
	Under Assumptions \ref{ass: Markovian}-\ref{ass: regularity} and Assumption \ref{ass: kernel} in Appendix, the following statements hold.
	\begin{enumerate}[(a)]
		\item Suppose $\omega^{\pi^\ast} \in \calW$, then with probability at least $1 - \frac{1}{NT}$,
		\begin{align}\label{eqn: regret bound for primal}
			\text{Regret}  \, (\widehat{\pi}) \lesssim \varepsilon^{(1)}_{NT} +  \widetilde \lambda^{\pi^\ast}_2(\calS \times \calA) \times \text{MMD}_k(\bar d^{\pi^b}_T, \lambda_2^{\pi^\ast}) \times \varepsilon^{(2)}_{NT},
		\end{align}
		where $\varepsilon^{(1)}_{NT}$ and $\varepsilon^{(2)}_{NT}$ are given in Theorem \ref{thm: feasible set}. 
		\item  Suppose $d^{\pi^\ast}_\gamma \ll {d}^{\pi^b}$ and $\omega^{\pi^\ast} \in \calW$, then with probability at least $1 - \frac{1}{NT}$,
		\begin{align}\label{eqn: regret bound for primal case 2}
			\text{Regret}  \, (\widehat{\pi}) \lesssim  \varepsilon^{(1)}_{NT}.
		\end{align}
		\item Suppose $d^{\pi^\ast}_\gamma \perp {d}^{\pi^b}$, then with probability at least $1 - \frac{1}{NT}$,
		\begin{align}\label{eqn: regret bound for primal case 3}
			\text{Regret}  \, (\widehat{\pi}) \lesssim \varepsilon^{(1)}_{NT}+\text{MMD}_k(\bar d^{\pi^b}_T, \lambda_2^{\pi^\ast}) \times \varepsilon^{(2)}_{NT}.
		\end{align}
	\end{enumerate}
Furthermore, when $c = 1/ 2$ defined in Assumption \ref{ass: regularity} (i.e., $\calH_k^{2c} = \calH_k$), then by letting $\varepsilon^{(2)}_{NT} \asymp 1$, all statements above hold.

\end{theorem}
\begin{remark}
	Among all assumptions we have made, Bellman completeness (i.e., Assumption \ref{ass: regularity})  and $\omega^{\pi^\ast} = \frac{\lambda_1^{\pi^\ast}}{{d}^{\pi^b}} \in \calW$ with $\norm{\omega^{\pi^\ast}}_\infty < +\infty$, are two main conditions for our regret guarantee stated in Theorem \ref{thm: regret bound for primal}. The first condition is related to the modeling assumption, while the second one is related to our batch data coverage induced by the behavior policy.
	
	In the existing literature, the weakest data coverage assumption for establishing the regret guarantee is $ d^{\pi^\ast}_\gamma \ll {d}^{\pi^b}$ with $\norm{d^{\pi^\ast}_\gamma/{d}^{\pi^b}}_\infty < + \infty$, i.e., absolute continuity with a uniformly bounded Radon–Nikodym derivative. See for example \cite{xie2021bellman} and \cite{zhan2022offline}. In contrast, our proposed method remains valid without requiring the absolute continuity. For example, in Case (a) of Theorem \ref{thm: regret bound for primal}, to achieve the regret guarantee of our proposed method, we only require the ratio of the absolutely continuous part  of $d^{\pi^\ast}_\gamma$ with respect to  ${d}^{\pi^b}$ over ${d}^{\pi^b}$, i.e., $\frac{\lambda_1^{\pi^\ast}}{{d}^{\pi^b}}$,  is uniformly bounded above, and the MMD between the singular part of $d^{\pi^\ast}_\gamma$ with respect to  ${d}^{\pi^b}$  and  ${d}^{\pi^b}$, i.e., $\text{MMD}_k(\bar d^{\pi^b}_T, \lambda_2^{\pi^\ast})$,  is finite, which is ensured by Assumption \ref{ass: kernel}. Therefore our method requires much weaker assumptions (essentially no assumptions) on the batch data coverage induced by the behavior policy, and is thus more applicable than existing algorithms from the perspective of the data coverage. However, our method requires a stronger condition on modeling $Q^\pi$, which could be regarded as a trade-off.  We require Bellman completeness and symmetric, additive functional class for modeling $Q^\pi$ (i.e., Assumption \ref{ass: regularity}), while some existing pessimistic algorithms \citep[e.g.,][]{jiang2020minimax} only require the realizability condition on modeling $Q^\pi$ (i.e., there is no misspecification error for $Q^\pi$). 
	
	It is worth mentioning that, when $d^{\pi^\ast}_\gamma \ll {d}^{\pi^b}$ as discussed in the Case (b) of Theorem \ref{thm: regret bound for primal}, $\omega^{\pi^\ast}$ becomes $d^{\pi^\ast}_\gamma/{d}^{\pi^b}$, with $ \widetilde \lambda^{\pi^\ast}_2(\calS \times \calA)  =0$. In this case, we recover the existing results such as \cite{jiang2020minimax}  on the regret bound by running our algorithm. However, compared with their results,  we require a stronger condition, i.e., Assumption \ref{ass: regularity} for modeling $Q^\pi$ instead of realizability condition on $Q^\pi$ only. The main reason for requiring a stronger condition here is due to the unknown information that $d^{\pi^\ast}_\gamma \ll {d}^{\pi^b}$. When $d^{\pi^\ast}_\gamma \perp {d}^{\pi^b}$ as discussed in Case (c) of Theorem \ref{thm: regret bound for primal}, it can be seen that $\widetilde \lambda^{\pi^\ast}_2(\calS \times \calA) = 1$. We showcase that the regret of our estimated policy can still possibly converge to $0$ in a satisfactory rate without any data coverage assumption. To achieve this, we rely on the Bellman completenss and the smoothness condition for enabling the extrapolation property. To the best of our knowledge, this is the first regret guarantee in  batch RL without assuming the absolute continuity. See Table \ref{tab:main results} in Appendix for a summary of our main assumptions for obtaining the regret of the proposed algorithm compared with existing ones. Finally, when the smoothness condition stated in Assumption \ref{ass: regularity} fails, i.e., $c = 1/2$, the upper bound on the regret of our estimated policy becomes a constant and our method may not guarantee to find an optimal policy asymptotically. 
\end{remark}
\begin{remark}
	Our proposed algorithm is motivated by Lemma \ref{lm: OPE decomposition with mean embedding}, which provides a careful decomposition on the estimation error of OPE without relying on the absolute continuity. The key reason for the possible success of our algorithm in handling insufficient data coverage (i.e., the singular part) relies on the smoothness condition we imposed on modeling $Q^\pi$, i.e., Assumption \ref{ass: regularity}, which provides a strong extrapolation property for our estimated $\widehat \calT^\pi Q$ for $Q \in \calF$ and $\pi \in \Pi$. Specifically, for any state-action pair $(s, a) \in F$, where $F$ is a measurable set such that $\bar d^{\pi^b}_T(F) = 0$ but $\bar d^{\pi^b}_T(F^c) = 1$, by leveraging the convolution operator $\calL_k$ defined in Appendix \ref{app: Theory},  the corresponding $\calT^\pi Q(s, a)$ can be extrapolated by the kernel smoothing of $\calT^\pi Q(\widetilde s, \widetilde a)$ for all $(\widetilde s, \widetilde a) \in F^c$. Therefore, $\widehat \calT^\pi Q$ can approximate $\calT^\pi Q$ well in terms of the RKHS norm. Such an extrapolation property is inherited by $Q$-function due to the Bellman equation that $\calT^\pi Q^\pi = Q^\pi$. 
\end{remark}

Next, we develop a finite-sample regret bound for $\widehat{\pi}_{\text{dual}}$. 
To proceed, we need one additional assumption. \begin{assumption}\label{ass: dual variable}
	$\calF$ is a convex set.
\end{assumption}

\begin{theorem}\label{thm: regret bound for dual problem}
	Under Assumptions \ref{ass: Markovian}-\ref{ass: kernel}, all statements in Theorem \ref{thm: regret bound for primal} hold for $\widehat{\pi}_{\text{dual}}$ as well.
\end{theorem}

\begin{remark}
	The implications behind Theorem \ref{thm: regret bound for dual problem} are the same as those of Theorem \ref{thm: regret bound for primal}. The benefit of considering $\widehat{\pi}_{\text{dual}}$ mainly comes from the computational efficiency. However, assuming $\calF$ convex could be restrictive. Therefore it will be interesting to study the behavior of duality gap when $\calF$ is modeled by a non-convex but ``asymptotically" convex set. This will allow the use of some deep neural network architectures such as \cite{zhang2019deep} in practice.  We leave it for future work. 
\end{remark}

%
%
%
%
\section{Simulation Studies}\label{sec: numerical}



In this section, we use Monte Carlo (MC) experiments to systematically study the proposed method in terms of the convergence of our algorithm, the effect of pessimism in finding the optimal policy, and the robustness to the covariate shift.
For simplicity, we only consider the contextual bandit problem.

For all numerical studies, we compare our method with two popular continuous-action policy optimization methods. 
The regression-based approach first estimates the reward function $r(s,a)$ as $\hat{r}(s,a)$ by running a regression method, and then estimates the value of any given policy $\pi$ with the plug-in estimator $\EE_{S_0 \sim \nu}\left[\hat{r}(S_0, \pi(S_0))\right]$, where the expectation is approximated via MC sampling of $\nu$. Recall that $\nu$ is known in advance. Given such a policy value estimator, we implement the off-the-shelf optimization algorithm to estimate the optimal policy within a given policy class. 
The kernel-based approach \citep{chen2016personalized, kallus2018policy} is similar to the regression approach but first estimates the policy value using the inverse probability weighting with kernel smoothing. 
Specifically, we estimate the value of any given policy $\pi$ by 
$$
(Nh)^{-1} \sum_{i=1}^N \frac{K( (\pi (S_{i,0}) - A_{i,0}) / h)}{ \pi^b(A_{i, 0} \given S_{i, 0})} \times R_{i,0},
$$ 
where $K(\cdot)$ is a kernel function, $h$ is the bandwidth parameter, and $\pi^b$ is  referred as to the generalized propensity score in this setting. 

Next, we describe the data generating process in our simulation study. We consider the mean reward function
$r(s, a) = \big(a - \boldsymbol{B}s \big)^\top \boldsymbol{C} \big(a - \boldsymbol{B}s \big)$, 
where 
the state-value $s$ has dimension $d_s = 5$, 
the action-value $a$ has dimension $d_a = 4$, 
the parameter matrix $\boldsymbol{B} \in \mathcal{R}^{d_a \times d_s}$, 
and  $\boldsymbol{C}$  is a $d_a \times d_a$ negative definite matrix. 
We randomly sample every entry of $\boldsymbol{B}$ from $\text{Uniform}(0,1)$. 
We construct $\boldsymbol{C} = - \boldsymbol{C}_0^\top \boldsymbol{C}_0$, where every entry of $\boldsymbol{C}_0$ is sampled from the standard normal. 
Our reward $R_0$ is generated following 
$R_0 = r(S_0, A_0) + \epsilon_0$, where the independent noise 
$\epsilon_0 \sim \mathcal{N}(0, 1)$. 
Therefore, the optimal policy $\pi^*(s) = s^\top \boldsymbol{B}$ for every $s \in \calS$, regardless of the state distribution. 


The training dataset $\{S_{i,0}, A_{i,0}, R_{i,0}\}_{1 \le i \le N}$ is generated as follows. 
The state $S_{i,0}$ is uniformly sampled from $[0,2]^{d_s}$, which is the same as the reference distribution. 
Except for the case where we study the effect of covariate shift, values of learned policies in this section are evaluated over the same distribution as the training distribution. 
The action is sampled following a behavior policy $\pi^b$ such that $A_{i,0} = \boldsymbol{B}S_{i,0} + \epsilon'$, where $\epsilon' \sim \mathcal{N}(0, \sigma_b^2  \boldsymbol{I}_{d_a})$. 
Therefore, a larger $\sigma_b$ indicates that the behavior policy is more different from the optimal one and also implies that the action space is explored more in the training data.


For all three methods, we search the optimal policy within the class of linear deterministic policies $\Pi = \{\pi \, \mid \, \text{for every $s \in \calS$,}\,\, \pi(s) =  \Tilde{\boldsymbol{B}}s \,  \, \text{for some $\boldsymbol{\Tilde{B}} \in [-1,1]^{d_s \times d_a} $} \}$. 
For the kernel-based method, we assume the true value of the generalized propensity score $\pi^b$ is \textit{known}, instead of plugging-in the estimated value. 
Although the latter is known to be asymptotically more  efficient in the causal inference literature, we observe that its empirical performance is close or sometimes worse than the oracle version in the studied setting, given the challenge to estimating a multi-dimensional conditional density function (See Appendix \ref{sec:appendix_numerical}). 
We tune the bandwidth for the kernel-based method and use the best one for it. 
To implement the optimization problems in two baseline methods, we adopt the widely used L-BFGS-B optimization algorithm following \citet{liu1989limited}. 
Regarding estimating the reward function $r(s,a)$, we use a neural network of three hidden layers with $32$ units for each layer and ReLU as the activation function. 
\textcolor{black}{For \textbf{STEEL}, we use the Laplacian kernel $k(x, y) = \exp \big(-||x-y||_1 / \gamma \big)$ in both the kernel ridge regression and the modeling of $\omega$, where the bandwidth $\gamma$ is picked based on the popular median heuristic, i.e., set as the median of $\{||(S_{i,0}, A_{i,0}) - (S_{j,0}, A_{j,0}) ||_1\}_{1 \le i, j \le N}$. 
For the other tuning parameters, we set $\zeta_{NT} = 0.001$, $(\varepsilon^{(1)}_{NT})^2 = 300$, and $(\varepsilon^{(2)}_{NT})^2 = 600$. 
}


For each setting below, we run $100$ repetitions and report the average regret of each method as well as its standard error. 
Except for when we study the convergence of three methods in terms of the sample size, we fix $N = 200$ data points for each repetition. 



\begin{figure*}[t]
\hspace{-.3cm}
\centering
 \includegraphics[width=0.6\textwidth]{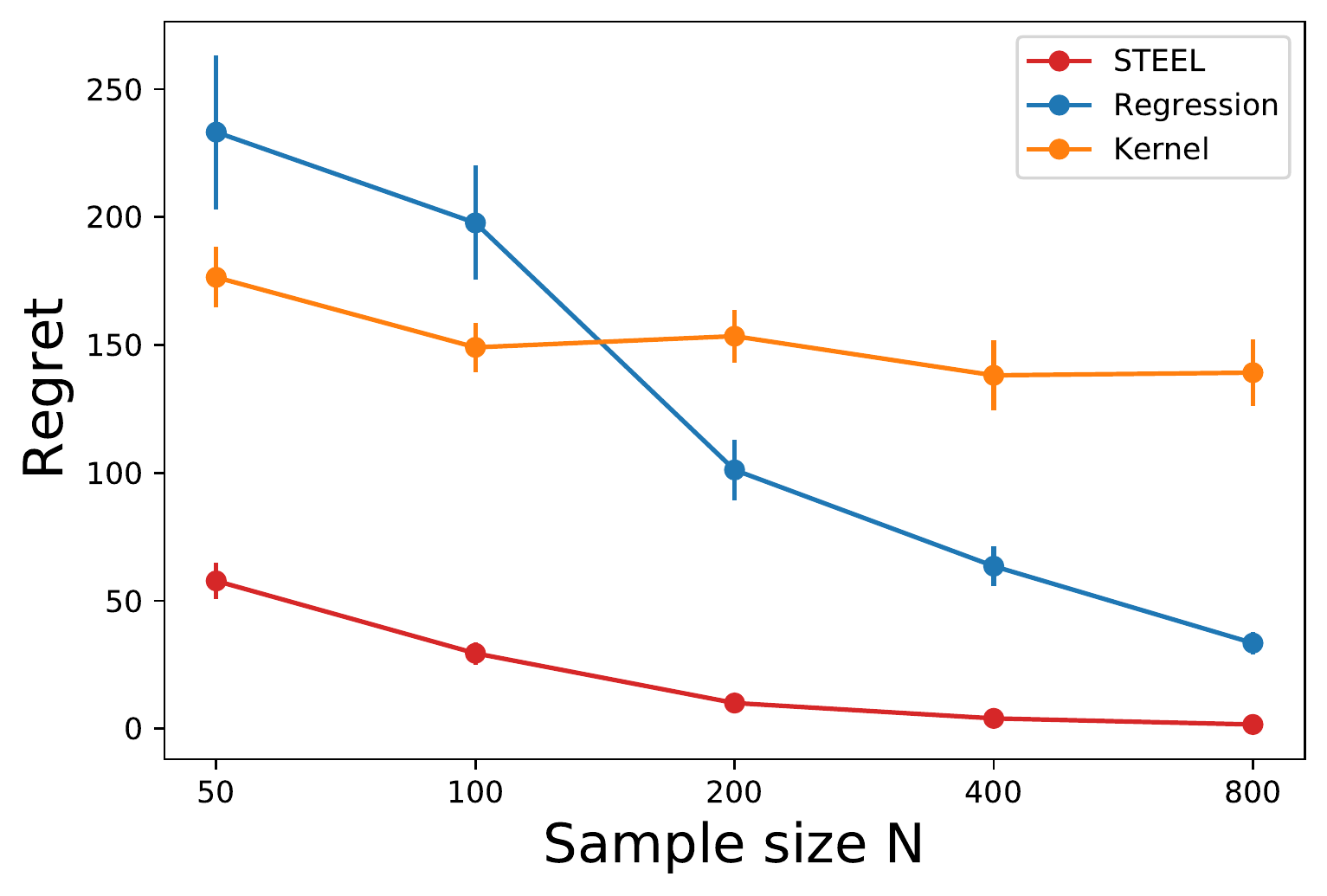} 
\caption{Convergence of different methods. 
The error bars indicate the standard errors of the averages. 
The x-axis is plotted on the logarithmic scale. 
}
\label{fig:convergence}
\end{figure*}

\textbf{Convergence. }
We first study the convergence of all three methods, by increasing the training sample size $N$. 
The results when $\sigma_b = 0.5$ are presented in Figure \ref{fig:convergence}. 
Overall, we observe that our method yields a desired convergence with the regret decaying to zero very quickly. 
The regression-based method also improves as the sample size increases, despite that the finite-sample performance is worse compare with ours. 
The kernel-based method suffers from the curse of dimensionality of the action space and the regret does not decay very significantly. Results under a few other settings are reported in the appendix, with similar findings.

\begin{figure*}[t]
\hspace{-.3cm}
\centering
 \includegraphics[width=0.6\textwidth]{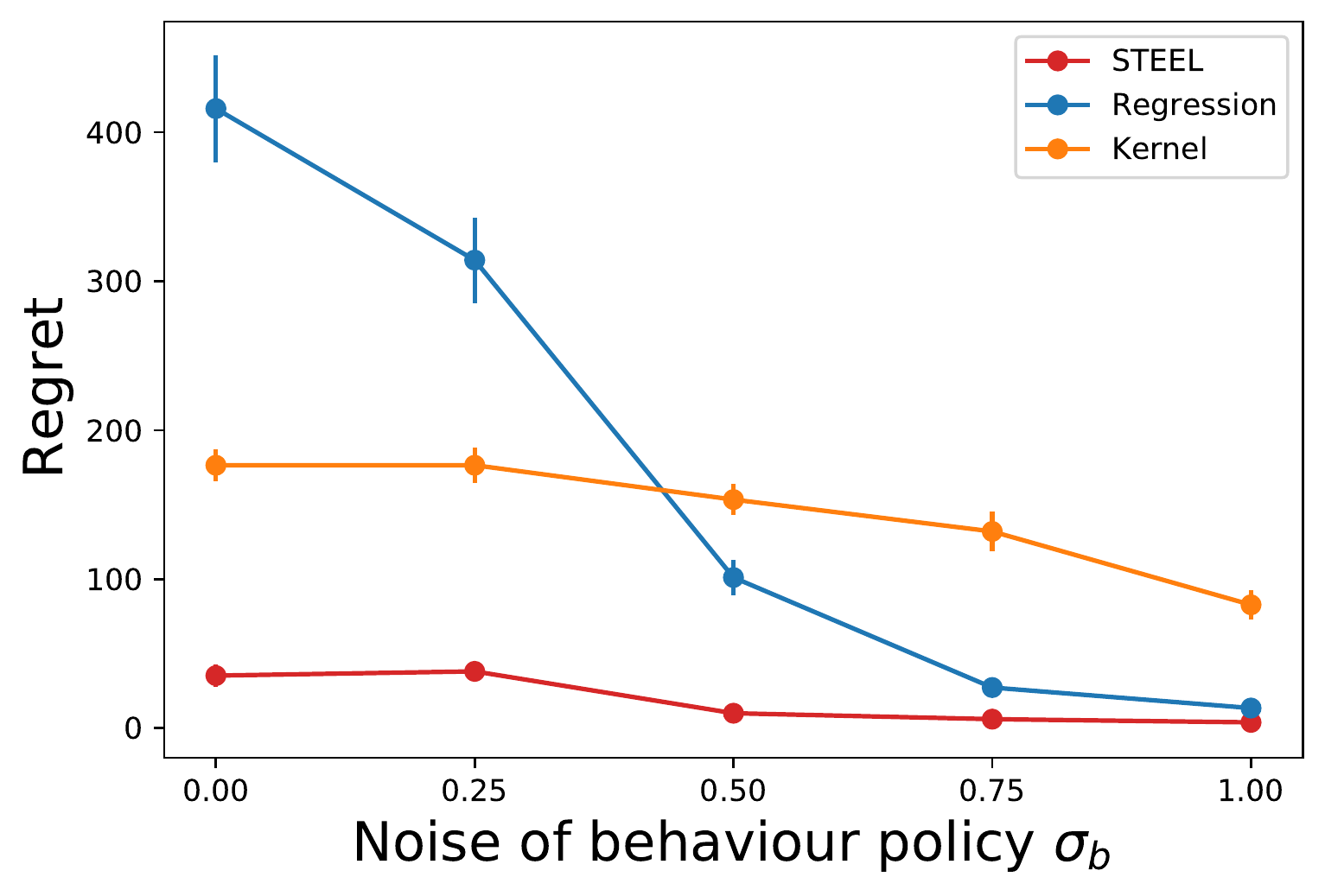} 
\caption{Effect of pessimism. 
The error bars indicate the standard errors of the averages.
}
\label{fig:pessimism}
\end{figure*}

\textbf{Effect of pessimism. }
Next, recall that $\sigma_b$ controls the degree of exploration in the training data. It is known that a well-explored action space in the training dataset is beneficial for policy learning, in the sense that we can accurately estimate the values of most candidate policies. In contrast, when $\sigma_b$ is small (i.e., the behavior policy is close to the optimal one), although the value estimation of the optimal policy may be accurate, that of a sub-optimal policy suffers from a large uncertainty and by chance the resulting estimator could be very large, which leads to returning a sub-optimal policy. 
Our method is designed with addressing such a large uncertainty in mind, by utilizing the pessimistic mechanism. 

In Figure \ref{fig:pessimism}, we empirically study the effect of $\sigma_b$ on different methods. 
The results are based on $N = 200$ data points and different values of $\sigma_b$. 
It can be seen clearly that, our method is fairly robust, when the training data is either well explored or not. 
In contrast, the performance of the two baseline methods deteriorates significantly when $\sigma_b$ is small.

\textbf{Performance of adaptive STEEL. }
In this section, we numerically study the performance of the adaptive version of STEEL, i.e.,  Algorithm \ref{alg:STEEL-adaptive}. 
Notice that the singularity can be caused by the covariate shift and the distance between the data and the target distributions will also increase with covariate shift, we design such a setting for our study. 
Specifically, the policy will be tested when the covaraites are sampled uniformly from $[\Delta x, 2 + \Delta x]^{d_s}$. 
A larger value of $\Delta x$ hence indicates a larger degree of covariate shift. 
We set a very large initial $\varepsilon^{(2)}_{NT} = 5000$, and report the performance of both the first-stage policy and the final policy from 
Algorithm \ref{alg:STEEL-adaptive}. 

\begin{figure*}[t]
\hspace{-.3cm}
\centering
\includegraphics[width=0.6\textwidth]{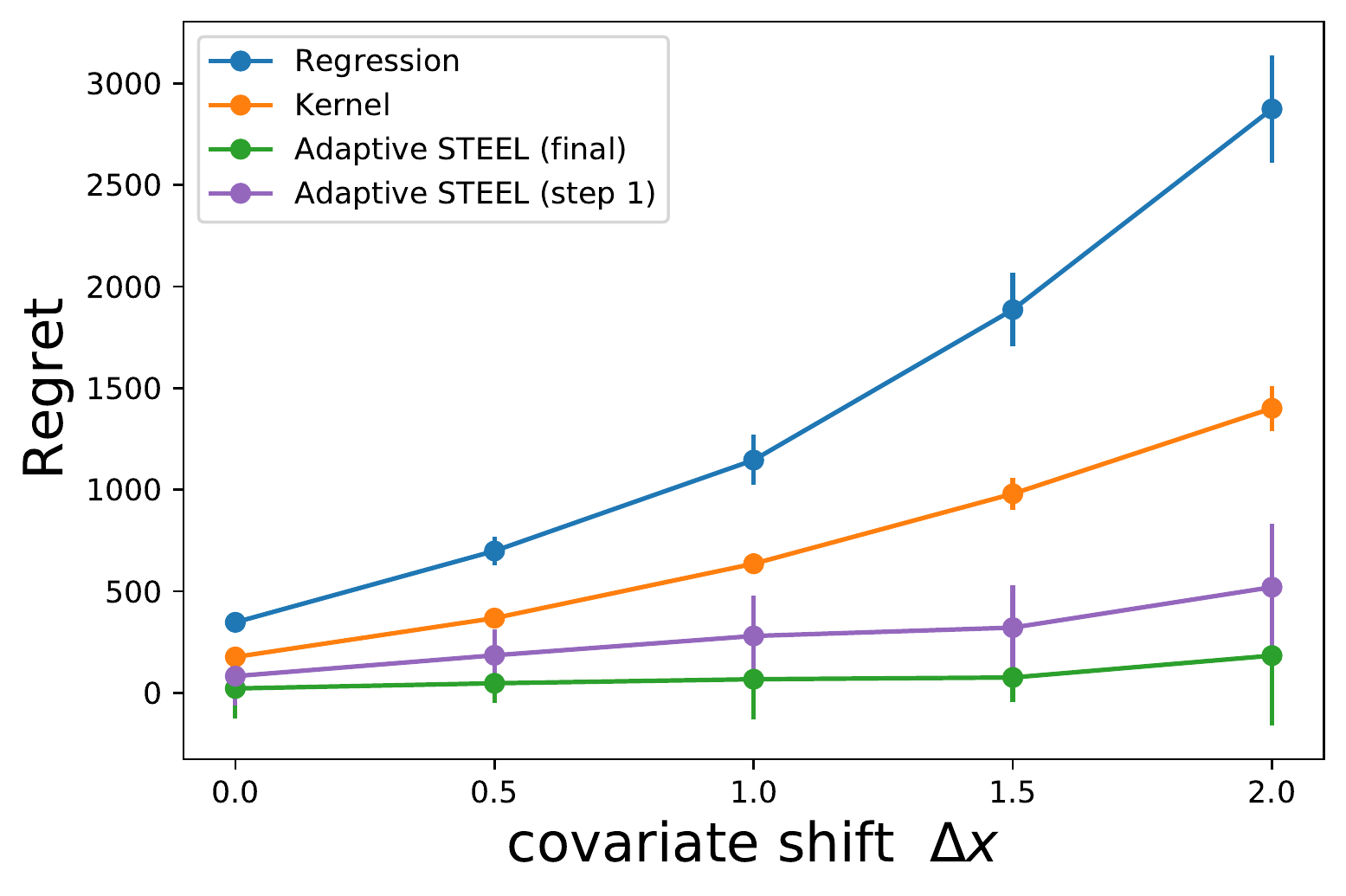}
\caption{
Robustness to covariate shift. 
The error bars indicate the standard errors of the averages.
}
\label{fig:shift}
\end{figure*}

The results when $\sigma_b = 0.25$ and $N  = 200$ are presented in Figure \ref{fig:shift}. 
It can be seen clearly that adaptive STEEL, without further tuning, consistently shows robust and better performance than other algorithms. 
The first-stage outcome policy from Algorithm \ref{alg:STEEL-adaptive}, although still has decent performance, has significantly higher regrets that grow with the degree of covariate shift due to its lack of adaptiveness from appropriately constructed uncertainty set. 


\section{Real Data Application}\label{sec:real_data}


In this section, we apply our method to personalized pricing with a real dataset. 
The dataset is from an online auto loan lending company in the United States \footnote{The dataset is available at \nolinkurl{https://business.columbia.edu/cprm} upon request.}. 
It was first studied by \citet{phillips2015effectiveness} and we follow similar setups as in subsequent studies \citep{ban2021personalized}. 


The dataset contains all auto loan records (208085 in total) in a major online auto loan lending company from July 2002 to November 2004. 
For each record, 
it contains some covariates (e.g., FIFO credit score, the term and amount of loan requested, etc.), 
the monthly payment offered by the company, 
and the decision of the applicant (accept the offer or not). 
We refer interested readers to \citet{phillips2015effectiveness} for more details of the dataset.

The monthly payment (together with the loan term) can be regarded as the offered price (\textit{action}) and the binary decision of the applicant reflects the corresponding demand. 
Specifically, we follow \citet{ban2021personalized} and \citet{bastani2022meta} to define the price as 
\begin{equation*}
    \text{price} =  \text{Monthly payment} \times \sum_{t=1}^{\text{Term}} (1 + \text{Prime rate})^{-t} - \text{amount of loan requested}, 
\end{equation*}
which considers the net present value of future payments. 
Here, the prime rate is the interest rate that the company itself needs to pay. 
We filter the dataset to exclude the outlier records (defined as having a price higher than \$10,000), and $99.49\%$ data points remain in the final dataset.

In our notations, the price is hence the action $A_0$. 
Let the binary decision of the applicant be $D_0$. 
The reward is then naturally computed as $R_0 = D_0 \times A_0$. 
We use the same set of features which are identified as significant in  \citet{ban2021personalized} to construct our feature vector $S_0$: it contains the FICO credit score, the loan amount approved, the prime rate, the competitor's rate, and the loan term. 

When running offline experiments for problems with \textit{discrete} action spaces, 
to compared different learned policies, 
the standard procedure is to find those data points where the recorded actions are consistent with the recommendations from a policy $\pi$ that we would like to evaluate. 
However, with a \textit{continuous} action space, such a procedure is typically impossible, as it is difficult for two continuous actions to have exactly the same value. 
Therefore, it is well acknowledged that, offline experiments without any model assumption is infeasible. 
We closely follow \citet{bastani2022meta} to design a semi-real experiment, 
where we fit a linear demand function from the real dataset and use it to evaluate the policies learned by different methods. 
The linear regression uses the concatenation of $S_0$ (which describes the baseline effect) and $S_0 A_0$  (which describes the interaction effect) as the covariate vector. 
To evaluate the statistical performance of the three methods considered in Section \ref{sec: numerical}, 
we repeat for $100$ random seeds, 
and for each random seed we sample $N$ data points for policy optimization. 

The implementation and tuning details of the three methods are almost the same with Section  \ref{sec: numerical}. 
The main modification is that, for the reward function class used in our method and the regression-based method, we use the neural network to model the demand function (instead of the expected reward itself), which multiplies with the action (price) gives the expected reward. 
Such a modification utilizes the problem-specific structure. 
All covariates are normalized and we always add an intercept term. 
For tuning parameters, we set $\zeta_{NT} = 0.001$, $(\varepsilon^{(1)}_{NT})^2 = 13$, and $(\varepsilon^{(2)}_{NT})^2 = 16$.

\begin{figure}
     \centering
     \begin{subfigure}[b]{0.45\textwidth}
         \centering
         \includegraphics[width=\textwidth]{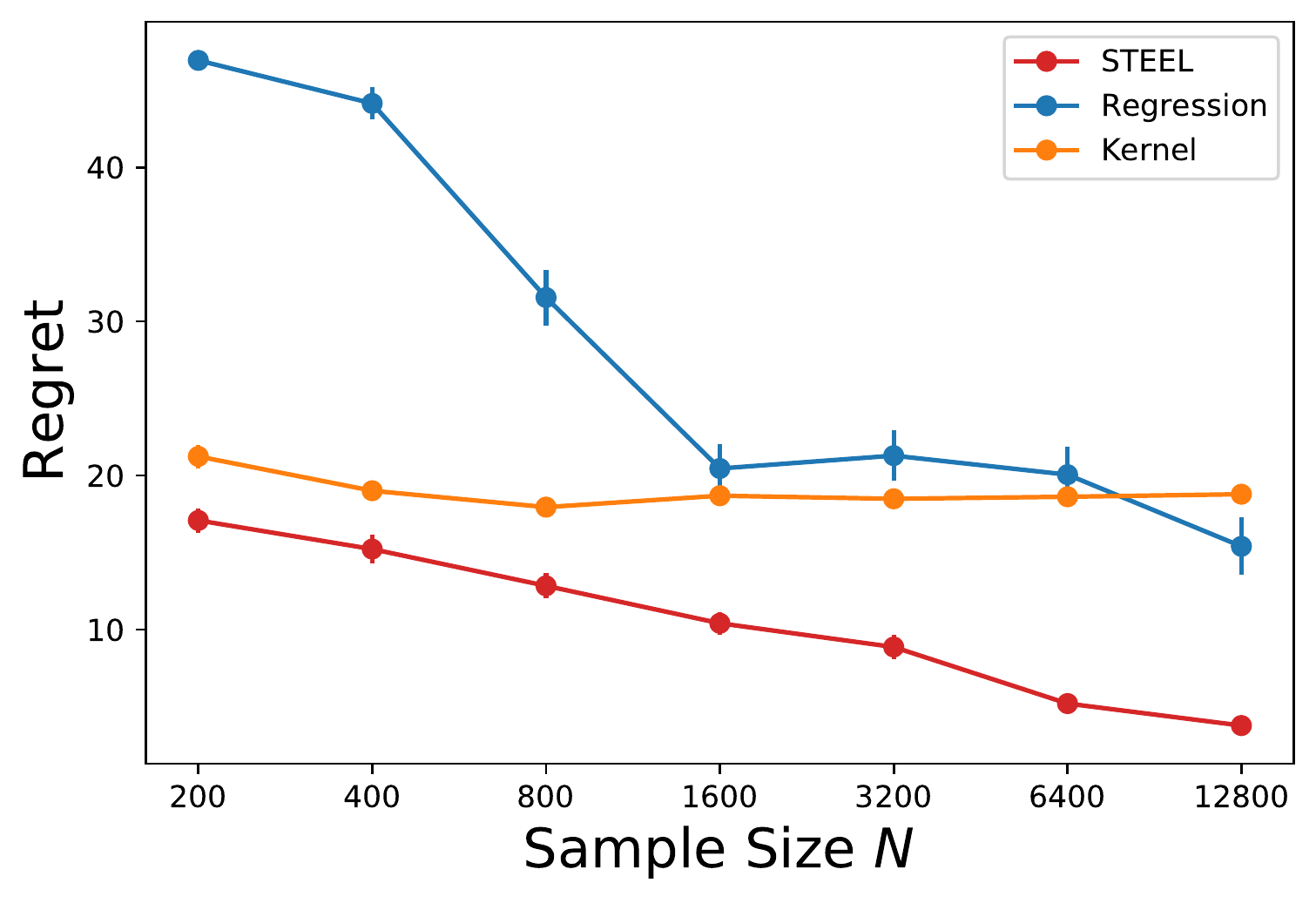}
         \caption{Regret trend with sample size. The error bars indicate the standard errors of the averages. }
         \label{fig:real_N}
     \end{subfigure}
     \hfill
     \begin{subfigure}[b]{0.45\textwidth}
         \centering
         \includegraphics[width=\textwidth]{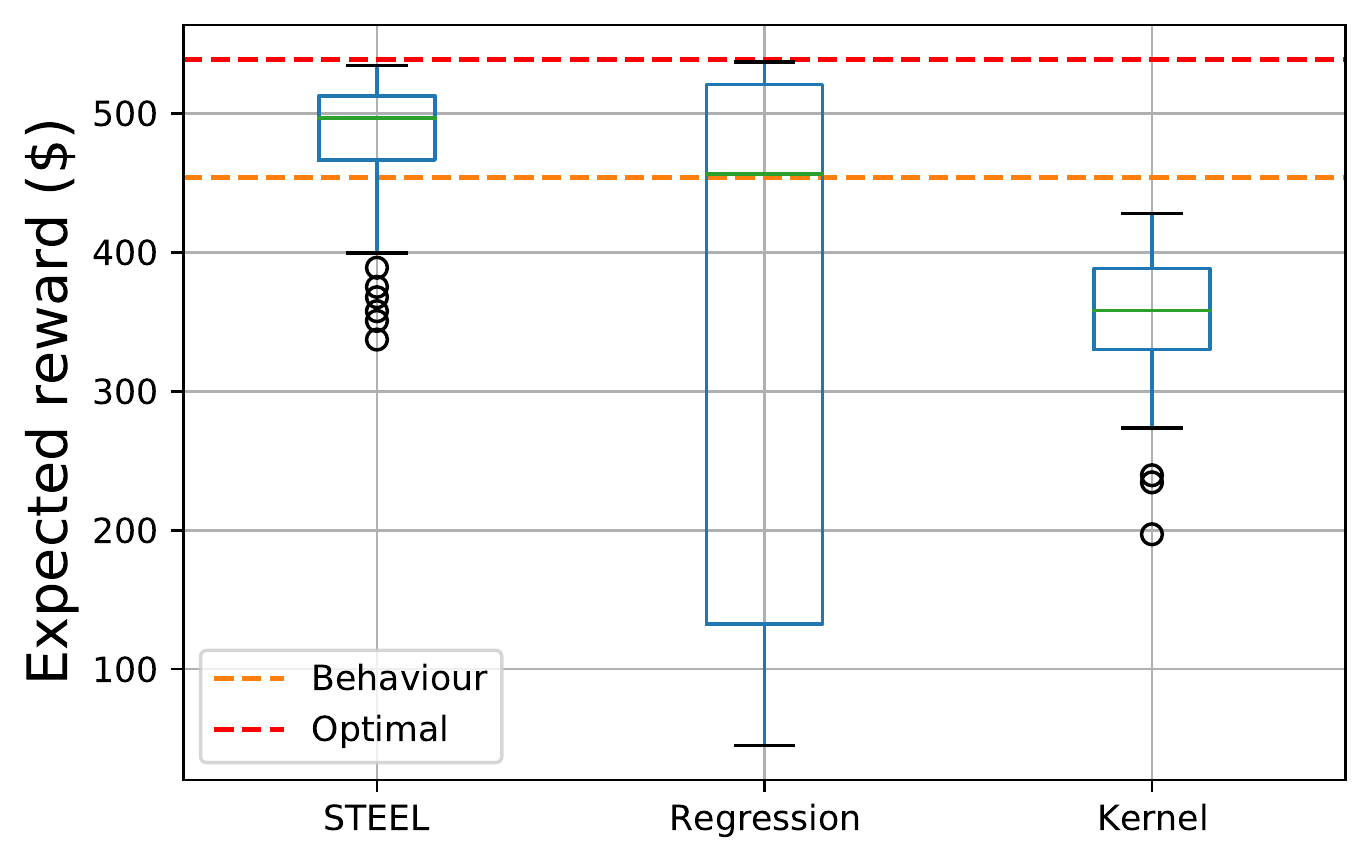}
         \caption{Value distribution over $100$ seeds when $N = 8000$. }
         \label{fig:Real_boxplot}
     \end{subfigure}
\caption{Performance in personalized pricing with real data. }
\label{fig:three graphs}
\end{figure}

In Figure \ref{fig:real_N}, we increase $N$ from $200$ to $12800$ and report the average regrets. 
We find that our method consistently outperforms the other two methods and its regret decays to zero. 
In Figure \ref{fig:Real_boxplot}, we zoom into the case where $N = 8000$, and report the values of the learned policies across the $100$ random seeds. 
We observe that, although the regression-based method can outperform the behaviour policy in about half of the random replications, 
it can perform very poor in the remaining replicates. 
In contrast, our method has an impressive performance and the robustness of our methods is consistent with our methodology design. 
The kernel-based method does not perform well in most settings. 
A deep dive shows that this is because this method over-estimates the values of some sub-optimal policies that assign many actions near the boundaries where we have few data points. 
The over-estimation is due to its poor extrapolation ability and the lack of the pessimism mechanism (i.e., it is not aware of the uncertainty).

Finally, we randomly pick a policy learned by our method (with the first random seed we used) and report its coefficient to provide some insights. 
Recall that we use a linear policy, mapping from the five features to the recommended price. 
The coefficients are overall aligned with the intuition: 
the learned personalized pricing policy offers a higher price when 
the FICO score is lower, 
the loan amount approved is higher, 
the prime rate (the interest rate that the company itself faces) is higher, 
the competitor's rate is higher (implying a consensus on the high risk of this loan), 
and the term is longer.



\setlength{\tabcolsep}{1pt}
\begin{table}[t]
\centering
\caption{Coefficients of the learned linear personalized pricing policy.}
\vspace{-.2cm}
\label{tab:comparison}
\begin{tabular}{ccccc}
\toprule
FICO score \; 
& Loan amount approved \;
&  Prime rate \;
& Competitor's rate \;
& Term \;
\\
\midrule
-0.135 & 0.041 & 0.180 & 0.027 & 0.514   \\
\bottomrule
\vspace{-0.8cm}
\end{tabular}
\end{table}




\section{Conclusion}\label{sec: conclusion}

In this paper, we study batch RL in the presence of singularities and propose a new policy learning algorithm to tackle this issue. Our proposed method  finds an optimal policy without requiring absolute continuity on the distribution of target policies with respect to the data distribution. Both theoretical guarantees and numerical evidence are presented to illustrate the superior performance of our proposed method, compared with existing works. In the current proposal, we leverage the MMD, together with distributionally robust optimization, to enable the extrapolation for addressing the singularity. Our idea can also be extended to other distances such as the Wasserstein distance or total variation distance. While these distances have some appealing properties, the related estimation becomes difficulty, showing a trade-off. Finally, it is also interesting to explore model-based solutions for addressing the singularity issue in batch RL.

\bibliographystyle{agsm}
\bibliography{reference}

\newpage
\appendix

\section{Relation to Existing Work}
Classical methods on batch RL mainly focus on either value iteration or policy iteration  \citep{watkins1992q,sutton1998introduction,antos2008fitted}. 
As discussed before, these methods require the full-coverage assumption on the batch data for finding the optimal policy, which is hard to satisfy in practice due to the inability to further interact with the environment. 
Failure of satisfying this condition often leads to unstable performance such as the lack of convergence or error magnification \citep{wang2021instabilities}.

Recently, significant efforts from the empirical perspective have been made trying to address the challenge from the insufficient data coverage \citep[e.g.,][]{kumar2020conservative,fujimoto2019off}. 
The key idea of these works is to restrict the policy class within the reach of the batch dataset, so as to relax the stringent full-coverage assumption. 
To achieve this, 
the underlying strategy is to adopt the principle of pessimism for modeling the state-value function, in order to discourage the exploration over state-action pairs less-seen in the batch data. 
See \cite{liu2020provably,rashidinejad2021bridging,jin2021pessimism,xie2021bellman,zanette2021provable,zhan2022offline,fu2022offline}. 
Thanks to these pessimistic-type algorithms, the full-coverage assumption can be relaxed to the partial coverage or the so-called single-trajectory concentration assumption, i.e., the distribution induced by the (in-class) optimal policy is absolutely continuous with respect to the one induced by the behavior policy. 
This enhances the applicability of batch RL algorithms to some extent. 
However, this partial coverage assumptions still cannot be verified and hardly holds, especially when the state-action space is large or when the imposed policy class to search from is very complex (e.g., neural networks) that leads to unavoidable over-exploration. 
Our STEEL algorithm can address this limitation without assuming any form of absolute continuity, and can find an (in-class) optimal  policy with a finite-sample regret guarantee. 
Thus, STEEL could be more generally applicable than existing solutions. 
The price to pay for this appealing property is a slightly stronger modeling assumption, i.e., Bellman completeness with respect to a smooth function class, which provides a desirable extrapolation property so that the singular part incurred by the distributional mismatch can be properly controlled.

Our STEEL algorithm is particularly useful for policy learning with a continuous action space, where the singularity arises naturally in the batch setting. 
This fundamental issue hinders the theoretical understanding of many existing algorithms such as deterministic policy gradient algorithms and their variants \citep[e.g.,][]{silver2014deterministic,lillicrap2015continuous}. To the best of our knowledge, only \cite{antos2008fitted,kallus2020doubly} in the RL literature tackle this problem. 
However, \cite{antos2008fitted} imposes a strong regularity condition on the action space, which is hard to interpret and essentially implies that the $L^\infty$-norm of any function over the action space is bounded by its $L^1$-norm (multiplied by some constant). On the other hand, \cite{kallus2020doubly} used kernel smoothing techniques on the deterministic policy, which is known to suffer from the curse of dimensionality when the action space is large.
Moreover, both \cite{antos2008fitted} and \cite{kallus2020doubly} cannot address the possible singularity over the state space. 
In contrast, our STEEL algorithm can handle the singularity in both the state and action spaces with a strong theoretical guarantee. 
To the best of our knowledage, our paper is the first to address the aforementioned issue. 

\section{Details of Theoretical Results}\label{app: Theory}

In this section, we list all additional technical assumptions and related discussions for our theoretical results.

\subsection{Additional Technical Assumptions}\label{subsec: additional technical assumptions}

\begin{assumption}\label{ass: kernel}
	The kernel $k(\bullet, \bullet): \calZ \times \calZ \rightarrow \mathbb{R}$ is positive definite, and
	$\sup_{Z \in \calZ}\sqrt{k(Z, Z)} = \kappa < + \infty$. For any $\pi \in \Pi$, $k$ is  measurable with respect to both $d_\gamma^\pi$ and $\lambda^\pi_2$.
\end{assumption}
The bounded and measurable conditions on the kernel $k$ in Assumption \ref{ass: kernel} are mild, which can be easily satisfied by many popular kernels such as the Gaussian kernel. Assumption \ref{ass: kernel} ensures that the conditions in Lemma \ref{lm: existence of embeddings} of the appendix hold so that the kernel mean embeddings exist. 
Assumption \ref{ass: kernel} also indicates that $\calH_k$ is compactly embedded in $L^2_{\bar d^{\pi^b}_T}(\calZ)$, e.g., for any $f \in \calH_k$, $f \in L^2_{\bar d^{\pi^b}_T}(\calZ)$. See Definition 2.1 and Lemma 2.3 of \cite{steinwart2012mercer} for more details.
Define an integral operator $\calL_k: L^2_{d^{\pi^b}}(\calZ) \rightarrow L^2_{d^{\pi^b}}(\calZ)$ as
\begin{align}\label{def: operator to H}
	\calL_k f(z) \triangleq \int_{\calZ} k(z, \widetilde z) f(\widetilde z)d^{\pi^b}(\widetilde z)\text{d}z, \quad \text{for $z \in \calZ$ and $f \in L^2_{\bar d^{\pi^b}_T}(\calZ)$}.
\end{align}
Assumption \ref{ass: kernel} implies that the self-adjoint operator $\calL_k$ is compact and enjoys a spectral representation such that for any $f \in L^2_{\bar d^{\pi^b}_T}(\calZ)$,
\begin{align}\label{def: spetral decomposition of Tk}
	\calL_k f = \sum_{i = 1}^{+\infty} e_j \langle\phi_j, f \rangle_{L^2_{\bar d^{\pi^b}_T}(\calZ)} \phi_j,
\end{align}
where $\langle \bullet, \bullet \rangle_{L^2_{\bar d^{\pi^b}_T}(\calZ)}$ is referred to as the inner product in $L^2_{d^{\pi_b}}(\calZ)$, $\{e_j\}_{j \geq 1}$ is a non-increasing sequence of eigenvalues towards $0$ and $\{\phi_j\}_{j \geq 1}$ are orthonormal bases of $L^2_{d^{\pi^b}_T}(\calZ)$. 
See Theorem 2.11 of \cite{steinwart2012mercer} for a justification of such spectral decomposition. 

Lastly, we provide a sufficient condition so that Assumption \ref{ass: function class}~\eqref{ass: vector-valued function class} holds.
\begin{assumption}\label{ass: Lips for covering} For any $\pi_1, \pi_2 \in \Pi$ and $Q_1, Q_2 \in \calF$, $\norm{g_{\pi_1, Q_1} - g_{\pi_2, Q_2}}_{L^2_{d^{\pi_b}}} \lesssim \norm{\norm{\pi_1 - \pi_2}_\infty}_{\ell_1} + \norm{Q_1 - Q_2}_\infty$, where $\calL_k^cg_{\pi_1, Q_1} = \calT^{\pi_1} Q_1$ and $\calL_k^cg_{\pi_2, Q_2} = \calT^{\pi_2} Q_2$.
	In addition, $\xi_{NT} \lesssim 1$ uniformly over $N$ and $T$.
\end{assumption}

Then we have the following lemma.
\begin{lemma}\label{lm: complexity for calG}
	Under Assumptions \ref{ass: reward}, \ref{ass: function class}~\eqref{ass: Q-function class}, \ref{ass: regularity}, and \ref{ass: Lips for covering}-\ref{ass: kernel}, Assumption \ref{ass: function class}~\eqref{ass: vector-valued function class} holds with $v(\calG) = v(\calF) + v(\Pi)$.
\end{lemma}


\subsection{More Discussions on Theorems}
In this subsection, we present Table \ref{tab:main results} for a summary of our main assumptions for obtaining the regret of
the proposed algorithm compared with existing ones.
\begin{table}[h!]
	\begin{center}
		\caption{Comparison of the main assumptions behind our algorithm and  several state-of-the-art methods. The operator $\calB^\ast$ in the approximate value iteration method is referred to as the optimal Bellman operator \citep{sutton1998introduction}. Pessimistic RL refers to some batch RL algorithms using pessimism. $V^\pi$ is the state-value function of a policy $\pi$ and $\bar \calV$ is the corresponding hypothesized class of functions.}
		\label{tab:main results}
		\scalebox{0.8}{\begin{tabular}{|c|c|c|}
				\hline
				Algorithm & Data Coverage& Modeling Assumption\\
				\hline
				Approximate Value Iteration & \multirow{2}{*}{$\norm{\frac{d^{\pi}_\gamma}{\bar d^{\pi^b}_T}}_\infty < +\infty, \forall \pi$} & $\forall f \in \calF, \calB^\ast f \in \calF$ {\scriptsize	\citep{munos2008finite}}\\
				\cline{1-1}\cline{3-3}
				Approximate Policy Iteration & & $\forall f \in \calF$ and $\pi \in \Pi$, $\calB^\pi f \in \calF$ {\scriptsize\citep{antos2008learning}}\\
				\hline
				\multirow{2}{*} & {$\sup_{f \in \mathcal F} \frac{|| f - \calB^{\pi^\ast} f||_{d^{\pi}_\gamma} }{|| f - \calB^{\pi^\ast} f ||_{\bar d^{\pi^b}}}<+\infty$ \, $d^{\pi^\ast}_\gamma \ll d^{\pi^b}$} 
    & $\forall f \in \calF$ and $\pi \in \Pi$, $\calB^\pi f \in \calF$ {\scriptsize\citep{xie2021bellman}}\\
    \cline{2-3}
				\cline{3-3}
				{Pessimistic RL}&{$\norm{\frac{d^{\pi^\ast}_\gamma}{\bar d^{\pi^b}_T}}_\infty < +\infty$} & $\frac{d^{\pi^\ast}_\gamma}{\bar d^{\pi^b}_T} \in \calW$ and $\forall \pi \in \Pi$, $Q^\pi \in \calF$ {\scriptsize\citep{jiang2020minimax}}\\
				\hline
				& \multirow{2}{*}{$\frac{d^{\pi}_\gamma(s)}{\bar d^{\pi^b}_T(s)}\lesssim 1, \forall\pi,s \in \calS$} & \multirow{4}{*}{$\frac{d^{\pi^\ast}_\gamma}{\bar d^{\pi^b}_T} \in \calW$ and $V^{\pi^\ast} \in \bar \calV$}\\
				PRO-RL \citep{zhan2022offline}&&\\ 
				without regularization &\multirow{2}{*}{$\frac{d^{\pi^\ast}_\gamma(s)}{\bar d^{\pi^b}_T(s)} \gtrsim 1, \forall s \in \calS$}& \\
				& & \\
				\hline
				& \multirow{2}{*}{${d_\gamma^{\pi^\ast} = \widetilde \lambda_1^{\pi^\ast}(\calS \times \calA) \lambda_1^{\pi^\ast}+ \widetilde \lambda_2^{\pi^\ast}(\calS \times \calA)\lambda_2^{\pi^\ast}}$} & \multirow{4}{*}{$\frac{\lambda_1^{\pi^\ast}}{\bar d^{\pi^b}_T} \in \calW; \forall \pi \in \Pi$, $Q \in \calF$, $\calB^\pi Q \in \calF \subseteq \calH_k^{2c}$}\\
				\textbf{STEEL}&&\\ 
				(general) &\multirow{2}{*}{$\norm{\frac{\lambda_1^{\pi^\ast}}{\bar d^{\pi^b}_T}}_\infty < + \infty$, $\text{MMD}_k(\bar d^{\pi^b}_T, \lambda_2^{\pi^\ast}) < +\infty$ }& \\
				& & \\
				\hline
			\end{tabular}
		}
	\end{center}
\end{table}
In the following, we provide more discussion on Theorem \ref{thm: regret bound for primal}.
\begin{remark}
	In Case (b) of Theorem \ref{thm: regret bound for primal}, where $d^{\pi^\ast}_\gamma \ll {d}^{\pi^b}$, to achieve the desirable regret guarantee, by implementing our algorithm without using the constraint set $\Omega_2(\pi, \calF,  \varepsilon^{(2)}_{NT})$, we indeed do not require our batch data to uniquely identify $Q^\pi$ or $\calV(\pi)$ for $\pi \neq \pi^\ast$ but only $\pi^\ast$. Specifically, for a fixed policy $\pi \neq \pi^\ast$,  we allow that there exist two different $Q^\pi_1$ and $Q^\pi_2$ under the batch data distribution such that
	\begin{align*}
		&\sup_{w \in \calW} \overline \EE\left[w(S, A)\left(R + \gamma Q^\pi_1(S', \pi(S')) - Q^\pi_1(S, A) \right)\right] \\
		= &\sup_{w \in \calW} \overline \EE\left[w(S, A)\left(R + \gamma Q^\pi_2(S', \pi(S')) - Q^\pi_2(S, A) \right)\right] = 0. 
	\end{align*}
	In addition, because we do not require $\omega^\pi$ for $\pi \neq \pi^\ast$ modeled correctly or being uniformly bounded above, the above equality cannot ensure that
	$$
	\calV(\pi)  = (1-\gamma) \EE_{S_0 \sim \nu}\left[Q^\pi_1(S_0, \pi(S_0))\right] =  (1-\gamma) \EE_{S_0 \sim \nu}\left[Q^\pi_2(S_0, \pi(S_0))\right].
	$$
	Thus $\calV(\pi)$ for $\pi \neq \pi^\ast$ are not required to be identified by our batch data either if our proposed algorithm is implemented without incorporating $\Omega_2(\pi, \calF,  \varepsilon^{(2)}_{NT})$.
	However, due to the unknown information on $\pi^\ast$ (e.g., whether Case (b) happens or not), our algorithm has to include the constraint set $\Omega_2(\pi, \calF,  \varepsilon^{(2)}_{NT})$ for handling the error induced by the possibly singular part, which implicitly imposes that $Q^\pi$ and $\calV(\pi)$ can be uniquely identified by our batch data for $\pi \in \Pi$ because of the strong RKHS norm used in $\Omega_2$. This is another trade-off for addressing the issue of a possible singularity.  Lastly, we would like to emphasize $\omega^{\pi^\ast} \in \calW$ and $\text{MMD}_k(\bar d^{\pi^b}_T, \lambda_2^{\pi^\ast})$ are only related to the in-class optimal policy $\pi^\ast$.
\end{remark}

In the following, we provide more discussion on the finite-sample regret bound for $\widehat{\pi}_{\text{dual}}$. Consider the following constrained optimization problem, which corresponds to the population counterpart of \eqref{eqn: regret bound for primal} with fixed $\varepsilon^{(1)}_{NT}$ and $\varepsilon^{(2)}_{NT}$ 
\begin{align*}
	\min_{Q \in \widetilde \Omega(\pi, \calF, \varepsilon^{(1)}_{NT}, \varepsilon^{(2)}_{NT})} \quad (1-\gamma) \EE_{S_0 \sim \nu}\left[Q(S_0, \pi(S_0))\right],
\end{align*}
where
\begin{align*}
	&\widetilde \Omega(\pi, \calF, \varepsilon^{(1)}_{NT}, \varepsilon^{(2)}_{NT}) \\
	=& \left\{Q \in \calF \mid\sup_{w \in \calW}  \overline\EE\left[w(S, A)\left(R + \gamma {Q}(S',  \pi(S')) - {Q}(S, A) \right)\right]  \leq  2\varepsilon^{(1)}_{NT}\right\}\\
	\cap & \left\{Q \in \calF \mid \norm{\calT^{ \pi} {Q}}_{\calH_k} \leq 2\varepsilon^{(2)}_{NT} \right\}.
\end{align*}
Define the corresponding Lagrangian function as
\begin{align}\label{def: population Lagrangian function}
	L(Q, \rho, \pi) &= (1-\gamma)\EE_{S_0 \sim \nu}\left[Q(S_0, \pi(S_0))\right]\\
	& +\rho_1 \times \left\{\sup_{w \in \calW}  \overline\EE \left[w(S, A)\left(R + \gamma {Q}(S', \pi(S')) - {Q}(S, A) \right)\right]  - 2\varepsilon^{(1)}_{NT}\right\}\nonumber \\ 
	& + \rho_2 \times \left\{\norm{\calT^\pi {Q}}_{\calH_k} - 2\varepsilon^{(2)}_{NT} \right\}\nonumber,
\end{align}
which could be roughly viewed as the population counterpart of $L_{NT}(Q, \rho, \pi)$. We ignore the dependency of $L(Q, \rho, \pi)$ on $NT$ for notation simplicity. Assumption \ref{ass: dual variable} is imposed so that together with Assumptions \ref{ass: Markovian}, \ref{ass: reward} and \ref{ass: function class}~\eqref{ass: Q-function class},  the following strong duality holds so that
\begin{align*}
	\min_{Q \in \widetilde \Omega(\pi, \calF, \varepsilon_{NT})} \quad (1-\gamma) \EE_{S_0 \sim \nu}\left[Q(S_0, \pi^\ast(S_0))\right] = \max_{\rho \succeq 0} \min_{Q \in \calF} L(Q, \rho, \pi).
\end{align*}
This implies that there is no duality gap between $\max_{\rho \succeq 0} \min_{Q\in \calF} L_{NT}(Q, \rho, \pi^\ast)$ and 
$$
\min_{Q \in \Omega_1(\pi, \calF, \calW, \varepsilon^{(1)}_{NT}) \cap \Omega_2(\pi, \calF,  \varepsilon^{(2)}_{NT}) } (1-\gamma)\EE_{S_0 \sim \nu}\left[Q(S_0, \pi(S_0))\right]
$$
asymptotically, which is essential for us to derive a finite-sample bound for the regret of $\widehat{\pi}_{\text{dual}}$ in Theorem \ref{thm: regret bound for dual problem}.

	The implications behind Theorem \ref{thm: regret bound for dual problem} are the same as those of Theorem \ref{thm: regret bound for primal}. The benefit of considering $\widehat{\pi}_{\text{dual}}$ mainly comes from the computational efficiency. However, assuming $\calF$ convex could be restrictive. Therefore it will be interesting to study the behavior of duality gap when $\calF$ is modeled by a non-convex but ``asymptotically" convex set. This will allow the use of some deep neural network architectures such as \cite{zhang2019deep} in practice.  We leave it for future work.

\section{Proof of Regret}

\textbf{Proof of Theorem \ref{thm: feasible set}}
By summarizing the results of Lemmas \ref{lm: ac part}-\ref{lm: RKHS convergence}, we can conclude our proof. 

\textbf{Proof of Theorem \ref{thm: regret bound for primal}}: By Theorem \ref{thm: feasible set}, with probability at least $1-1/(NT)$, the regret can be decomposed as
	\begin{align*}
		\text{Regret}(\widehat{\pi}) & = (1-\gamma) \EE_{S_0 \sim \nu}\left[Q^{\pi^\ast}(S_0, \pi^\ast(S_0))\right]- (1-\gamma) \EE_{S_0 \sim \nu}\left[Q^{\widehat \pi}(S_0, \widehat \pi(S_0))\right] \nonumber\\
		& \leq (1-\gamma) \EE_{S_0 \sim \nu}\left[Q^{\pi^\ast}(S_0, \pi^\ast(S_0))\right]- \min_{Q \in \Omega(\widehat \pi, \calF,  \calW, \varepsilon^{(1)}_{NT}, \varepsilon^{(2)}_{NT})}(1-\gamma) \EE_{S_0 \sim \nu}\left[Q(S_0, \widehat \pi(S_0))\right]\\
		& \leq (1-\gamma) \EE_{S_0 \sim \nu}\left[Q^{\pi^\ast}(S_0, \pi^\ast(S_0))\right] -  \min_{Q \in \Omega(\pi^\ast, \calF,  \calW, \varepsilon^{(1)}_{NT}, \varepsilon^{(2)}_{NT})}(1-\gamma) \EE_{S_0 \sim \nu}\left[Q(S_0, \pi^\ast(S_0))\right]\\
		& \leq \max_{Q \in \Omega(\pi^\ast, \calF,  \calW, \varepsilon^{(1)}_{NT}, \varepsilon^{(2)}_{NT}) } \left\{\calV(\pi^\ast) - (1-\gamma) \EE_{S_0 \sim \nu}\left[Q(S_0, \pi^\ast(S_0))\right]\right\},
	\end{align*}
where the first inequality is based on Theorem \ref{thm: feasible set} and the second one relies on our policy optimization algorithm \eqref{eqn: policy optimization algorithm}. The last line of the above inequalities transforms the regret of $\widehat{\pi}$ into the estimation error of OPE for $Q^{\pi^\ast}$ using $Q \in \Omega(\pi^\ast, \calF,  \calW, \varepsilon^{(1)}_{NT}, \varepsilon^{(2)}_{NT})$. Lastly, by leveraging results in Lemma \ref{lm: OPE decomposition with mean embedding}, as long as $\omega^{\pi^\ast} \in \calW$, we  obtain that
\begin{align*}
	&\max_{Q \in \Omega_1(\pi^\ast, \calF, \calW, \varepsilon^{(1)}_{NT}) \cap \Omega_2(\pi^\ast, \calF, \varepsilon^{(2)}_{NT})} \left\{\calV(\pi^\ast) - (1-\gamma) \EE_{S_0 \sim \nu}\left[Q(S_0, \pi^\ast(S_0))\right]\right\}\\
	\leq & \sup_{Q \in \Omega_1(\pi^\ast, \calF,\calW, \varepsilon^{(1)}_{NT})}\sup_{w \in \calW} \overline \EE\left[w(S, A)\left(R + \gamma Q(S', \pi^\ast(S')) - Q(S, A) \right)\right] \nonumber\\
	+ &  \widetilde \lambda_2^{\pi^\ast}(\calS \times \calA) \times \text{MMD}(\bar d^{\pi^b}_T, \lambda_2^{\pi^\ast}) \times \max_{Q \in \Omega_2(\pi^\ast, \calF, \varepsilon^{(2)}_{NT}) }\norm{\calT^{\pi^\ast} Q}_{\calH_k}\\
	\leq & \sup_{Q \in \Omega_1(\pi^\ast, \calF,\calW, \varepsilon^{(1)}_{NT})}\sup_{w \in \calW} \left\{\overline \EE\left[w(S, A)\left(R + \gamma Q(S', \pi^\ast(S')) - Q(S, A) \right)\right] \right. \\
	&\left. - \overline \EE_{NT}\left[w(S, A)\left(R + \gamma Q(S', \pi^\ast(S')) - Q(S, A) \right)\right] \right\} \\
 +& \sup_{Q \in \Omega_1(\pi^\ast, \calF,\calW, \varepsilon^{(1)}_{NT})}\sup_{w \in \calW} \overline \EE_{NT}\left[w(S, A)\left(R + \gamma Q(S', \pi^\ast(S')) - Q(S, A) \right)\right]\\
	+& \widetilde \lambda^{\pi^\ast}_2(\calS \times \calA) \times \text{MMD}(\bar d^{\pi^b}_T, \lambda_2^{\pi^\ast}) \times  \sup_{Q \in \Omega_2(\pi^\ast, \calF, \varepsilon^{(2)}_{NT})}\left\{ \norm{\calT^{\pi^\ast} Q}_{\calH_k} - \norm{\widehat \calT^{\pi^\ast} Q}_{\calH_k} \right\}\\	
	&+ \widetilde \lambda^{\pi^\ast}_2(\calS \times \calA) \times \text{MMD}(\bar d^{\pi^b}_T, \lambda_2^{\pi^\ast}) \times  \sup_{Q \in \Omega_2(\pi^\ast, \calF, \varepsilon^{(2)}_{NT})} \norm{\widehat \calT^{\pi^\ast} Q}_{\calH_k}\\
	\lesssim & \varepsilon^{(1)}_{NT} +  \lambda^{\pi^\ast}_2(\calS \times \calA) \times \text{MMD}(\bar d^{\pi^b}_T, \lambda_2^{\pi^\ast}) \times \varepsilon^{(2)}_{NT},
\end{align*}
where we use results in Lemmas \ref{lm: ac part}-\ref{lm: RKHS convergence} for the last inequality. When $c = 1/2$, by slightly modifying the proof, Theorem \ref{thm: feasible set} holds with $\varepsilon^{(2)}_{NT} \asymp 1$. Then we have the last result of Theorem \ref{thm: regret bound for primal}. This concludes our proof.

\textbf{Proof of Theorem \ref{thm: regret bound for dual problem}}: Denote the solution of 
$$
\max_{\pi \in \Pi, \rho \succeq 0}\min_{Q \in \calF} L_{NT}(Q, \rho, \pi)
$$
by $(\widehat{\pi}_{\text{dual}}, \widehat \rho, \widehat Q)$, where $\widehat \rho \succeq 0$. As shown in Theorem \ref{thm: feasible set},  with probability at least $1- 1/(NT)$, for every $\pi \in \Pi$,
$$
Q^\pi \in \Omega(\pi, \calF,  \calW, \varepsilon^{(1)}_{NT}, \varepsilon^{(2)}_{NT}).
$$
Then with probability at least $1 - 1/(NT)$, we have the following regret decomposition for any constant $C > 0$.
\begin{align*}
	\text{Regret}(\widehat{\pi}_{\text{dual}}) & = (1-\gamma) \EE_{S_0 \sim \nu}\left[Q^{\pi^\ast}(S_0, \pi^\ast(S_0))\right]- (1-\gamma) \EE_{S_0 \sim \nu}\left[Q^{\widehat \pi_{\text{dual}}}(S_0, \widehat \pi_{\text{dual}}(S_0))\right] \nonumber\\
	& \leq (1-\gamma) \EE_{S_0 \sim \nu}\left[Q^{\pi^\ast}(S_0, \pi^\ast(S_0))\right]-(1-\gamma) \EE_{S_0 \sim \nu}\left[Q^{\widehat \pi_{\text{dual}}}(S_0, \widehat \pi_{\text{dual}}(S_0))\right]\\
	& - \widehat \rho_1 \times \left\{\sup_{w \in \calW}  \overline\EE_{NT}\left[w(S, A)\left(R + \gamma {Q^{\widehat \pi_{\text{dual}}}}(S', \widehat \pi_{\text{dual}}(S')) - {Q^{\widehat \pi_{\text{dual}}}}(S, A) \right)\right]  - \varepsilon^{(1)}_{NT}\right\}\nonumber \\ 
	& - \widehat \rho_2 \times \left\{\norm{\widehat \calT^\pi {Q^{\widehat \pi_{\text{dual}}}}}_{\calH_k}- \varepsilon^{(2)}_{NT}\right\}\nonumber \\
	& \leq (1-\gamma) \EE_{S_0 \sim \nu}\left[Q^{\pi^\ast}(S_0, \pi^\ast(S_0))\right]- \min_{Q \in \calF}\left\{ (1-\gamma) \EE_{S_0 \sim \nu}\left[Q(S_0, \widehat \pi_{\text{dual}}(S_0))\right] \right.\\
	&\left. - \widehat \rho_1 \times \left\{\sup_{w \in \calW}  \overline\EE_{NT}\left[w(S, A)\left(R + \gamma {Q}(S', \widehat{\pi}_{\text{dual}}(S')) - {Q}(S, A) \right)\right]  - \varepsilon^{(1)}_{NT}\right\}\nonumber  \right. \\ 
	& \left. - \widehat \rho_2 \times \left\{\norm{\widehat \calT^{\widehat \pi_{\text{dual}}} {Q}}_{\calH_k} -  \varepsilon^{(2)}_{NT}\right\}\nonumber \right\}\\
	& \leq (1-\gamma) \EE_{S_0 \sim \nu}\left[Q^{\pi^\ast}(S_0, \pi^\ast(S_0))\right]- \max_{\rho \succeq 0}\min_{Q \in \calF}\left\{ (1-\gamma) \EE_{S_0 \sim \nu}\left[Q(S_0, \pi^\ast(S_0))\right] \right.\\
	&\left. +\rho_1 \times \left\{\sup_{w \in \calW}  \overline\EE_{NT}\left[w(S, A)\left(R + \gamma {Q}(S',  \pi^\ast(S')) - {Q}(S, A) \right)\right]  - \varepsilon^{(1)}_{NT}\right\}\nonumber  \right. \\ 
	& \left. +  \rho_2 \times \left\{ \norm{\widehat \calT^{ \pi^\ast} {Q}}_{\calH_k} - \varepsilon^{(2)}_{NT} \right\}\nonumber \right\}\\
	& \leq (1-\gamma) \EE_{S_0 \sim \nu}\left[Q^{\pi^\ast}(S_0, \pi^\ast(S_0))\right]- \max_{0 \preceq \rho \preceq C}\min_{Q \in \calF}\left\{ (1-\gamma) \EE_{S_0 \sim \nu}\left[Q(S_0, \pi^\ast(S_0))\right] \right.\\
	&\left. +\rho_1 \times \left\{\sup_{w \in \calW}  \overline\EE_{NT}\left[w(S, A)\left(R + \gamma {Q}(S',  \pi^\ast(S')) - {Q}(S, A) \right)\right]  - \varepsilon^{(1)}_{NT}\right\}\nonumber  \right. \\ 
	& \left. +  \rho_2 \times \left\{\norm{\widehat \calT^{ \pi^\ast} {Q}}_{\calH_k} - \varepsilon^{(2)}_{NT}\right\}\nonumber \right\},
\end{align*}
where the first inequality is due to $Q^\pi \in \Omega(\pi, \calF,  \calW, \varepsilon^{(1)}_{NT}, \varepsilon^{(2)}_{NT})$, the second inequality holds as we minimize over $Q \in \calF$ and by Assumption \ref{ass: regularity}, $Q^{\widehat \pi_{\text{dual}}} \in \calF$, the third inequality is based on the optimization property of \eqref{eqn: dual policy optimization}, and the last one holds because of the restriction on the dual variable $\rho$. By results in Lemmas \ref{lm: ac part}-\ref{lm: RKHS convergence},
we can further obtain that
\begin{align*}
	\text{Regret}(\widehat{\pi}_{\text{dual}}) 
	& \leq (1-\gamma) \EE_{S_0 \sim \nu}\left[Q^{\pi^\ast}(S_0, \pi^\ast(S_0))\right]- \max_{0 \preceq \rho \preceq C}\min_{Q \in \calF}\left\{ (1-\gamma) \EE_{S_0 \sim \nu}\left[Q(S_0, \pi^\ast(S_0))\right] \right.\\
	&\left. +\rho_1 \times \left\{\sup_{w \in \calW}  \overline\EE\left[w(S, A)\left(R + \gamma {Q}(S',  \pi^\ast(S')) - {Q}(S, A) \right)\right]  - 2\varepsilon^{(1)}_{NT}\right\}\nonumber  \right. \\ 
	& \left. +  \rho_2 \times \left\{\norm{\calT^{ \pi^\ast} {Q}}_{\calH_k} -  \varepsilon^{(2)}_{NT}\right\}\nonumber \right\},
\end{align*}
with probability at least $1 - O(1)/(NT)$. Consider the following constraint optimization problem.
\begin{align*}
	\min_{Q \in \widetilde \Omega(\pi^\ast, \calF, \varepsilon^{(1)}_{NT}, \varepsilon^{(2)}_{NT}) } \quad (1-\gamma) \EE_{S_0 \sim \nu}\left[Q(S_0, \pi^\ast(S_0))\right],
\end{align*}
where
\begin{align*}
	&\widetilde \Omega(\pi^\ast, \calF, \varepsilon^{(1)}_{NT}, \varepsilon^{(2)}_{NT}) \\
	=& \left\{Q \in \calF \mid\sup_{w \in \calW}  \overline\EE\left[w(S, A)\left(R + \gamma {Q}(S',  \pi^\ast(S')) - {Q}(S, A) \right)\right]  \leq  2\varepsilon^{(1)}_{NT}\right\}\\
	\cap & \left\{Q \in \calF \mid\norm{\calT^{ \pi^\ast} {Q}}_{\calH_k} \leq  \varepsilon^{(2)}_{NT}\right\}.
\end{align*}
It can be verified that the objective function is convex functional with respect to $Q$ and the constraint set is constructed by convex mapping of $Q$ under the Assumption \ref{ass: dual variable}. Moreover, when $Q^{\pi^\ast} \in \calF$ by Assumption \ref{ass: regularity}, the inequality is strictly satisfied due to the Bellman equation, which implies that $Q^{\pi^\ast}$ is the interior point of $\widetilde \Omega(\pi^\ast, \calF, \varepsilon_{NT})$. Lastly, by Assumption \ref{ass: function class}~\eqref{ass: Q-function class}, the objective function is always bounded below. Then by Theorem 8.6.1 of \cite{luenberger1997optimization}, strong duality holds. Therefore
\begin{align*}
	&\min_{Q \in \widetilde \Omega(\pi^\ast, \calF, \varepsilon^{(1)}_{NT}, \varepsilon^{(2)}_{NT})} \quad (1-\gamma) \EE_{S_0 \sim \nu}\left[Q(S_0, \pi^\ast(S_0))\right] \\
	=& \max_{\rho \succeq 0}\min_{Q \in \calF}\left\{ (1-\gamma) \EE_{S_0 \sim \nu}\left[Q(S_0, \pi^\ast(S_0))\right] \right.\\
	&\left. +\rho_1 \times \left\{\sup_{w \in \calW}  \overline\EE\left[w(S, A)\left(R + \gamma {Q}(S',  \pi^\ast(S')) - {Q}(S, A) \right)\right]  - 2\varepsilon^{(1)}_{NT}\right\}\nonumber  \right. \\ 
	& \left. +  \rho_2 \times \left\{\norm{\calT^{ \pi^\ast} {Q}}_{\calH_k} - \varepsilon^{(2)}_{NT} \right\}\nonumber \right\}\\
	=& \max_{\rho \succeq 0}\min_{Q \in \calF} L(Q, \rho, \pi). 
\end{align*}
In the following, we show that
$$
\max_{\rho \succeq 0}\min_{Q \in \calF} L(Q, \rho, \pi) = \max_{0 \preceq \rho \preceq C}\min_{Q \in \calF} L(Q, \rho, \pi)
$$
for some $C > 0$.
For every $\pi \in \Pi$, let $\rho^\ast(\pi)$ be the optimal dual variables, i.e.,
\begin{align}\label{def: Lagarange multipler}
	\rho^\ast(\pi) \in \argmax_{\rho \succeq 0}\min_{Q \in \calF}\,  L(Q, \rho, \pi). 
\end{align}
By the strong duality and complementary slackness, one must have that
\begin{align*}
	\rho_1^{\ast}(\pi^\ast) \times \left\{\sup_{w \in \calW}  \overline\EE\left[w(S, A)\left(R + \gamma {Q}^\ast(S',  \pi^\ast(S')) - {Q}^\ast(S, A) \right)\right]  - 2\varepsilon^{(1)}_{NT}\right\} = 0\\ 
	\rho_2^{\ast}(\pi^\ast)\times \left\{\norm{\calT^{ \pi^\ast} {Q}^\ast}_{\calH_k} - \varepsilon^{(2)}_{NT}\right\}= 0,
\end{align*}
where $Q^\ast$ is the optimal primal solution.  Since the optimal value for the primal problem is always finite duo to Assumption \ref{ass: function class}~\eqref{ass: Q-function class}, we claim that
$\rho_1^{\ast}(\pi^\ast) , \rho_2^{\ast}(\pi^\ast) $ are finite. We can show this statement by contradiction. Without loss of generality,  if $\rho_1^{\ast}(\pi^\ast) = \infty$ and $\rho_2^{\ast}(\pi^\ast)$ is finite, one must have
$$
\sup_{w \in \calW}  \overline\EE\left[w(S, A)\left(R + \gamma {Q}^\ast(S',  \pi^\ast(S')) - {Q}^\ast(S, A) \right)\right]  < 2\varepsilon^{(1)}_{NT}
$$
so that $Q^\ast$ is an optimal solution of 
\begin{align*}
	&\min_{Q \in \calF}\left\{ (1-\gamma) \EE_{S_0 \sim \nu}\left[Q(S_0, \pi^\ast(S_0))\right] \right.\\
	&\left. +\rho^\ast_1 \times \left\{\sup_{w \in \calW}  \overline\EE\left[w(S, A)\left(R + \gamma {Q}(S',  \pi^\ast(S')) - {Q}(S, A) \right)\right]  - 2\varepsilon^{(1)}_{NT}\right\}\nonumber  \right. \\ 
	& \left. +  \rho^\ast_2 \times \left\{\norm{\calT^{ \pi^\ast} {Q}}_{\calH_k} - \varepsilon^{(2)}_{NT} \right\}\nonumber \right\},
\end{align*}
with the optimal value $-\infty$. This violates the complementary slackness. Therefore, there always exists an optimal dual solution $\rho^\ast(\pi^\ast) = (\rho_1^\ast(\pi^\ast), \rho_2^{\ast}(\pi^\ast))$ such that $\max\{\rho_1^\ast(\pi^\ast), \rho_2^{\ast}(\pi^\ast)\}  \leq C$  some constant $C>0$ of the above dual problem. This further indicates that
\begin{align*}
	&\min_{Q \in \widetilde \Omega(\pi^\ast, \calF, \varepsilon^{(1)}_{NT}, \varepsilon^{(2)}_{NT})} \quad (1-\gamma) \EE_{S_0 \sim \nu}\left[Q(S_0, \pi^\ast(S_0))\right] \\
	=& \max_{0 \preceq \rho \preceq C}\min_{Q \in \calF}\left\{ (1-\gamma) \EE_{S_0 \sim \nu}\left[Q(S_0, \pi^\ast(S_0))\right] \right.\\
	&\left. +\rho_1 \times \left\{\sup_{w \in \calW}  \overline\EE\left[w(S, A)\left(R + \gamma {Q}(S',  \pi^\ast(S')) - {Q}(S, A) \right)\right]  - 2\varepsilon^{(1)}_{NT}\right\}\nonumber  \right. \\ 
	& \left. +  \rho_2 \times \left\{ \norm{\calT^{ \pi^\ast} {Q}}_{\calH_k} - \varepsilon^{(2)}_{NT}) \right\}\nonumber \right\}.
\end{align*}
By leveraging the above equality, we obtain that
\begin{align*}
	\text{Regret}(\widehat{\pi}_{\text{dual}}) 
	& \leq (1-\gamma) \EE_{S_0 \sim \nu}\left[Q^{\pi^\ast}(S_0, \pi^\ast(S_0))\right]- \min_{Q \in \widetilde \Omega(\pi^\ast, \calF, \varepsilon^{(1)}_{NT}, \varepsilon^{(2)}_{NT})} \quad (1-\gamma) \EE_{S_0 \sim \nu}\left[Q(S_0, \pi^\ast(S_0))\right] \\
	& \leq \max_{Q \in \widetilde \Omega(\pi^\ast, \calF, \varepsilon^{(1)}_{NT}, \varepsilon^{(2)}_{NT})}\left\{ (1-\gamma) \EE_{S_0 \sim \nu}\left[Q^{\pi^\ast}(S_0, \pi^\ast(S_0))\right]- (1-\gamma) \EE_{S_0 \sim \nu}\left[Q(S_0, \pi^\ast(S_0))\right] \right\}\\
	& \lesssim \varepsilon^{(1)}_{NT} +  \widetilde \lambda^{\pi^\ast}_2(\calS \times \calA) \times \text{MMD}(\bar d^{\pi^b}_T, \lambda_2^{\pi^\ast}) \times \varepsilon^{(2)}_{NT},
\end{align*}
where the last inequality is given by Lemma \ref{lm: OPE decomposition with mean embedding} and the definition of $\widetilde \Omega(\pi^\ast, \calF, \varepsilon^{(1)}_{NT}, \varepsilon^{(2)}_{NT})$. For the last part of the proof, see a similar argument in the proof of Theorem \ref{thm: regret bound for primal}. This concludes our proof.

\section{Supporting Lemmas}
\textbf{Proof of Lemma \ref{lm: OPE decomposition}}: Without loss of generality, assume $d^\pi_\gamma$ is the probability density function of the discounted visitation probability measure over $\calS \times \calA$. Then by the backward Bellman equation, we can show that for any $(s, a) \in \calS \times \calA$,
\begin{align*}
	d^\pi_\gamma(s, a) = (1-\gamma)\nu(s) \pi(a \given s) + \gamma \int_{\calS \times \calA} d^\pi_\gamma(s', a')q(s \given s', a')\pi(a \given s) ds'da',
\end{align*}
Multiplying $Q^\pi(s, a) - \widetilde{Q}(s, a)$ and integrating out over $\calS \times \calA$ on both sides  gives that
\begin{align}\label{eqn: temp 1}
	\EE_{(S, A) \sim d_\gamma^\pi} \left[Q^\pi(S, A) - \widetilde{Q}(S, A) \right] = \calV(\pi) - \widetilde{\calV}(\pi) + \gamma \EE_{(S, A) \sim d_\gamma^\pi} \left[Q^\pi(S', \pi(S')) - \widetilde{Q}(S', \pi(S')) \right].
\end{align}
By using the Bellman equation for $Q^\pi$, i.e., for every $(s, a) \in \calS \times \calA$,
\begin{align}\label{eqn: Bellman equation}
	Q^\pi(s, a) = \EE\left[R + \gamma Q^\pi(S', \pi(S')) \given S = s, A= a\right],
\end{align}
we can conclude our proof by showing that Equation \eqref{eqn: temp 1} can be simplified to that
$$
\calV(\pi) - \widetilde{\calV}(\pi)= \EE_{(S, A) \sim d_\gamma^\pi} \left[ R + \gamma \widetilde{Q}(S', \pi(S')) - \widetilde{Q}(S, A) \right].
$$

\noindent\textbf{Proof of Lemma \ref{lm: OPE decomposition with mean embedding}}: By Lebesgue's decomposition theorem, we can show that
\begin{align*}
	\calV(\pi) - \widetilde{\calV}(\pi) &= \EE_{(S, A) \sim d_\gamma^\pi} \left[ R + \gamma \widetilde{Q}(S', \pi(S')) - \widetilde{Q}(S, A) \right]\\
	& = \widetilde \lambda_1^\pi(\calS \times \calA) \times  \EE_{(S, A) \sim \lambda_1^\pi} \left[ R + \gamma \widetilde{Q}(S', \pi(S')) - \widetilde{Q}(S, A) \right]\\
	& + \widetilde \lambda_2^\pi(\calS \times \calA)  \times \EE_{(S, A) \sim \lambda_2^\pi} \left[ R + \gamma \widetilde{Q}(S', \pi(S')) - \widetilde{Q}(S, A) \right]\\
	& = \widetilde \lambda_1^\pi(\calS \times \calA)  \times \overline \EE\left[\omega^\pi(S, A) \left(R + \gamma \widetilde{Q}(S', \pi(S')) - \widetilde{Q}(S, A)\right) \right]\\
	& + \widetilde \lambda_2^\pi(\calS \times \calA)  \times \EE_{(S, A) \sim \lambda_2^\pi} \left[ R + \gamma \widetilde{Q}(S', \pi(S')) - \widetilde{Q}(S, A) \right],
\end{align*}
where the last equality uses the change of measure.
By the conditions that $\omega^\pi \in \calW$ and $\calW$ is symmetric, we can derive an upper bound for the first term in the above inequality, i.e.,
\begin{align*}
	& \overline \EE\left[\omega^\pi(S, A) \left(R + \gamma \widetilde{Q}(S', \pi(S')) - \widetilde{Q}(S, A)\right) \right]\\
	\leq & \sup_{w \in \calW} \overline \EE\left[w(S, A)\left(R + \gamma \widetilde{Q}(S', \pi(S')) - \widetilde{Q}(S, A) \right)\right]. 
\end{align*}
Furthermore, under the conditions in Lemma \ref{lm: existence of embeddings} and $\calT^\pi \widetilde{Q} \in \calH_k$, we have that
\begin{align*}
	& \left\abs{\EE_{(S, A) \sim \lambda_2^\pi} \left[ R + \gamma \widetilde{Q}(S', \pi(S')) - \widetilde{Q}(S, A) \right]\right}\\
	=& \left\abs{\langle \calT^\pi \widetilde{Q}, \mu_{\pi} \rangle_{\calH_k} \right}\\
	=& \max\left\{\langle \calT^\pi \widetilde{Q}, \mu_{\pi} \rangle_{\calH_k}, \langle -\calT^\pi \widetilde{Q}, \mu_{\pi} \rangle_{\calH_k}   \right\}\\
	\leq	& \max\left\{\sup_{\substack{\mu_{\mathbb{P}} \in \calH_k \\ \norm{\mu_{\pi^b} - \mu_{\mathbb{P}}}_{\calH_k} \leq \text{MMD}_k(\bar d^{\pi^b}_T, \lambda_2^\pi) }}  \langle \calT^\pi \widetilde{Q}, \mu_\mathbb{P}\rangle_{\calH_k}, \sup_{\substack{\mu_{\mathbb{P}} \in \calH_k\\ \norm{\mu_{\pi^b} - \mu_{\mathbb{P}}}_{\calH_k} \leq \text{MMD}_k(\bar d^{\pi^b}_T, \lambda_2^\pi)}}  \langle -\calT^\pi \widetilde{Q}, \mu_\mathbb{P} \rangle_{\calH_k}   \right\},
\end{align*}
where the first equality is based on Lemma \ref{lm: OPE decomposition}. Furthermore, by Cauchy–Schwarz inequality,  we can show that
\begin{align}
	\sup_{\substack{\mu_{\mathbb{P}} \in \calH_k \\ \norm{\mu_{\pi^b} - \mu_{\mathbb{P}}}_{\calH_k} \leq \text{MMD}_k(\bar d^{\pi^b}_T, \lambda_2^\pi) }}  \langle \calT^\pi \widetilde{Q}, \mu_\mathbb{P}\rangle_{\calH_k} & = \langle \calT^\pi \widetilde{Q}, \mu_{\pi^b} \rangle_{\calH_k} + \sup_{\substack{\mu_{\mathbb{P}} \in \calH_k \\ \norm{\mu_{\pi^b} - \mu_{\mathbb{P}}}_{\calH_k} \leq \text{MMD}_k(\bar d^{\pi^b}_T, \lambda_2^\pi) }}  \langle \calT^\pi \widetilde{Q}, \mu_\mathbb{P} - \mu_{\pi^b}\rangle_{\calH_k}\\
	& \leq \langle \calT^\pi \widetilde{Q}, \mu_{\pi^b} \rangle_{\calH_k} + \text{MMD}_k(\bar d^{\pi^b}_T, \lambda_2^\pi) \norm{\calT^\pi \widetilde{Q}}_{\calH_k},
\end{align}
where the equality in the last line holds if $\mu_{\mathbb{P}} = \mu_{\pi^b} + \text{MMD}_k(\bar d^{\pi^b}_T, \lambda_2^\pi) \calT^\pi \widetilde{Q} / \norm{\calT^\pi \widetilde{Q}}_{\calH_k}$ for $\calT^\pi \widetilde{Q} \neq 0$ or $\calT^\pi \widetilde{Q}$ = 0. Therefore we can conclude that
$$
\sup_{\substack{\mu_{\mathbb{P}} \in \calH_k \\ \norm{\mu_{\pi^b} - \mu_{\mathbb{P}}}_{\calH_k} \leq \text{MMD}_k(\bar d^{\pi^b}_T, \lambda_2^\pi) }}  \langle \calT^\pi \widetilde{Q}, \mu_\mathbb{P}\rangle_{\calH_k}  =\langle \calT^\pi \widetilde{Q}, \mu_{\pi^b} \rangle_{\calH_k} + \text{MMD}_k(\bar d^{\pi^b}_T, \lambda_2^\pi)\norm{\calT^\pi \widetilde{Q}}_{\calH_k}.
$$
By a similar argument, we can show that
$$
\sup_{\substack{\mu_{\mathbb{P}} \in \calH_k \\ \norm{\mu_{\pi^b} - \mu_{\mathbb{P}}}_{\calH_k} \leq \text{MMD}_k(\bar d^{\pi^b}_T, \lambda_2^\pi) }}  \langle -\calT^\pi \widetilde{Q}, \mu_\mathbb{P}\rangle_{\calH_k}  =\langle -\calT^\pi \widetilde{Q}, \mu_{\pi^b} \rangle_{\calH_k} + \text{MMD}_k(\bar d^{\pi^b}_T, \lambda_2^\pi)\norm{\calT^\pi \widetilde{Q}}_{\calH_k},
$$
and note that
$$
\langle \calT^\pi \widetilde{Q}, \mu_{\pi^b} \rangle_{\calH_k} = \overline \EE\left[\left(R + \gamma \widetilde{Q}(S', \pi(S')) - \widetilde{Q}(S, A) \right)\right]. 
$$

Summarizing two upper bounds together gives that for every $\pi$
\begin{align*}
	&\left\abs{\calV(\pi) - \widetilde{\calV}(\pi)  \right}\\
	\leq &\widetilde \lambda_1^\pi(\calS \times \calA)  \times \left\abs{ \overline \EE\left[\omega^\pi(S, A) \left(R + \gamma \widetilde{Q}(S', \pi(S')) - \widetilde{Q}(S, A)\right) \right]\right}\\
	+ &  \widetilde \lambda_2^\pi(\calS \times \calA)  \times \left\abs{\EE_{(S, A) \sim \lambda_2^\pi} \left[ R + \gamma \widetilde{Q}(S', \pi(S')) - \widetilde{Q}(S, A) \right]\right}\\
	\leq & \widetilde \lambda_1^\pi(\calS \times \calA)  \times  \sup_{w \in \calW} \overline \EE\left[w(S, A)\left(R + \gamma \widetilde{Q}(S', \pi(S')) - \widetilde{Q}(S, A) \right)\right] \\
 + & \left\abs{\overline \EE\left[\left(R + \gamma \widetilde{Q}(S', \pi(S')) - \widetilde{Q}(S, A)\right) \right]\right}
	+   \widetilde \lambda^\pi_2(\calS \times \calA) \text{MMD}_k(\bar d^{\pi^b}_T, \lambda_2^\pi) \norm{\calT^\pi \widetilde{Q}}_{\calH_k}.
\end{align*}

\begin{lemma}\label{lm: ac part}
	Under Assumptions \ref{ass: Markovian}-\ref{ass: stationary} and \ref{ass: function class}, we have with probability at least $1- \frac{1}{NT}$, for every $Q \in \calF, w \in \calW$ and $\pi \in \Pi$,
	\begin{align*}
		&\left\abs{\overline \EE\left[w(S, A)\left(R + \gamma Q(S', \pi(S')) - Q(S, A) \right)\right] - \overline\EE_{NT}\left[w(S, A)\left(R + \gamma {Q}(S', \pi(S')) - {Q}(S, A) \right)\right] \right} \\
		\lesssim  &\log(NT)\sqrt{\left(v(\Pi) +  v(\calF)   \right)/NT}.
	\end{align*}
\end{lemma}
\begin{proof}
	We apply Lemma \ref{lm: empirical process for scalar} to show the statement. Define
	\begin{align*}
		\calG_1 = \left\{(s, a, s') \rightarrow w(s, a)\left(\widetilde R(s, a, s') + \gamma Q(s', \pi(s')) - Q(s, a)\right) \,\mid \, \pi \in \Pi, Q \in \calF, w \in \calW   \right\}. 
	\end{align*}
	By Assumption \ref{ass: function class}, we can show that 
	$$
	\log \calN(\epsilon, \calG_1, \norm{\bullet}_\infty) \lesssim( v(\Pi) + v(\calF)) \log(1/\epsilon).
	$$
	Then by applying Lemma \ref{lm: empirical process for scalar}, we can show that with probability at least $1 - 1/(NT)$, 
	\begin{align*}
		& \left\abs{\overline \EE\left[\left(R + \gamma Q(S', \pi(S')) - Q(S, A) \right)\right] - \overline \EE_{NT}\left[\left(R + \gamma Q(S', \pi(S')) - Q(S, A) \right)\right]\right}\\
		\lesssim &\log(NT)\sqrt{\left( v(\Pi) +  v(\calF) \right)/NT}.
	\end{align*}
\end{proof}

\textbf{Proof of Lemma \ref{lm: complexity for calG}}: Recall that we define the following class of vector-valued functions as
\begin{align*}
	\calG & =  \left\{g: \calS \times \calA \times \calS \rightarrow \calH_k \, \mid \,  g(s, a, s') = \left(Y(Q, \pi, s, a, s') -  \calT^\pi_{\zeta_{NT}}Q(s, a)\right)k((s, a), \bullet) \right.\\
	& \left. \, \text{with} \quad Q \in \calF, \pi \in \Pi  \right\},
\end{align*}
where $Y(Q, \pi, s, a, s') = \widetilde{R}(s, a, s') + \gamma Q(s', \pi(s')) - Q(s, a)$.
Next, we show that for any $g \in \calG$, $\sup_{(s, a, s') \in \calS \times \calA \times \calS}\sup_{g \in \calG}\norm{g(s, a, s')}_{\calH_k} < + \infty$, i.e., the envelop function of $\calG$ is uniformly bounded above. Since for any $Q \in \calF$, $\norm{Q}_\infty \leq  c_\calF$ by Assumption \ref{ass: function class}~\eqref{ass: Q-function class}, we can show that for any $(s, a, s') \in \calS \times \calA \times \calS$,
\begin{align*}
	\sup_{g \in \calG}\norm{g(s, a, s')}^2_{\calH_k} \leq &  \sup_{Q \in \calF, \pi \in \Pi} \left(Y(Q, \pi, s, a, s') -  \calT^\pi_{\zeta_{NT}}Q(s, a)\right)^2 K((s, a), (s, a))\\
	\leq & 2\left(R_{\max} + (1+\gamma)c_\calF\right)^2\kappa^2 \\
	+ & 2\left(\sup_{\pi \in \Pi, Q \in \calF}\norm{\calT^\pi Q}_{\calH_k} +\sup_{\pi \in \Pi, Q \in \calF}\norm{\calL_k^{-c}\calT^\pi Q}_{L^2_{d^{\pi_b}}} \right)^2\kappa^3  \triangleq c^2_\calG,
\end{align*}
where the second inequality is based on Assumption \ref{ass: reward} and the following argument.
\begin{align*}
	\norm{\calT^\pi_{\zeta_{NT}}Q}_\infty & \leq \kappa \norm{\calT^\pi_{\zeta_{NT}}Q}_{\calH_k} \leq \kappa \sup_{\pi \in \Pi, Q \in \calF}\norm{\calT^\pi Q}_{\calH_k} + \kappa\zeta_{NT}^{c - 1/ 2} \sup_{\pi \in \Pi, Q \in \calF}\norm{\calL_k^{-c}\calT^\pi Q}_{L^2_{d^{\pi_b}}}\\
	& \leq \kappa \left(\sup_{\pi \in \Pi, Q \in \calF}\norm{\calT^\pi Q}_{\calH_k} + \sup_{\pi \in \Pi, Q \in \calF}\norm{\calL_k^{-c}\calT^\pi Q}_{L^2_{d^{\pi_b}}}\ \right),
\end{align*}
where the second inequality is based on the condition that $\zeta_{NT} \lesssim 1$ and $c > 1/2$.
In the following,  we calculate the uniform entropy of $\calG$ with respect to any empirical probability measure, i.e., $
\sup_{\mathbb{P}} \log(\calN(\epsilon c_\calG, \calG, \norm{\bullet}_{\calH_k, \infty}))$. 
Then, for any $\epsilon > 0$, it can be seen that for any $(s, a, s') \in \calS \times \calA \times \calS$,  $\pi_1, \pi_2 \in \Pi$ and $ Q_1, Q_2 \in \calF$,
{\small
	\begin{align*}
		& \norm{\left(Y(Q_1, \pi_1, s, a, s') -  \calT^{\pi_1}_{\zeta_{NT}}Q_1(s, a)\right)k((s, a), \bullet) - \left(Y(Q_2, \pi_2, s, a, s') -  \calT^{\pi_2}_{\zeta_{NT}}Q_2(s, a)\right)k((s, a), \bullet)}_{\calH_k}\\
		& \leq \kappa \abs{Y(Q_1, \pi_1, s, a, s') -Y(Q_2, \pi_2, s, a, s') } + \kappa \abs{\calT^{\pi_1}_{\zeta_{NT}}Q_1(s, a) -\calT^{\pi_2}_{\zeta_{NT}}Q_2(s, a)}\\
		& \lesssim \kappa\left( \norm{Q_1 - Q_2}_\infty + \norm{\norm{\pi_1 - \pi_2}_\infty}_{\ell_1} \right) + \kappa \abs{\calT^{\pi_1}_{\zeta_{NT}}Q_1(s, a) -\calT^{\pi_1}_{\zeta_{NT}}Q_2(s, a)}+ \kappa\abs{\calT^{\pi_1}_{\zeta_{NT}}Q_2(s, a) -\calT^{\pi_2}_{\zeta_{NT}}Q_2(s, a)}\\
		& \leq \kappa\left( \norm{Q_1 - Q_2}_\infty + \norm{\norm{\pi_1 - \pi_2}_\infty}_{\ell_1} \right) + \kappa \left\abs{\calT^{\pi_1}_{\zeta_{NT}}(Q_1-Q_2)(s, a)\right}+ \kappa\abs{\calT^{\pi_1}_{\zeta_{NT}}Q_2(s, a) -\calT^{\pi_2}_{\zeta_{NT}}Q_2(s, a)},
	\end{align*}
}where the first inequality is based on Assumption \ref{ass: kernel}, the second one is given by Lipschitz condition stated in  Assumption \ref{ass: function class}~\eqref{ass: Q-function class}, and the last one is due to the linearity of the operator $\calT_{\zeta_{NT}}^\pi$. For the second term in the above inequality, we can show that
\begin{align*}
	\left\abs{\calT^{\pi_1}_{\zeta_{NT}}(Q_1-Q_2)(s, a)\right} & \leq \kappa \norm{\calT^{\pi_1}_{\zeta_{NT}}(Q_1-Q_2)}_{\calH_k}\\
	& \leq \kappa \norm{\calT^{\pi_1}(Q_1-Q_2)}_{\calH_k}\\
	& =  \norm{\calL^{c-1/2}_k(g_{\pi, Q_1} - g_{\pi, Q_2})}_2\\
	& \lesssim  \norm{g_{\pi, Q_1} - g_{\pi, Q_2})}_2\\
	& \lesssim  \norm{Q_1-Q_2}_\infty,
\end{align*}
where the first inequality is based on Assumption \ref{ass: kernel}. The second and third lines hold because of Assumption \ref{ass: regularity} and the spectral decomposition of $\calL_k$. The last inequality holds because of the Lipschitz condition in Assumption \ref{ass: Lips for covering}. The third inequality is based on the following argument.
\begin{align*}
	\norm{\calL^{c-1/2}_k\left(g_{\pi, Q_1} - g_{\pi, Q_2}\right)}^2_2 &= \sum_{i=1}^{+\infty}e_i^{2c-1}e^2_{i, \pi, Q_1, Q_2}\\
	& \lesssim \sum_{i=1}^{+\infty}e^2_{i, \pi, Q_1, Q_2} =  \norm{g_{\pi, Q_1} - g_{\pi, Q_2}}_2,
\end{align*}
where the inequality holds as $\{e_i\}_{i \geq 1}$ is non-decreasing towards $0$ and $c > 1/2$. Now we focus on the last term $\abs{\calT^{\pi_1}_{\zeta_{NT}}Q_2(s, a) -\calT^{\pi_2}_{\zeta_{NT}}Q_2(s, a)}$. Similar as before, we can show that
\begin{align*}
	\left\abs{\calT^{\pi_1}_{\zeta_{NT}}Q_2(s, a) -\calT^{\pi_2}_{\zeta_{NT}}Q_2(s, a)\right}
	& \leq  \kappa \norm{(\calL_k + \zeta_{NT}I )^{-1}\calL_k(\calT^{\pi_1}Q_2 - \calT^{\pi_2}Q_2)}_{\calH_k}\\
	& \lesssim  \norm{\calT^{\pi_1}Q_2 - \calT^{\pi_2}Q_2}_{\calH_k}\\
	& =  \norm{\calL^{c-1/2}_kg_{\pi_1, Q_2} - \calL^{c-1/2}_kg_{\pi_2, Q_2}}_2\\
	& \lesssim  \norm{g_{\pi_1, Q_2}  - g_{\pi_2, Q_2} }_2\\
	& \lesssim  \norm{\norm{\pi_1 - \pi_2}_\infty}_{\ell_1},
\end{align*}
where we use Assumptions \ref{ass: regularity}-\ref{ass: Lips for covering} again.

Therefore, we can show that
$$
\log(\calN(\epsilon c_\calG, \calG, \norm{\bullet}_{\calH_k, \infty})) \lesssim \log \calN(\epsilon , \calF,  \norm{\bullet}_\infty) + \sum_{i = 1}^{d_A}\log \calN(\epsilon , \Pi_i,  \norm{\bullet}_\infty) \asymp \left(v(\Pi) +  v(\calF)  \right)\log(1/\epsilon).
$$

\begin{lemma}\label{lm: RKHS convergence}
	Under Assumptions \ref{ass: Markovian}-\ref{ass: function class}~\eqref{ass: Q-function class} \eqref{ass: vector-valued function class}, and \ref{ass: regularity}, we have with probability at least $1- \frac{1}{NT}$, for every $Q \in \calF$ and $\pi \in \Pi$,
	$$
	\norm{\calT^\pi Q - \widehat{\calT}^\pi Q}_{\calH_k} \lesssim \left(\log(NT)\sqrt{\left(v(\Pi) +  v(\calF) + v(\calG)  \right)/NT}\right)^{\frac{2c-1}{2c+1}}.
	$$
\end{lemma}
\begin{proof}
Recall that we define an operator $T^\pi_{\zeta_{NT}}: \calF \rightarrow \calH_k$ such that
\begin{align}\label{def: bias term oracle RKHS}
	T^\pi_{\zeta_{NT}}Q = \argmin_{f \in \calH_k} \overline{\EE}\left[\left(R + \gamma Q(S', \pi(S')) - Q(S, A) - f(S, A)\right)^2\right] + \zeta_{NT} \norm{f}_{\calH_k}^2,
\end{align}
for every $Q \in \calF$.
It can be seen that
$$
T^\pi_{\zeta_{NT}}Q  = \left(\calL_k + \zeta_{NT}\calI \right)^{-1}\calL_k\left(\calT^\pi Q\right),
$$
where $\calI$ is the identity operator.
In order to bound $\sup_{\pi \in \Pi, Q \in \calF} \norm{\widehat \calT^\pi Q - \calT^\pi Q}_{\calH_k}$, we first decompose it as
\begin{align*}
	& \sup_{\pi \in \Pi, Q \in \calF} \norm{\widehat \calT^\pi Q - \calT^\pi Q}_{\calH_k}\\
	\leq &\underbrace{\sup_{\pi \in \Pi, Q \in \calF} \norm{\widehat \calT^\pi Q -T^\pi_{\zeta_{NT}} Q}_{\calH_k}}_{\text{Term (I) }} + \underbrace{\sup_{\pi \in \Pi, Q \in \calF} \norm{\calT^\pi Q -T^\pi_{\zeta_{NT}} Q}_{\calH_k}}_{\text{Term (II)}}.
\end{align*}

We then derive an upper bound for Term (II), which is the ``bias" term for estimating $ \calT^\pi Q $. We follow the proof of Theorem 4 in \cite{smale2005shannon} and extend their pointwise results to the uniform covergence over $\Pi$ and $\calF$. Specifically, under Assumption \ref{ass: regularity}, for every $\pi \in \Pi$ and $Q \in \calF$, there exists $g_{\pi, Q}$ such that $\calL_k^c g_{\pi, Q} = \calT^\pi Q$. Recall that 
$$
g_{\pi, Q} = \sum_{i = 1}^{\infty} e_{i, \pi, Q}\phi_i \quad \text{with} \quad \norm{\{e_{i, \pi, Q}\}_{i \geq 1}}_{\ell_2} = \norm{g_{\pi, Q}}_{L^2_{d^{\pi_b}}} < +\infty
$$
and thus
$$
\calT^\pi Q = \sum_{i = 1}^{\infty} e_i^{c}e_{i, \pi, Q}\phi_i\quad \text{with} \quad \norm{\{e_{i, \pi, Q}\}_{i \geq 1}}_{\ell_2} < +\infty.
$$
By the definition of $T^\pi_{\zeta_{NT}} Q$, we can show that for every $\pi \in \Pi$ and $Q \in \calF$,
\begin{align*}
	\calT^\pi_{\zeta_{NT}} Q - \calT^\pi Q &= 
	\left(\calL_k + \zeta_{NT}\calI \right)^{-1}\calL_k\left(\calT^\pi Q\right) - \calT^\pi Q\\
	& = -\sum_{i = 1}^{+\infty} \frac{\zeta_{NT}}{\zeta_{NT} + e_i}e_i^{c}e_{i, \pi, Q}\phi_i.
\end{align*}
Then for $1/2 < c \leq 3/2$, we have
\begin{align*}
	\left\norm{ \calT^\pi_{\zeta_{NT}} Q - \calT^\pi Q\right}^2_{\calH_k} &= \sum_{i = 1}^{+\infty} \left(\frac{\zeta_{NT}}{\zeta_{NT} + e_i}e_i^{c-1/2}e_{i, \pi, Q}\right)^2\\
	&= \zeta_{NT}^{2c - 1}\sum_{i = 1}^{+\infty} \left(\frac{\zeta_{NT}}{\zeta_{NT} + e_i}\right)^{3-2c}\left(\frac{e_i}{\zeta_{NT}+e_i}\right)^{2c-1}e_{i, \pi, Q}^2\\
	&\leq \zeta_{NT}^{2c - 1}\sum_{i = 1}^{+\infty} e_{i, \pi, Q}^2,
\end{align*}
which implies that for any $\pi \in \Pi$ and $Q \in \calF$,
$$
\norm{\calT^\pi Q -T^\pi_{\zeta_{NT}} Q}_{\calH_k} \leq \zeta_{NT}^{c - 1/2}\norm{\calL_k^{-c}\calT^\pi Q}_{L^2_{d^{\pi_b}}},
$$
which further gives that
$$
\text{Term (II)} \leq \zeta_{NT}^{c - 1/2}\sup_{\pi \in \Pi, Q \in \calF}\norm{\calL_k^{-c}\calT^\pi Q}_{L^2_{d^{\pi_b}}}.
$$

Next, we derive an upper bound for the ``variance" of $\widehat \calT^\pi Q $, i.e., Term (I). Following the result of \cite{smale2007learning}, we define the sampling operator $U_{\bm z}: \calH_k \rightarrow \mathbb{R}^{NT}$ with batch data $\{Z_{i, t} = (S_{i, t}, A_{i, t} )\}_{1 \leq i \leq N, 0 \leq t < T}$ over $\calZ = \calS \times \calA$ as
$$
U_{\bm z}(f) = \{ f(Z_{i, t}) \}_{1 \leq i \leq N, 0 \leq t < T} \in \mathbb{R}^{NT}
$$
for any function $f$ defined over $\calZ$.
The adjoint of the sampling operator is defined as $U_{\bm z}^\top: \mathbb{R}^{NT} \rightarrow \calH_k$ such that
$$
U_{\bm z}^\top \theta = \sum_{i = 1}^N\sum_{t = 0}^{T-1} \theta_{i, t} k({Z_{i, t}}, \bullet), \quad \text{with} \quad \theta \in \mathbb{R}^{NT}.
$$
Then we have
$$
\widehat \calT^\pi Q = \left(\frac{1}{NT}U_{\bm z}^\top U_{\bm z} + \zeta_{NT} I_{NT}  \right)^{-1} \frac{1}{NT} U_{\bm z}^\top Y_{Q, \pi},
$$
where
$$
Y_{Q, \pi} = \left\{ Y_{i, t}(Q, \pi) =  R_{i, t} + \gamma Q(S_{i, t+1}, \pi(S_{i, t+1})) - Q(S_{i, t}, A_{i, t})   \right\}_{1 \leq i \leq N, 0 \leq t < T} \in \mathbb{R}^{NT},
$$
and $I_{NT}$ is an identity matrix with the dimension $NT$.
It can be seen that for any $Q \in \calF$ and $\pi \in \Pi$,
\begin{align*}
	& \widehat \calT^\pi Q - \calT^\pi_{\zeta_{NT}} Q\\
	= & \left(\frac{1}{NT}U_{\bm z}^\top U_{\bm z} + \zeta_{NT} I_{NT}  \right)^{-1} \left(\frac{1}{NT}U_{\bm z}^\top Y_{Q, \pi} - \frac{1}{NT}U_{\bm z}^\top U_{\bm z}\left(\calT^\pi_{\zeta_{NT}}Q \right) -\zeta_{NT} \calT^\pi_{\zeta_{NT}}Q  \right)\\
	= & \left(\frac{1}{NT}U_{\bm z}^\top U_{\bm z} + \zeta_{NT} I_{NT}  \right)^{-1} \left( \overline\EE_{NT}\left[\left(Y_{Q, \pi} - \calT^\pi_{\zeta_{NT}}Q(S, A) \right) k(Z, \bullet)\right]  -\calL_k(\calT^\pi Q- \calT^\pi_{\zeta_{NT}}Q)  \right),
\end{align*}
where the last equation is based on the definition of the sampling operator, its adjoint and the closed form of $ \calT^\pi_{\zeta_{NT}}Q$. Now we can show that
\begin{align*}
	&\norm{\widehat \calT^\pi Q - \calT^\pi_{\zeta_{NT}} Q}_{\calH_k}\\
	\leq & \frac{1}{\zeta_{NT}} \norm{ \overline \EE_{NT}\left[\left(Y_{Q, \pi} - \calT^\pi_{\zeta_{NT}}Q(S, A) \right) k(Z, \bullet)\right]  -\calL_k(\calT^\pi Q- \calT^\pi_{\zeta_{NT}}Q) }_{\calH_k}.
\end{align*}
Note that  for any $(T-1) \geq t \geq 0$ and $1 \leq i \leq N$, by Assumption \ref{ass: stationary},
$$
\EE\left[\left(Y_{i, t}(Q, \pi)  - \calT^\pi_{\zeta_{NT}}Q(S_{i, t}, A_{i, t}) \right) k(Z_{i, t}, \bullet)\right] = \calL_k(\calT^\pi Q- \calT^\pi_{\zeta_{NT}}Q).
$$
To bound the above empirical process for vector-valued functions, we leverage the result developed in Lemma \ref{lm: empirical process for vector-valued}. 
By Lemma \ref{lm: empirical process for vector-valued}, we can show that with probability at least $1 - \frac{1}{NT}$,
\begin{align*}
	&\norm{\left( \EE_{NT}\left[\left(Y_{Q, \pi} - \calT^\pi_{\zeta_{NT}}Q(S, A) \right) k(Z, \bullet)\right]  -\calL_k(\calT^\pi Q- \calT^\pi_{\zeta_{NT}}Q)  \right)}_{\calH_k}\\ \lesssim &  \log(NT)\sqrt{\left(v(\Pi) +  v(\calF) + v(\calG)\right)/NT}.
\end{align*}

Summarizing Terms (I) and (II) gives that
\begin{align*}
	& \sup_{\pi \in \Pi, Q \in \calF} \norm{\widehat \calT^\pi Q - \calT^\pi Q}_{\calH_k} \\
	\lesssim & \zeta_{NT}^{-1}\log(NT)\sqrt{\left(v(\Pi) +  v(\calF) + v(\calG \right)/NT} + \zeta_{NT}^{c - 1/2}\sup_{\pi \in \Pi, Q \in \calF}\norm{\calL_k^{-c}\calT^\pi Q}_{L^2_{d^{\pi_b}}}.
\end{align*}
By choosing $\zeta_{NT} \asymp \left(\log(NT)\sqrt{\left(v(\Pi) +  v(\calF) + v(\calG)  \right)/NT} \right)^{2/(2c+1)}$, we optimize the right-hand-side of the above inequality and obtain that
\begin{align*}
	\sup_{\pi \in \Pi, Q \in \calF} \norm{\widehat \calT^\pi Q - \calT^\pi Q}_{\calH_k} & \lesssim\sup_{\pi \in \Pi, Q \in \calF}\norm{\calL_k^{-c}\calT^\pi Q}^{2/(2c+1)}_{L^2_{d^{\pi_b}}} \left(\log(NT)\sqrt{\left(v(\Pi) +  v(\calF)  \right)/NT}\right)^{\frac{2c-1}{2c+1}}.
\end{align*}
\end{proof}

\section{Additional Definitions and Supporting Lemmas}\label{app: technical lemmas}
We define the $\epsilon$-covering number below, which is used in the main text.
\begin{definition}[$\epsilon$-covering number]\label{def: covering}
	An $\epsilon$-cover of a set $\Theta$ with respect to some semi-metric $ \widetilde d$ is a set of finite elements $\{\theta_i\}_{i\geq 1} \subseteq \Theta$ such that for every $\theta \in \Theta$, there exists $\theta_j$ such that $\widetilde d(\theta_j, \theta) \leq \epsilon$.   An $\epsilon$-covering number of a set $\Theta$ denoted by $\calN(\epsilon, \Theta, \widetilde d)$ is the infimum of the cardinality of $\epsilon$-cover of $\Theta$.
\end{definition}

The following lemma provides a sufficient condition for the existence of mean embeddings for both $\lambda_2^\pi$ and $\bar d^{\pi^b}_T$. For notation simplicity, let $Z = (S, A)$ and denote $\calZ = \calS \times \calA$.
\begin{lemma}\label{lm: existence of embeddings}
	If the kernel $k(\bullet, \bullet): \calZ \times \calZ \rightarrow \mathbb{R}$ is measurable with respect to  both $\lambda_2^\pi$ and $\bar d^{\pi^b}_T$ and $\max\{\overline{\EE}[\sqrt{k(Z, Z)}], \EE_{Z \sim \lambda_2^\pi}[\sqrt{k(Z, Z)}] \} < +\infty$, then there exist mean embeddings $\mu_{\pi^b}, \mu_\pi  \in \calH_k$ such that for any $f \in \calH_k$
	\begin{align*}
		\overline{\EE}\left[f(Z)\right] = \langle \mu_{\pi^b}, f  \rangle  \quad \text{and} \quad \EE_{Z \sim \lambda_2^\pi}\left[f(Z)\right] = \langle \mu_{\pi}, f  \rangle. 
	\end{align*}
\end{lemma}
The proof of Lemma \ref{lm: existence of embeddings} can be found in Lemma 3 of \cite{gretton2012kernel}. Note that if there exists a positive constant $\delta$ such that $\text{MMD}_k(\bar d^{\pi^b}_T, \lambda_2^\pi) \leq \delta$, then under the conditions in Lemma \ref{lm: existence of embeddings}, we can show that the mean embeddings $\mu_{\pi^b}$ and $\mu_{\pi}$ must satisfy that
\begin{align}\label{eqn: embeding inequality}
	\norm{\mu_{\pi^b} - \mu_{\pi}}_{\calH_k} \leq \delta.
\end{align}
See Lemma 4 of \cite{gretton2012kernel} for more details.  Equation \eqref{eqn: embeding inequality} motivates us to bound the singular part of the OPE error $\abs{\calV(\pi) - \widetilde{\calV}(\pi)}$ by its worst-case performance over all the mean embeddings that are within $\delta$-distance to $\mu_{\pi^b}$.
\vspace{0.5cm}

\begin{lemma}\label{lm: empirical process for scalar}
	Let $\{ \{Z_{i, t}\}_{0 \leq t < T} \}_{1 \leq i \leq N}$ be i.i.d. copies of stochastic process $\{Z_t\}_{t \geq 0}$. Suppose $\{Z_t\}_{t \geq 0}$ is a stationary and exponential $\beta$-mixing process with $\beta$-mixing coefficient $\beta(q) \leq \beta_0 \exp(-\beta_1q)$ for some $\beta_0 \geq 0$ and $\beta_1 > 0$. Let $\calG$ be  a class of measurable functions that take $Z_t$ as input. For any $g \in \calG$, assume $\EE[g(Z_t)] = 0$ for any $t \geq 0$. Suppose the envelop function of $\calG$ is uniformly bounded by some constant $C > 0$. In addition, if $\calG$ belongs to the class of VC-typed functions such that $\calN(\epsilon, \calG, \norm{\bullet}_\infty) \lesssim (1/\epsilon)^\alpha$. Then with probability at least $1 - 1/(NT)$,
	$$
	\sup_{g \in \calG}\abs{\frac{1}{NT}\sum_{i = 1}^N\sum_{t = 0}^{T-1}g(Z_{i, t})} \lesssim \log(NT) \sqrt{\frac{\alpha}{NT}}.
	$$
\end{lemma}

\begin{proof}
	To prove the lemma, it is sufficient to bound $\sup_{g \in \calG}\abs{\sum_{i = 1}^N\sum_{t = 0}^{T-1}g(Z_{i, t})}$. In the following,  we apply Berbee's coupling lemma \citep{berbee1979random} and follow the remark below Lemma 4.1 of \cite{dedecker2002maximal}. Specifically, Let $q$ be some positive integer. We can always construct a sequence $\{\widetilde{Z}_{i, t} \}_{t\geq 0}$ such that with probability at least $1 - (NT\beta(q))/q$,
	$$
	\sup_{g \in \calG}\abs{\sum_{i = 1}^N\sum_{t = 0}^{T-1}g(Z_{i, t})} = \sup_{g \in \calG}\abs{\sum_{i = 1}^N\sum_{t = 0}^{T-1}g(\widetilde Z_{i, t})},
	$$
	and meanwhile the block sequence $\widetilde X_{i, k}(g) = \{g(\widetilde{Z}_{i, (k-1)q + j})\}_{0 \leq j < q}$ are identically distributed for $k \geq 1$ and $i\geq1$.  In addition, the sequence $\{  \widetilde X_{i, k}(g) \, \mid \, k = 2\omega, \omega \geq 1  \}$ are independent and so are the sequence $\{  \widetilde X_{i, k}(g) \, \mid \, k = 2\omega + 1, \omega \geq 0  \}$. 
	Let $I_r =\{\floor{T/q}q, \cdots, T-1\}$ with $\text{Card}(I_r) < q$. Then we can show that with probability at least $1 - (NT\beta(q))/q$,
	\begin{align*}
		&\sup_{g \in \calG}\abs{\sum_{i = 1}^N\sum_{t = 0}^{T-1}g(Z_{i, t})} \\
		\leq & \sup_{g \in \calG}\abs{\sum_{i = 1}^N\sum_{t = 0}^{q\floor{T/q} - 1}g(\widetilde Z_{i, t})} +  \sup_{g \in \calG}\abs{\sum_{i = 1}^N\sum_{t \in I_r}g(Z_{i, t})}. 
	\end{align*}
	In the following, assuming that the above inequality holds, we bound each of the above two terms separately. First of all, without loss of generality, we assume $\floor{T/q}$ is an even number. Then for the first term, we have
	\begin{align*}
		& \sup_{g \in \calG}\abs{\sum_{i = 1}^N\sum_{t = 0}^{q\floor{T/q} - 1}g(\widetilde Z_{i, t})} \\
		\leq & \sum_{j = 0}^{2q -1}  \sup_{g \in \calG}\abs{\sum_{i = 1}^N\sum_{k = 0}^{\floor{T/q}/2 - 1}g(\widetilde Z_{i, 2kq + j})}.
	\end{align*}
	By the construction, 	$\sup_{g \in \calG}\abs{\sum_{i = 1}^N\sum_{k = 0}^{\floor{T/q}/2-1}g(\widetilde Z_{i, 2kq + j})}$ is a suprema empirical process of i.i.d. sequences. Then by conditions in Lemma \ref{lm: empirical process for scalar} and Mcdiarmid’s inequality, we have with probability at least $1- \varepsilon$,
	$$
	\sup_{g \in \calG}\abs{\sum_{i = 1}^N\sum_{k = 0}^{\floor{T/q}/2 - 1}g(\widetilde Z_{i, 2kq + j})}\lesssim \EE\left[\sup_{g \in \calG}\left\abs{\sum_{i = 1}^N\sum_{k = 0}^{\floor{T/q}/2 - 1}g(\widetilde Z_{i, 2kq + j})\right}\right] + \sqrt{\frac{NT\log(1/\varepsilon)}{q}}.
	$$
	
	Given the condition that
	\begin{align*}
		\calN(\epsilon, \calG, \norm{\bullet}_{\infty}) \lesssim (1/ \epsilon)^\alpha.
	\end{align*}
	By a standard maximal inequality using uniform entropy integral \citep[e.g.,][]{van2011local}, we can show that with probability at least $1- \varepsilon$,
	$$
	\EE\left[\sup_{g \in \calG}\abs{\sum_{i = 1}^N\sum_{k = 0}^{\floor{T/q}/2 - 1}g(\widetilde Z_{i, 2kq + j})}\right] \lesssim \sqrt{\frac{\alpha NT}{q}}.
	$$
	By letting $\varepsilon = 1 /(NT)$, we can show that with probability at least $1 - 1/(NT)$,
	$$
	\sup_{g \in \calG}\abs{\sum_{i = 1}^N\sum_{k = 0}^{\floor{T/q}/2 - 1}g(\widetilde Z_{i, 2kq + j})}\lesssim \sqrt{\frac{\alpha NT}{q}} +\sqrt{\frac{NT\log(NT)}{q}}.
	$$
	
	Next, we bound $\sup_{g \in \calG}\abs{\sum_{i = 1}^N\sum_{t \in I_r}g(Z_{i, t})}$. Since for $i \geq 1$, $\sum_{t \in I_r}g(Z_{i, t})$ are i.i.d. sequences. Then by a similar argument as before, we can show that with probability at least $1 - 1/(NT)$
	$$
	\sup_{g \in \calG}\abs{\sum_{i = 1}^N\sum_{t \in I_r}g(Z_{i, t})} \lesssim q \left( \sqrt{\alpha N} +  \sqrt{N\log(NT)} \right).
	$$
	Without loss of generality, we assume that $\log(NT) \leq T$. Otherwise, we can apply a standard maximal inequality on $	\sup_{g \in \calG}\abs{\frac{1}{N}\sum_{i = 1}^N(\sum_{t = 0}^{T-1}g(Z_{i, t})/T)}$ by treating it as $N$ i.i.d. sequences for obtaining the desirable result.
	Summarizing together and by letting $q \asymp \log(NT)$, we can show that  with probability at least $1 - 1/(NT)$, 
	\begin{align*}
		&\sup_{g \in \calG}\abs{\sum_{i = 1}^N\sum_{t = 0}^{T-1}g(Z_{i, t})} \\
		\lesssim & \log(NT) \sqrt{\frac{\alpha NT}{\log(NT)}}  + \log(NT)  \sqrt{\frac{NT\log(NT)}{\log(NT)}} + \log(NT) \left( \sqrt{\alpha N} +  \sqrt{N\log(NT)} \right)\\
		\lesssim & \log(NT)\sqrt{NT\alpha},
	\end{align*}
	where the last inequality uses $\log(NT)\leq T$. This concludes our proof via dividing both sides by $NT$.
\end{proof}

\bigskip

Before presenting Lemma \ref{lm: empirical process for vector-valued}, let $\calG$ be  a class of vector-valued functions that take values in $\calZ$ as input and output an element in $\calH_k$. Then the metric related to the covering number $\calN(\epsilon c_\calG, \calG, L^2(\mathbb{P}))$ is defined as
$$
\left(\EE_{\mathbb{P}}(\norm{g_1(Z)- g_2(Z)}^2_{\calH_k})\right)^{1/2}.
$$
Here we suppose that the envelop function of $\calG$, defined as $G(z) = \sup_{g \in \calG}\norm{g(z)}_{\calH_k}$, is uniformly bounded by some constant $c_{\calH_k}$, i.e., $\norm{G}_\infty \leq c_{\calH_k}$.

\begin{lemma}\label{lm: empirical process for vector-valued}
	Let $\{ \{Z_{i, t}\}_{0 \leq t < T} \}_{1 \leq i \leq N}$ be i.i.d. copies of stochastic process $\{Z_t\}_{t \geq 0}$. Suppose $\{Z_t\}_{t \geq 0}$ is a stationary and exponential $\beta$-mixing process with $\beta$-mixing coefficient $\beta(q) \leq \beta_0 \exp(-\beta_1q)$ for some $\beta_0 \geq 0$ and $\beta_1 > 0$. Let $\calG$ be  a class of vector-valued functions that take $Z_t$ as input and output an element in $\calH_k$. For any $g \in \calG$, assume $\EE[g(Z_t)] = 0$ for any $t \geq 0$. Suppose the envelop function of $\calG$, defined as $G(z) = \sup_{g \in \calG}\norm{g(z)}_{\calH_k}$, is uniformly bounded by some constant $c_{\calH_k}$, i.e., $\norm{G}_\infty \leq c_{\calH_k}$. In addition, if $\calG$ belongs to the class of VC-typed functions such that $\sup_{\mathbb{P}}\calN(\epsilon c_{\calH_k}, \calG, L^2(\mathbb{P})) \lesssim (1/\epsilon)^\alpha$. Then with probability at least $1 - 1/(NT)$,
	$$
	\sup_{g \in \calG}\left\norm{\frac{1}{NT}\sum_{i = 1}^N\sum_{t = 0}^{T-1}g(Z_{i, t})\right}_{\calH_k} \lesssim \log(NT) \sqrt{\frac{\alpha}{NT}}.
	$$
\end{lemma}

\begin{proof}
	We use the same strategy as Lemma \ref{lm: empirical process for scalar} to prove this lemma. It is sufficient to bound $\sup_{g \in \calG}\norm{\sum_{i = 1}^N\sum_{t = 0}^{T-1}g(Z_{i, t})}_{\calH_k}$. In the following,  we apply Berbee's coupling lemma \cite{berbee1979random} and follow the remark below Lemma 4.1 of \cite{dedecker2002maximal}. Specifically, Let $q$ be some positive integer. We can always construct a sequence $\{\widetilde{Z}_{i, t} \}_{t\geq 0}$ such that with probability at least $1 - (NT\beta(q))/q$,
	$$
	\sup_{g \in \calG}\norm{\sum_{i = 1}^N\sum_{t = 0}^{T-1}g(Z_{i, t})}_{\calH_k} = \sup_{g \in \calG}\norm{\sum_{i = 1}^N\sum_{t = 0}^{T-1}g(\widetilde Z_{i, t})}_{\calH_k},
	$$
	and meanwhile the block sequence $\widetilde X_{i, k}(g) = \{g(\widetilde{Z}_{i, (k-1)q + j})\}_{0 \leq j < q}$ are identically distributed for $k \geq 1$.  In addition, the sequence $\{  \widetilde X_{i, k}(g) \, \mid \, k = 2\omega, \omega \geq 1s  \}$ are independent and so are the sequence $\{  \widetilde X_{i, k}(g) \, \mid \, k = 2\omega + 1, \omega \geq 0  \}$. Let $I_r =\{\floor{T/q}q, \cdots, T-1\}$ with $\text{Card}(I_r) < q$. Then we can show that with probability at least $1 - (NT\beta(q))/q$,
	\begin{align*}
		&\sup_{g \in \calG}\norm{\sum_{i = 1}^N\sum_{t = 0}^{T-1}g(Z_{i, t})}_{\calH_k} \\
		\leq & \sup_{g \in \calG}\norm{\sum_{i = 1}^N\sum_{t = 0}^{q\floor{T/q} - 1}g(\widetilde Z_{i, t})}_{\calH_k} +  \sup_{g \in \calG}\norm{\sum_{i = 1}^N\sum_{t \in I_r}g(Z_{i, t})}_{\calH_k}. 
	\end{align*}
	In the following, assuming that the above inequality holds, we bound each of the above two terms separately. First of all, without loss of generality, we assume $\floor{T/q}$ is an even number. Then for the first term, we have
	\begin{align*}
		& \sup_{g \in \calG}\norm{\sum_{i = 1}^N\sum_{t = 0}^{q\floor{T/q} - 1}g(\widetilde Z_{i, t})}_{\calH_k} \\
		\leq & \sum_{j = 0}^{2q -1}  \sup_{g \in \calG}\norm{\sum_{i = 1}^N\sum_{k = 0}^{\floor{T/q}/2 - 1}g(\widetilde Z_{i, 2kq + j})}_{\calH_k}.
	\end{align*}
	By the construction, 	$\sup_{g \in \calG}\norm{\sum_{i = 1}^N\sum_{k = 0}^{\floor{T/q}/2 - 1}g(\widetilde Z_{i, 2kq + j})}_{\calH_k}$ is a suprema empirical process of i.i.d. vector-valued functions. Then we apply McDiarmid's inequality for vector-valued functions (See Theorem 7 of \cite{rivasplata2018pac}). It can be seen that the bounded difference of $\sup_{g \in \calG}\norm{\sum_{i = 1}^N\sum_{k = 0}^{\floor{T/q}/2-1}g(\widetilde Z_{i, 2kq + j})}_{\calH_k}$ is $c_{\calH_k}$. Then with probability at least $1 - \varepsilon$, we have
	{\small$$
	\left\abs{\sup_{g \in \calG}\left\norm{\sum_{i = 1}^N\sum_{k = 0}^{\floor{T/q}/2 - 1}g(\widetilde Z_{i, 2kq + j})\right}_{\calH_k} - \EE\left[\sup_{g \in \calG}\left\norm{\sum_{i = 1}^N\sum_{k = 0}^{\floor{T/q}/2-1}g(\widetilde Z_{i, 2kq + j})\right}_{\calH_k}\right]\right} \lesssim \sqrt{NT / q} + \sqrt{NT/q \log(1/\varepsilon)}.
	$$
}
	Next, we derive the upper bound for  $\EE\left[\sup_{g \in \calG}\norm{\sum_{i = 1}^N\sum_{k = 0}^{\floor{T/q}/2-1}g(\widetilde Z_{i, 2kq + j})}_{\calH_k}\right]$. By the symmetric argument for vector-valued functions (e.g., Lemmas C.1 and C.2 of \cite{park2022towards}), we can show that
	\begin{align*}
		& \EE\left[\sup_{g \in \calG}\norm{\sum_{i = 1}^N\sum_{k = 0}^{\floor{T/q}/2-1}g(\widetilde Z_{i, 2kq + j})}_{\calH_k}\right]\\
		\leq & 2\EE\left[\sup_{g \in \calG}\norm{\sum_{i = 1}^N\sum_{k = 0}^{\floor{T/q}/2-1}\sigma_{i, 2kq + j} g(\widetilde Z_{i, 2kq + j})}_{\calH_k}\right],
	\end{align*}
	where $\{\sigma_{i, 2kq + j}\}_{1 \leq i \leq N, 0 \leq k \leq \floor{T/q}/2-1}$ are i.i.d. Rademacher  variables. Define the empirical Rademacher complexity as
	$$
	\widehat \calR(\calG) = \frac{1}{N\floor{T/q}/2}\EE\left[\sup_{g \in \calG}\left\norm{\sum_{i = 1}^N\sum_{k = 0}^{\floor{T/q}/2-1}\sigma_{i, 2kq + j} g(\widetilde Z_{i, 2kq + j})\right}_{\calH_k} \given \left\{ \widetilde Z_{i, 2kq + j} \right\}_{1 \leq i \leq N, 0 \leq k \leq \floor{T/q}/2-1} \right].
	$$
	Notice that the entropy condition in the lemma gives that
	$$
	\sup_{\mathbb{P}} \log(\calN(\epsilon c_{\calH_k}, \calG, L^2(\mathbb{P}))) \lesssim \alpha \log(\frac{1}{\epsilon}).
	$$
	Then by Theorem C.12 of \cite{park2022towards} with some slight modification, we can show that
	\begin{align*}
		\widehat \calR(\calG)  & \lesssim  \frac{c_{\calH_k}}{\sqrt{N\floor{T/q}/2}}\int_{0}^{1} \sup_{\mathbb{P}} \sqrt{	 \log(\calN(\epsilon c_{\calH_k}, \calG, L^2(\mathbb{P})))}d\epsilon\\
		& \lesssim \frac{c_{\calH_k}\sqrt \alpha}{\sqrt{N\floor{T/q}/2}}.
	\end{align*}
	This implies that with probability at least $1 - \varepsilon$,
	\begin{align*}
		& \sup_{g \in \calG}\norm{\sum_{i = 1}^N\sum_{k = 0}^{\floor{T/q}/2}g(\widetilde Z_{i, 2kq + j})}_{\calH_k}\\
		\lesssim & c_{\calH_k}\sqrt{\alpha N\floor{T/q}/2} + \sqrt{NT / q} + \sqrt{NT/q \log(1/\varepsilon)}.
	\end{align*}
	
	Next, we bound $\sup_{g \in \calG}\norm{\sum_{i = 1}^N\sum_{t \in I_r}g(Z_{i, t})}_{\calH_k}$. Since for $i \geq 1$, $\sum_{t \in I_r}g(Z_{i, t})$ are i.i.d. sequences. Then by a similar argument as before, we can show that with probability at least $1 - \varepsilon$,
	$$
	\sup_{g \in \calG}\norm{\sum_{i = 1}^N\sum_{t \in I_r}g(Z_{i, t})}_{\calH_k} \lesssim q \left(\sqrt{\alpha N}+ \sqrt{N\log(1/\varepsilon)} \right).
	$$
	Summarizing together and by letting $q = 2\log(NT)$ and $\varepsilon = 1/(NT)$, we can show that  with probability at least $1 - 1/(NT)$, 
	\begin{align*}
		&\sup_{g \in \calG}\norm{\sum_{i = 1}^N\sum_{t = 0}^{T-1}g(Z_{i, t})}_{\calH_k} \\
		\lesssim & \log(NT)\sqrt{NT\alpha}.
	\end{align*}
	This concludes our proof by dividing both sides by $NT$.
\end{proof}

\section{Additional Numerical Details and Results}
\label{sec:appendix_numerical}

\begin{figure*}[ht!]
\hspace{-.3cm}
\centering
 \includegraphics[width=0.6\textwidth]{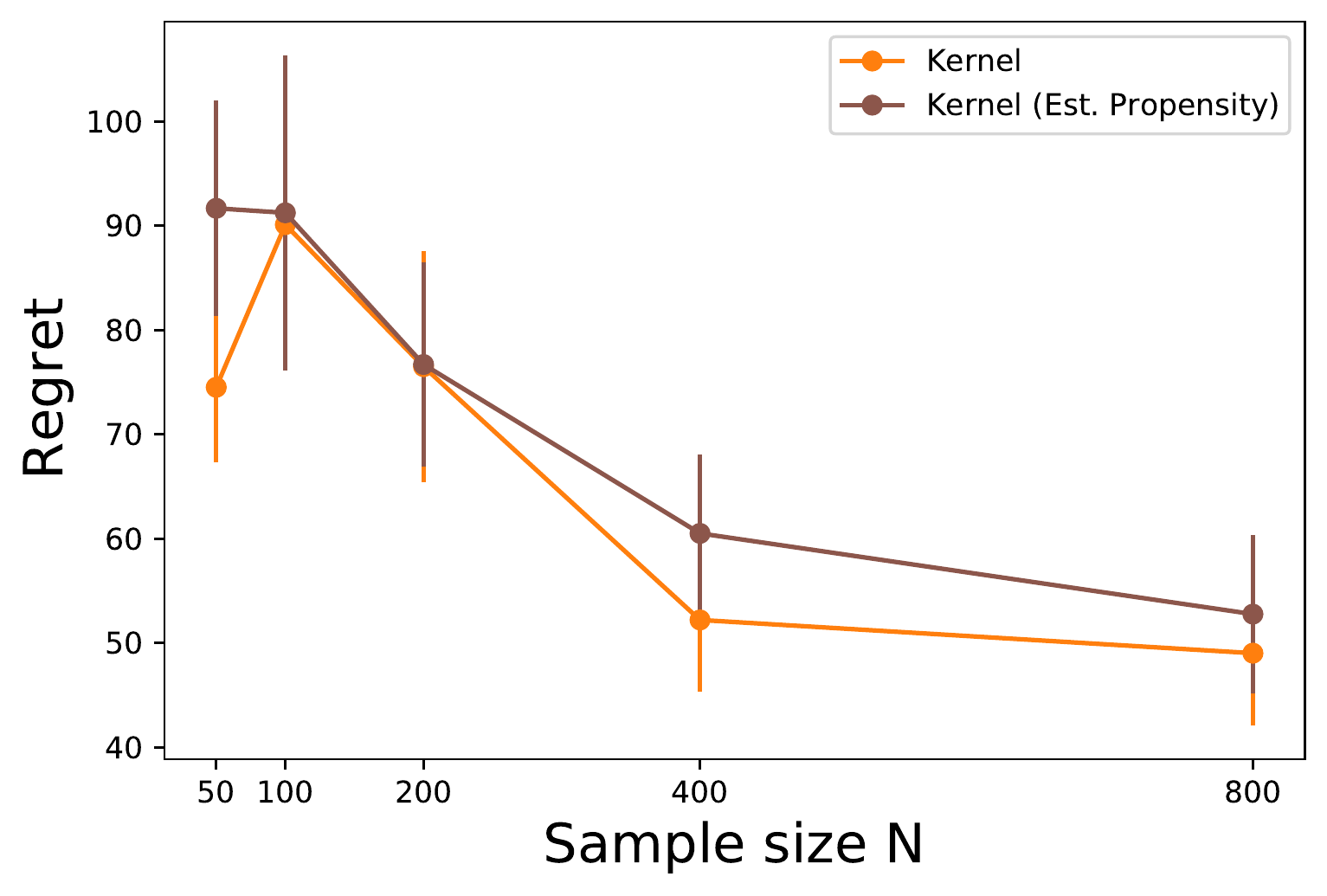} 
\caption{Performance of the kernel-based method: comparison between the variant with the true propensity scores and that with the estimated propensity scores. 
}
\label{fig:Convergence_kernel_est}
\end{figure*}

In Figure \ref{fig:Convergence_kernel_est}, we compare the performance of the kernel-based method with the true propensity scores and that with the estimated propensity scores. 
It is observed that the latter yields higher regrets consistently. 

\begin{figure*}[ht!]
\hspace{-.3cm}
\centering
 \includegraphics[width=0.6\textwidth]{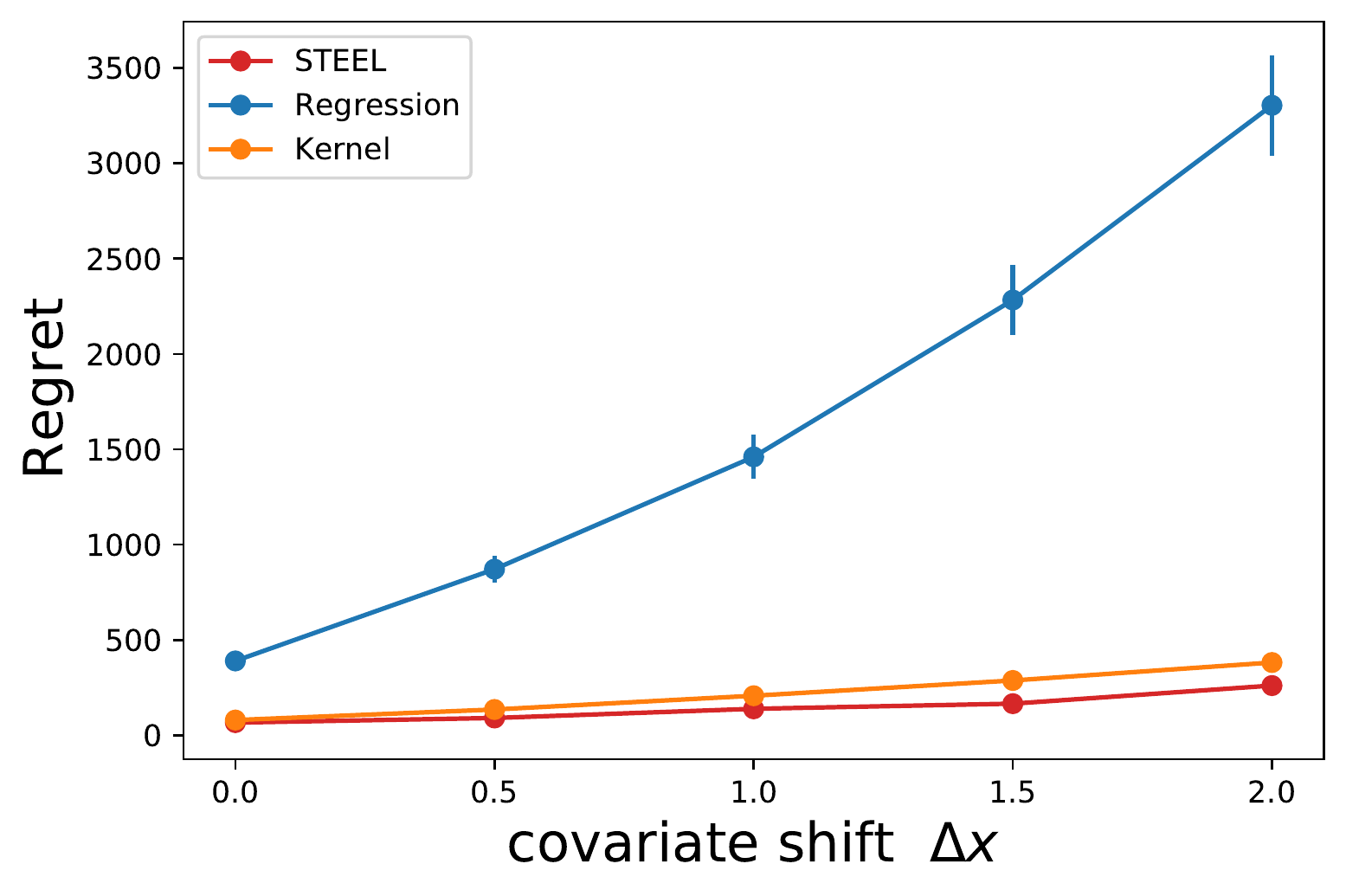} 
\caption{Performance with a non-smooth true reward function. 
}
\label{fig:non_smooth}
\end{figure*}

It is also of interest to study the performance when the reward function class is mis-specified, although we use highly nonlinear functional class as neural networks in our numerical study. 
To investigate that, we modify the true reward function in simulation as follows such that it becomes non-smooth: 
when the first dimension of covariate is larger than 1, we add an 0.5 constant shift to the reward function such that it has a jump near that boundary. 
The results are presented in Figure \ref{fig:non_smooth}. 
It can be observed that STELL still yield desired performance, although the advantage over kernel-methods (which does not rely on a pre-specified reward class) becomes smaller. 


\end{document}